\documentclass[10pt,journal,compsoc]{IEEEtran}
\usepackage[utf8]{inputenc} %
\usepackage[T1]{fontenc}    %
\usepackage{array}
\usepackage[caption=false,font=normalsize,
labelfont=sf,textfont=sf]{subfig}
\usepackage{textcomp}
\usepackage{stfloats}
\usepackage{url}
\usepackage{verbatim}
\usepackage{graphicx}
\def\BibTeX{{\rm B\kern-.05em{\sc i\kern-.025em b}\kern-.08em
    T\kern-.1667em\lower.7ex\hbox{E}\kern-.125emX}}
\usepackage{balance}
\usepackage[square, sort, numbers]{natbib}

\usepackage{anyfontsize}

\usepackage{amsmath,amsfonts,bm}

\def\eqref#1{equation~\ref{#1}}

\def\1{\bm{1}}

\DeclareMathAlphabet{\mathsfit}{\encodingdefault}{\sfdefault}{m}{sl}
\SetMathAlphabet{\mathsfit}{bold}{\encodingdefault}{\sfdefault}{bx}{n}

\newcommand{\R}{\mathbb{R}}

\usepackage{hyperref}
\usepackage{url}
\usepackage{url}            %
\usepackage{booktabs}       %
\usepackage{nicefrac}     
\usepackage{comment}
\PassOptionsToPackage{table, dvipsnames}{xcolor}
\PassOptionsToPackage{table, dvipsnames}{colortbl}
\usepackage{pifont} %
\usepackage{wrapfig}

\usepackage{algorithm}
\usepackage[italicComments=true]{algpseudocodex}

\usepackage[noabbrev, capitalize, nameinlink]{cleveref}
\definecolor{mygreen}{rgb}{0.1,0.5,0.1}
\definecolor{pastelgreen}{rgb}{0.2,0.5,0.3}
\definecolor{pastelorange}{rgb}{0.8,0.4,0.1}
\definecolor{tablegreen}{rgb}{0.85,1.,0.85}
\definecolor{tableorange}{rgb}{1.,0.77,0.77}
\definecolor{myorange}{rgb}{1.,0.6,0.0}
\definecolor{myblue}{HTML}{4885ED}
\definecolor{citered}{rgb}{0.5, 0.0, 0.0}
\definecolor{citeblue}{rgb}{0.0, 0.0, 0.5}

\definecolor{bestcolor}{rgb}{0.1,0.6,0.1}
\definecolor{secondcolor}{rgb}{0.28,0.3,0.93} 

\newcommand{\best}[1]{\textbf{\textcolor{bestcolor}{#1}}}
\newcommand{\second}[1]{\textit{\textcolor{secondcolor}{#1}}}

\newif\ifshowcomments
\showcommentstrue   %
\ifshowcomments
    
    \newcommand{\mfr}[1]{\textcolor{red}{#1}}
    \newcommand{\adri}[1]{\textcolor{purple}{[adri: #1]}}
    \newcommand{\nuria}[1]{\textcolor{magenta}{[nuria: #1]}}
\else
    
    \newcommand{\mfr}[1]{\textcolor{red}{}}
    \newcommand{\adri}[1]{}
    \newcommand{\nuria}[1]{}
\fi

\newcommand{\tick}{\cellcolor{tablegreen}\textcolor{mygreen}{\ding{51}}}
\newcommand{\cross}{\cellcolor{tableorange}\textcolor{red}{\ding{55}}}

\newcommand{\triang}{{\tiny \ding{115}}}
\newcommand{\crossb}{{\scriptsize \ding{59}}}

\DeclareMathOperator{\D}{\mathcal{D}}
\DeclareMathOperator{\T}{\mathcal{T}}
\DeclareMathOperator{\Acc}{{\scriptstyle\text{Acc}}}
\DeclareMathOperator{\TPR}{{\scriptstyle\text{TPR}}}
\DeclareMathOperator{\TNR}{{\scriptstyle\text{TNR}}}
\DeclareMathOperator{\FPR}{{\scriptstyle\text{FPR}}}
\DeclareMathOperator{\FNR}{{\scriptstyle\text{FNR}}}

\DeclareMathOperator{\EOp}{{\scriptstyle\text{EOp}}}
\DeclareMathOperator{\EOds}{{\scriptstyle\text{EOdds}}}
\DeclareMathOperator*{\Exp}{\mathbb{E}}

\begin{document}

\title{Towards Algorithmic Fairness by means of\\Instance-level Data Re-weighting\\based on Shapley Values}
\author{Adrian Arnaiz-Rodriguez, Nuria Oliver
\thanks{The authors are with ELLIS Alicante, Spain (e-mail: adrian@ellisalicante.org; nuria@ellisalicante.org)}}

\markboth{Arnaiz-Rodriguez and Oliver: FairShap}
{Arnaiz-Rodriguez, Oliver, 
\MakeLowercase{: Towards Algorithmic Fairness by means of Instance-level Data Re-weighting based on Shapley Values}}

\maketitle

\begin{abstract}
Algorithmic fairness is of utmost societal importance, yet state-of-the-art large-scale machine learning models require training with massive datasets that are frequently biased. In this context, pre-processing methods that focus on modeling and correcting bias in the data emerge as valuable approaches. In this paper, we propose \texttt{FairShap}, a novel instance-level data re-weighting method for fair algorithmic decision-making through data valuation by means of Shapley Values. \texttt{FairShap} is model-agnostic and easily interpretable. It measures the contribution of each training data point to a predefined fairness metric. We empirically validate \texttt{FairShap} on several state-of-the-art datasets of different nature, with a variety of training scenarios and machine learning models and show how it yields fairer models with similar levels of accuracy than the baselines. We illustrate \texttt{FairShap}'s interpretability by means of histograms and latent space visualizations. Moreover, we perform a utility-fairness study and analyze \texttt{FairShap}'s computational cost depending on the size of the dataset and the number of features. We believe that \texttt{FairShap} represents a novel contribution in interpretable and model-agnostic approaches to algorithmic fairness that yields competitive accuracy even when only biased training datasets are available.
\end{abstract}


\section{Introduction}
\IEEEPARstart{M}{achine} learning (ML) models are increasingly used to support human decision-making in a broad set of use cases, including in high-stakes domains, such as healthcare, education, finance, policing, or immigration. In these scenarios, algorithmic design, implementation, deployment, evaluation and auditing should be performed cautiously to minimize the potential negative consequences of their use, and to develop fair, transparent, accountable, privacy-preserving, reproducible and reliable systems~\citep{barocas2017fairness,smuha2019ethics,oliver2022srip}. 
To achieve algorithmic fairness, a variety of fairness metrics that mathematically model different definitions of equality have been proposed in the literature~\citep{carey2022statistical}. Group fairness focuses on ensuring that different demographic groups are treated fairly by an algorithm~\citep{hardt2016equality, gomez2017eodds}, and individual fairness aims to give a similar treatment to similar individuals~\citep{dwork2012fairness}. 
In the past decade, numerous machine learning methods have been proposed to achieve algorithmic fairness~\citep{mehrabi2021survey}. 

Algorithmic fairness may be addressed in the three stages of the ML pipeline: first, by modifying the input data (\emph{pre-processing}) via e.g. re-sampling, data cleaning, re-weighting or learning fair representations~\citep{kamiran2012data, zemel2013learning}; second, by including a fairness metric in the optimization function of the learning process (\emph{in-processing})~\citep{zhang2018mitigating, kamishima2012fairness}; and third, by adjusting the model's decision threshold to ensure fair decisions across groups (\emph{post-processing})~\citep{hardt2016equality}. These approaches are not mutually exclusive and may be combined to obtain better results.

From a practical perspective, pre-processing fairness methods tend to be easier to understand for a diverse set of stakeholders, including legislators~\citep{feldman2015certifying, hacker2022varieties}. Furthermore, to mitigate potential biases in the data, there is increased societal interest in using demographically-representative data to train ML models~\citep{madaio2022assessing, gebru2021datasheets, hagendorff2020ethics}. 
However, the vast majority of the available datasets used to train ML models in real world scenarios are not demographically representative and hence could be biased. Moreover, datasets that are carefully created to be fair lack the required size and variety to train large-scale deep learning models.

In this context, pre-processing algorithmic fairness methods that focus on modeling and correcting bias in the data emerge as valuable approaches~\citep{chouldechova2020snapshot}. Methods of special relevance are those that identify the value of each data point not only from the perspective of the algorithm's performance, but also from a fairness perspective~\citep{feldman2015certifying}, and methods that are able to leverage small but fair datasets to improve fairness when learning from large-scale yet biased datasets. 

\emph{Data valuation} approaches are particularly well suited for this purpose. 
The data valuation methods proposed to date~\citep{ghorbani2019data} measure the contribution of each data point to the utility of the model --usually defined as accuracy-- and use this information as a pre-processing step to improve the performance of the model.
However, they have not been used for algorithmic fairness. In this paper, we fill this gap by proposing \texttt{FairShap}, an instance-level, data re-weighting method for fair algorithmic decision-making which is model-agnostic and interpretable through data valuation. \texttt{FairShap} leverages the concept of Shapley Values~\citep{shapley1953value} to measure the contribution of \emph{each} data point to a pre-defined group \emph{fairness} metric. As the weights are computed on a reference dataset ($\T$), \texttt{FairShap} makes it possible to use fair but small datasets to debias large yet biased datasets. 

\begin{figure*}[t]
\centering
\includegraphics[width=.95\textwidth]{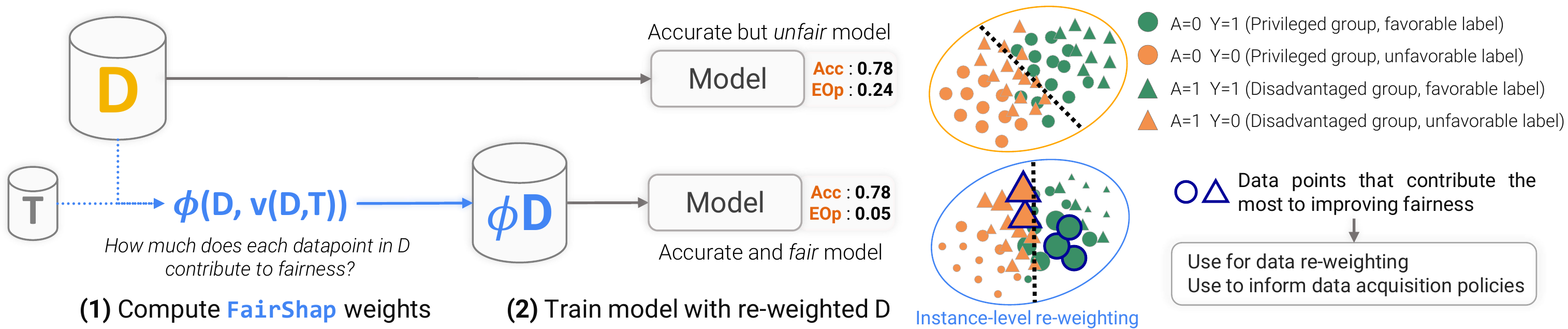}
\caption{Left: \texttt{FairShap}'s workflow. The weights are computed using a reference dataset $\T$, which can be an external dataset or the validation set of $D$. Right: Illustrative example of \texttt{FairShap}'s impact on individual instances and on the decision boundary. Note how data re-weighting with \texttt{FairShap} is able to shift the data distribution yielding a fairer model with similar levels of accuracy.}
\label{fig:fig1}
\end{figure*}

\cref{fig:fig1} illustrates the workflow of data re-weighting by means of \texttt{FairShap}: First, the weights for each data point in the training set $x_i$, $\Phi_i$, are computed by leveraging a reference dataset $\T$ which is either a fair dataset --when available-- or the validation set of the dataset $D$. Second, once the weights are obtained, the training data is re-weighted. Third, an ML model is trained using the re-weighted data and then applied to the test set. In addition, the weights obtained via \texttt{FairShap} may be used to inform data selection policies to automatically identify the data points that would yield models with competitive levels of both performance and fairness. \cref{sec:pruning} presents the experimental results of using \texttt{FairShap} for this purpose. 

\texttt{FairShap} has several advantages: (1) it is easily interpretable, as it assigns a numeric value (weight) to each data point in the training set; (2) it enables detecting which data points are the most important to improve fairness while preserving accuracy; (3) it makes it possible to leverage small but fair datasets to learn fair models from large-scale yet biased datasets; and (4) it is model agnostic.

\section{Related Work} \label{sec:related}
\noindent \textbf{Group Algorithmic Fairness.} Group bias in algorithmic decision-making is based on the conditional independence between the joint probability distributions of the sensitive attribute ($A$), the label ($Y$), and the predicted outcome ($\hat{Y}$). Barocas et al. \citep{barocas2017fairness} define three concepts used to evaluate algorithmic fairness: \emph{independence} ($\hat{Y}\bot A$), \emph{separation} ($\hat{Y}\bot A | Y$), and \emph{sufficiency} ($Y\bot A | \hat{Y}$). The underlying idea is that a \emph{fair} classifier should have the same error classification rates for different protected groups. Three popular metrics to assess group algorithmic fairness are --from weaker to stronger notions of fairness--  \emph{demographic parity} (DP), i.e. equal acceptance rate~\citep{dwork2012fairness, gomez2017eodds}; \emph{equal opportunity} (EOp), i.e. equal true positive rate, TPR, for all groups~\citep{chouldechova2017fair, hardt2016equality}; and \emph{equalized odds} (EOdds), i.e. equal TPR and false positive rate, FPR, for all groups~\citep{gomez2017eodds, hardt2016equality}. Numerous algorithms have been proposed to maximize these metrics while maintaining  accuracy~\citep{mehrabi2021survey}. \texttt{FairShap} focuses on improving the two strongest of these group-based fairness metrics: EOp and EOdds.

\textbf{Data Re-weighting for Algorithmic Fairness.}  
Data \emph{re-weighting} is a pre-processing technique that assigns weights to the training data to optimize a certain fairness measure. Compared to other pre-processing approaches, data re-weighting is easily interpretable~\citep{barocas2016big}. There are two broad approaches to perform data re-weighting: group and instance-level re-weighting. 

In \emph{group re-weighting}, the same weight is assigned to all data points belonging to the same group, which in the context of algorithmic fairness are defined according to their values of a sensitive attribute $A$ (e.g. gender, race, age, etc...). Kamiran and Calders re-weight the groups defined by $A$ and $Y$ based on statistics of the under-represented label(s) and the disadvantaged group(s) in a model-agnostic manner~\citep{kamiran2012data}. Krasanakis et al. assume that there is an underlying set of labels that would correspond to an unbiased distribution and use an inference model based on label error perturbation to define weights that yield better fairness performance \citep{krasanakis2018adaptive}. Other authors have proposed adjusting the loss function values in the sensitive groups to iteratively learn weights that address the labeling bias and thus improve the fairness of the models \citep{jiang2020identifying}. Chai and Wang find the weights by solving an optimization problem that entails several rounds of model training \citep{chai2022adaptative}. Finally, re-weighing has been combined with a regularization term to adjust the weights in an iterative optimization process based on distributionally robust optimization (DRO) \citep{jung2023reweighting}.

However, note that many of these works \citep{krasanakis2018adaptive, jiang2020identifying, chai2022adaptative, jung2023reweighting} propose re-weighting methods that adjust the weights repeatedly through an ongoing learning process, thus resembling in-processing rather than pre-processing approaches~\citep{caton2020fairness} as the computed weights depend on the model. This iterative process adds uncertainty to the weight computation~\citep{ali2021accounting} and requires retraining the model in each iteration, which could be computationally very costly or even intractable for large datasets and/or complex models. Conversely, data-valuation methods are based on the concept that the value of the data should be orthogonal to the choice of the learning algorithm and hence data-valuation approaches should be purely data-driven and hence model-agnostic~\citep{sim2022data}.

In contrast to group re-weighting, \emph{instance-level re-weighting} seeks to assign individual weights to each data point by considering the protected attributes and the sample misclassification probability. 
Most of the previous work has proposed the use of Influence Functions (IFs) for instance-level re-weighting. IFs estimate the changes in model performance when specific points are removed from the training set by computing the gradients or Hessian of the model~\citep{koh2017understanding, pruthi2022datainfluence, paul2021dataimportance, Sundararajan2017axiomatic}.
In the context of fairness, IFs have been used to estimate the impact of data points on fairness metrics. Black and Fredrikson propose a leave-one-out (LOO) method to estimate such an influence~\citep{black2021loofairness}. In~\citep{wang2022instancelevel}, the data weights are estimated by means of a neural tangent kernel by leveraging a kernelized combination of training examples. Finally, Li and Hiu propose an algorithm that uses the Hessian of the matrix of the loss function to estimate the effect of changing the weights to identify those that most improve the fairness of the model~\citep{li2022individualreweighing}. 

While promising, IFs are not exempt from limitations, such as their fragility, their dependency on the model --and thus making them in-processing rather than pre-processing methods, their need for strongly convex and twice-differentiable models~\citep{basu2021influence} and their limited interpretability, which is increasingly a requirement by legal stakeholders~\citep{feldman2015certifying, hacker2022varieties}. Also, IFs approximate the leave-one-out score only for strongly convex loss functions, which limits the analysis by overlooking correlations between data points~\citep{koh2017understanding, kwon2022betashap, hammoudeh2024training}. In contrast, Shapley Values (SV) have demonstrated greater stability than LOO and superior performance in data selection tasks, as well as in both stochastic and deterministic learning scenarios~\citep{wang2023data}.
Finally, IFs do not satisfy validated properties that have been attributed to data valuation methods, such as the awareness to data preference, which are essential to making the methods more precise, practical and interpretable \citep{ghorbani2019data,wu2022davinz}. 

\textbf{Data Valuation.} Data valuation (DV) methods, such as the \emph{Shapley Value}~\citep{shapley1953value} or \emph{Core}~\citep{gillies1959core}, measure how much a player contributes to the total utility of a team in a given coalition-based game. 
They have shown promise in several domains and tasks, including federated learning~\citep{wang2019measure}, data minimization~\citep{brophy2020exit}, data acquisition policies, data selection for transfer learning, active learning, data sharing, exploratory data analysis and mislabeled example detection~\citep{schoch2022csshapley}. 

In the ML literature, Shapley Values (SVs) have been proposed to tackle a variety of tasks, such as transfer learning and counterfactual generation~\citep{fern2021text, albini2022counterfactual}. In the eXplainable AI (XAI) field \citep{molnar2020interpretable}, SVs have been used to achieve feature explainability by measuring the contribution of each feature to the individual prediction~\citep{lundberg2017unified}. Ghorbani and Zhou recently proposed an instance-level data re-weighting approach by means of the SVs to determine the contribution of each data point to the model's accuracy \citep{ghorbani2019data}. In this case, the SVs are used to modify the training process or to design data acquisition/removal policies. The goal is to maximize the model's accuracy in the test set. We are not aware of any peer-reviewed publication where SVs are used in the context of algorithmic fairness.

In this paper, we propose \texttt{FairShap}, an interpretable, instance-level data re-weighting method for algorithmic fairness based on SVs for data valuation. We direct the reader to Table 4 %
in the Appendix for a comparison between \texttt{FairShap} and related methods regarding their desirable qualities. 
In addition to data re-weighting, \texttt{FairShap} may be used to inform data acquisition policies.

\section{Background}
\subsection{Data Valuation}
\label{sec:sv}

\noindent Let $\D=\{(x_i,y_i)\}_{i=1}^n$ be the dataset used to train a machine learning model $M$. The Shapley Value (SV) of a data point $(x_i,y_i)$ --or $i$ for short-- that belongs to the dataset $\D$ is a data valuation function, $\phi_i(\D,v)\in \R$ --or $\phi_i(v)$ for short, that estimates the contribution of each data point $i$ to the performance or valuation function $v(M,\D,\T)$ --or $v(\D)$ for short-- of model $M$ trained with dataset $\D$ and tested on \emph{reference} dataset $\T$, which is either an external dataset or a subset of $\D$. The Shapley Value is given by Eq. \ref{eq:sv}. Note how its computation considers all subsets $S$ in the powerset of $\D$, $\mathcal{P}(\D)$. 
\begin{equation} \label{eq:sv}
\phi_i(\D,v) := \frac{1}{|\D|} \sum_{S\in \mathcal{P}(\D \backslash \{i\})} \frac{v(S\cup \{i\})-v(S)}{\binom{|\D|-1}{|S|}}
\end{equation}
The valuation function $v(\D)$ is typically defined as the accuracy of $M$ trained with dataset $\D$ and tested with $\T$. In this case, the Shapley Value, $\phi_i(\Acc)$, measures how much each data point $i \in \D$ contributes to the accuracy of $M$. The values, $\phi_i(\Acc)$, might be used for several purposes, including domain adaptation data re-weighting~\citep{ghorbani2019data}. 

\textbf{Axiomatic properties of the Shapley Values.}\label{par:svaxioms} The SVs satisfy the following axiomatic properties: 

    \emph{Efficiency}:  $v(\D)=\sum_{i \in \D} \phi_i(v)$, i.e. the value of the entire training dataset $\D$ is equal to the sum of the Shapley Values of each of the data points in $\D$.
    
    \emph{Symmetry}: $\forall S \subseteq \D: v(S \cup {i}) = v(S\cup {j}) \rightarrow \phi_i=\phi_j$, i.e. if two data points add the same value to the dataset, their Shapley Values must be equal.
    
    \emph{Additivity}: $\phi_i(\D,v_1+v_2)=\phi_i(\D,v_1)+\phi_i(\D,v_2)$, $\phi_i(D,v_1+v_2)=\phi_i(D,v_1)+\phi_i(D,v_2)$, i.e. if the valuation function is split into additive 2 parts, we can also compute the Shapley values in 2 additive parts.
    
    \emph{Null Element}: $\forall S \subseteq \D: v(S \cup {i}) = v(S) \rightarrow \phi_i=0$, i.e. if a data point does not add any value to the dataset then its Shapley value is 0.

\subsection{Algorithmic Fairness}

The group fairness criteria addressed in this work are based on the statistical relationships between the predicted outcomes $\hat{Y}$, the observed outcomes $Y$, and the protected group attribute $A$. Following common practice and without loss of generality, we consider $\hat{Y} = Y = 1$ as the positive outcome and $A = a$ as the advantaged group. In this work, we focus on two widely recognized fairness metrics from the literature~\citep{mehrabi2021survey}: Equalized Odds (EOdds) and Equal Opportunity (EOp)

First, \emph{equal opportunity}~\citep{hardt2016equality} defines fairness as predicting the positive outcome independently of the protected group attribute, conditioned on the observed outcome being positive. Formally, equal opportunity is defined as:
\begin{align} \label{eq:equal_opportunity}
    P(\hat{Y} = 1 \mid A = a, Y = 1) =  P(\hat{Y} = 1 \mid A = b, Y = 1).
\end{align}

Second, \emph{equalized odds}~\citep{hardt2016equality} is a stricter version of equal opportunity, requiring predictive independence conditioned on both positive and negative observed outcomes. Formally, equalized odds is defined as:
\begin{align} \label{eq:equalized_odds}
    P(\hat{Y} = 1 \mid A = a, Y = y)  =  P(\hat{Y} = 1 \mid A = b, Y = y)
\end{align}
with $y \in \{0, 1\}$.

These fairness criteria are measured in practice as follows. For EOp, the fairness metric is the difference in TPR between different groups:
\begin{align} \label{eq:equal_opportunity_met}
\text{EOp} := \text{TPR}_{A=a} - \text{TPR}_{A=b}.
\end{align}
For EOdds, the fairness metric considers both the TPR and FPR differences between groups:
\begin{align} \label{eq:equalized_odds_met}
\text{EOdds} := &\frac{1}{2}(\text{FPR}_{A=a} - \text{FPR}_{A=b}) \\\nonumber
+ &\frac{1}{2}(\text{TPR}_{A=a} - \text{TPR}_{A=b}).
\end{align}

\section{\texttt{FairShap}: Fair Shapley Value} \label{sec:fsv}
\noindent \texttt{FairShap} proposes valuation functions that consider the model's fairness while sharing the same axioms as the Shapley Values. Specifically, \texttt{FairShap} considers the family of group fairness metrics that are defined by TPR, TNR, FPR, FNR and their $A$-$Y$ conditioned versions, namely Equalized Odds and Equal Opportunity. 

A straightforward implementation of \texttt{FairShap} is intractable. Thus, to address such a limitation, \texttt{FairShap} leverages the efficiency axiom of the SV and the decomposability of the fairness metrics~\citep{gultchin2022beyond, wang2022instancelevel}%
To obtain fair data valuations, \texttt{FairShap} computes $\phi(\D,v)$ by means of a distance-based approach and a reference dataset, $\T$, which may be a small and fair external dataset or a partition (typically the validation set) of $\D$. The resulting model trained with the re-weighted dataset according to \texttt{FairShap}'s weights maintains similar levels of accuracy while increasing its fairness. Furthermore, no model is required to compute the fair data valuations and thus \texttt{FairShap} is model agnostic.

In the following, we derive the expressions to compute the weights of a dataset according to \texttt{FairShap} in a binary classification case ($Y$ is a binary variable) and with binary protected attributes. The extension to non-binary protected attributes and multi-class scenarios is provided in  Appendix~C.5. %
We first present the necessary building blocks for \texttt{FairShap}: the pairwise contribution of training point $i$  to the correct classification of test point $j$ and the SVs for the probabilities of predicted label given the actual one (TPR, TNR, FPR, FNR).

\textbf{Pairwise contributions $\Phi_{i,j}$.} Statistical algorithmic fairness depends on the disparity in a model's error rates on different groups of data points in the test set when the groups are defined according to their values of a protected attribute, $A$. 
To measure the data valuation for a training data point to the fairness of the model, it is essential to identify the contribution of that training data point to the model's accuracy on the different groups of the test set defined by their protected attribute. 

Let $\Phi_{i,j}$ be the contribution of the training point $(x_i,y_i) \in \D$ to the probability of correct classification of the test point $(x_j, y_j) \in \T$. $\Phi_{i,j}$ measures the expected change in the model's correct prediction of $j$ due to the inclusion of $i$ in the dataset, namely: 
\begin{align}\label{eq:pairwiseSV}
    \Phi_{i,j} = \mathbb{E}_{S\sim\mathcal{P}(\D \backslash \{i\})}[&p(y=y_j|x_j, S \cup \{i\}) \\\nonumber
    &- p(y=y_j|x_j, S)],
\end{align}
where $p(y=y_j|x_j, S \cup \{i\}) - p(y=y_j|x_j, S)$ is the leave-one-out contribution of the training datapoint $i$ to the correct classification of the test point $j$ in a model trained with the dataset $S$, i.e. $\text{LOO}(i,j,S)$. Let $\bm{\Phi} \in \R^{|\D|\times |\T|}$ be the matrix where each element corresponds to the contribution of the pairwise train-test data points, leveraging the efficiency axiom, ${\phi_i(\Acc) := \mathbb{E}_{j \sim p(\T)}[\Phi_{i,j}]= \overline{\bm{\Phi}}_{i,:}\in \R}$.
While a direct implementation of $\Phi_{i,j}$ is very expensive to compute ($O(2^N)$), an efficient implementation (${O(N \log N)}$) by Jia et al. \citep{jia2019knn} (see Appendix C.1) %
is available. It consists of a closed-form solution of $\phi(\Acc)$ by means of a deterministic distance-based approach and thus model independent. 
Although only exact for $k$-NN models, this method is able to efficiently compute the SVs both in the case of tabular and non-structured (embeddings) data~\citep{jiang2023opendataval}. It yields very efficient runtime performance when compared to other estimators of the SVs for data valuation while avoiding the errors of  Monte Carlo approximations of the SVs. Note that %
\citep{jia2019knn} does not delve into the formulation, meaning and possible uses of $\Phi_{i,j}$, as we do in this paper. 

\textbf{Fair Shapley Values.}  TPR and TNR are the building blocks of the group fairness metrics that \texttt{FairShap} uses as valuation functions: EOdds and EOp. Let $\phi_i(\TPR)$ and $\phi_i(\TNR)$ be two valuation functions that measure the contribution of training point $i$ to the TPR and TNR, respectively. Note that $\text{TPR}=\text{Acc}|_{Y=1}$ and $\text{TNR}=\text{Acc}|_{Y=0}$.
Therefore, 
$\phi_i(\TPR)$ corresponds to the expected change in the model's probability of correctly predicting the positive class when point $i$ is included in the training dataset $\D$, considering all possible training dataset subsets and the distribution of the reference dataset:
\begin{align} 
\phi_i(\TPR) :=&\ \mathbb{E}_{j \sim p(\T|Y=1)}\left[\mathbb{E}_{S\sim\mathcal{P}(\D \backslash \{i\})}\left[\text{LOO}(i,j,S)\right]\right] \label{eq:svtpr} \\
=&\ \mathbb{E}_{j \sim p(\T|Y=1)}\left[\Phi_{i,j}\right]. \nonumber
\end{align}
The value for the entire dataset is $\bm{\phi(\TPR)} = [\phi_0(\TPR), \cdots, \phi_n(\TPR)] \in \R^{|\D|}$. $\bm{\phi(\TNR)}$ is obtained similarly but for $Y=0$ and $y=0$. In addition, $\phi_i(\FNR) = \frac{1}{|\D|} - \phi_i(\TPR)$ and $\phi_i(\FPR)= \frac{1}{|\D|} - \phi_i(\TNR)$. These four functions fulfill the SV axioms. More details about these metrics can be found in Appendix B.%

Intuitively, $\bm{\phi(\TPR)}$ and $\bm{\phi(\TNR)}$ quantify how much the training data points contribute to the correct classification when $y=1$ and $y=0$, respectively. To illustrate $\bm{\phi(\TPR)}$ and $\bm{\phi(\TNR)}$, Figure 9 in Appendix D.1. %
depicts the $\bm{\phi(\TPR)}$ and $\bm{\phi(\TNR)}$ of a simple synthetic example with two normally distributed classes.

Once $\phi_i(\TPR)$, $\phi_i(\TNR)$, $\phi_i(\FPR)$ and $\phi_i(\FNR)$ have been obtained, we can compute the \texttt{FairShap} weights for a given dataset. However, there are two scenarios to consider, depending on whether the sensitive attribute ($A$) and the target variable or label ($Y$) are the same or not.

\textbf{FairShap when $A$ = $Y$.} In this case, the group fairness metrics (Eop and EOdds) collapse to measure the disparity between TPR and TNR or FPR and FNR for the different values of $Y$~\citep{berk2021fairness}, which, in a binary classification case, may be expressed as the Equal Opportunity measure computed as $\text{EOp}:=\text{TPR}-\text{FPR} \in [-1,1]$ or its scaled version $\text{EOp}=(\text{TPR}+\text{TNR})/2 \in [0,1]$. Thus, the $\phi_i(\EOp)$ of data point $i$ may be expressed as
\begin{equation} \label{eq:sveopaeqy} 
\phi_i(\EOp):=\frac{\phi_i(\TPR)+\phi_i(\TNR)}{2}.
\end{equation}

For more details on the equality of the group fairness metrics when $A=Y$ and how to obtain $\phi_i(\EOp)$, we refer the reader to Appendix C.3. %

\textbf{FairShap when $A$ $\neq$ $Y$.} \label{sec:AneqYSV} This is the most common scenario. In this case, group fairness metrics, such as EOp or EOdds, use true/false positive/negative rates conditioned not only on $Y$, but also on $A$. Therefore, we define $\text{TPR}_{A=a}=\text{Acc}_{Y=y, A=a}$, or $\text{TPR}_a$ for short, and thus
\begin{equation}
 \phi_i(\TPR_{a}) := \mathbb{E}_{j \sim p(\T|Y=1, A=a)}[\Phi_{i,j}] = \overline{\bm{\Phi}}_{i,:|Y=1, A=a} \label{eq:svtpra} 
\end{equation}
where the value for the entire dataset is 
 $\bm{\phi}(\TPR_a) = [\phi_0(\TPR_a), \cdots, \phi_n(\TPR_a)]$. Intuitively, $\phi_i(\TPR_{a})$ measures the contribution of the training point $i$ to the TPR of the testing points belonging to a given protected group ($A=a$). $\phi_i(\TNR_{a})$ is obtained similarly but for $y = 0$.

Given EOp and EOdds as per \cref{eq:equal_opportunity_met,eq:equalized_odds_met}, then $\phi_i(\EOp)$ is given by
\begin{equation} \label{eq:sveop}
\phi_i(\EOp) := \phi_i(\TPR_a) - \phi_i(\TPR_b)
\end{equation}
and $\phi_i(\EOds)$ is expressed as 
\begin{equation} \label{eq:sveodds}
\phi_i(\EOds) := \frac{1}{2} (\phi_i(\FPR_a)-\phi_i(\FPR_b))+ \frac{1}{2}(\phi_i(\TPR_a)-\phi_i(\TPR_b))
\end{equation}
where their corresponding $\bm{\phi(\EOp)}$ and $\bm{\phi(\EOds)}$ vectors.
A detailed view on the complete formula and a step-by-step derivation of the equations above can be found in Appendices B and C.4.  %
Additionally, Appendix B.1. %
present a synthetic example showing the impact of $\phi(\cdot)$ on the decision boundaries and the fairness metrics. 

 \cref{alg:fairshap} provides the pseudo-code to compute the data weights according to \texttt{FairShap}.

\begin{algorithm}[t]
\begin{small}
\caption{\texttt{FairShap}: Instance-level data re-weighting for group algorithmic fairness by means of data valuation}
\label{alg:fairshap}
\begin{algorithmic}[1]
\State \textbf{Input} Training set $\D$, reference set $\T$, protected groups $A$, parameter $k$

\Procedure{\texttt{FairShap}}{$\D$, $\T$, $k$}
    \State Initialize $\bm{\Phi}$ as a matrix of zeros with dimensions $|\D| \times |\T|$
    
    \For{$j$ in $\T$} %
        \State Order $i \in \D$ according to the $L_2$ distance to $j \in \T \rightarrow (x_1, x_2, \ldots, x_N)$
        \State Compute $\Phi_{N,j} = \frac{\mathbb{I}[y_{x_N}=y_j]}{N}$
        
        \For{$i$ from $N-1$ to $1$} 
            \LComment{How much does $i$ contribute to $j$'s likelihood of correct classification ($\Phi_{i,j}$)?}
            \State $\Phi_{i,j} = \Phi_{i+1, j} + \mathbb{I}[y_i=y_{j}] - I[y_{i+1}=y_{j}]\frac{1}{\max(k,i)}$
        \EndFor
    \EndFor
    \vskip 0.1in
    \State $\bm{\phi}(\TPR_{a})\:=\text{\cref{eq:svtpra}}$
    \State $\bm{\phi}(\FPR_{a}) \:= [\phi_i(\FPR_a)= \frac{1}{|\D|} - \phi_i(\TNR_a):\: \forall i\in \mathcal{D}]\: \forall a\in A$
    \State  $\bm{\phi(\EOp)}\:\:\:= \text{\cref{eq:sveop}}$
    \State $\bm{\phi(\EOds)}=\text{\cref{eq:sveodds}}$ 
    \State \textbf{Output}:
    \State Shapley Value matrix $\bm{\Phi} \in \mathbb{R}^{|\D| \times |\T|}$
    \State \texttt{FairShap} arrays $\bm{\phi(\EOp)} \in \mathbb{R}^{|\D|}$ and $\bm{\phi(\EOds)} \in \mathbb{R}^{|\D|}$
\EndProcedure
\end{algorithmic}
\end{small}
\end{algorithm}

\textbf{Instance-level data re-weighting with \texttt{FairShap}} The definition of a data valuation function states that the higher the value, the more the data point contributes to the measure. Yet, it does not necessarily mean that a higher value is more desirable: it depends on the value function of choice. In the case of accuracy, a higher value denotes a larger contribution to accuracy \citep{ghorbani2019data}. In the case of fairness, we prioritize points with high $-\phi(\EOp)=\phi_i(\TPR_B)-\phi_i(\TPR_A)$, where $B$ is the discriminated group (i.e. $\TPR_A>\TPR_B$). Giving more weight to data points with a positive $-\phi_i(\EOp)$ will contribute to increasing the TPR of the discriminated group, balancing the difference in TPR between groups and thus yielding a smaller EOp and a fairer model. In the experimental section, we denote the re-weighting with $-\phi(\EOp)$ as $\phi(\EOp)$ for simplicity and the same for $\phi_i(\EOds)$.

\section{Experiments}\label{sec:experiments}
\noindent In this section, we present the experiments performed to evaluate \texttt{FairShap}. We report results on a variety of benchmark datasets of different nature for $A=Y$ and $A\neq Y$, and with fair and biased reference datasets $\T$.

\begin{figure}[ht]
\centering
\subfloat[]{
\includegraphics[width=0.45\textwidth]{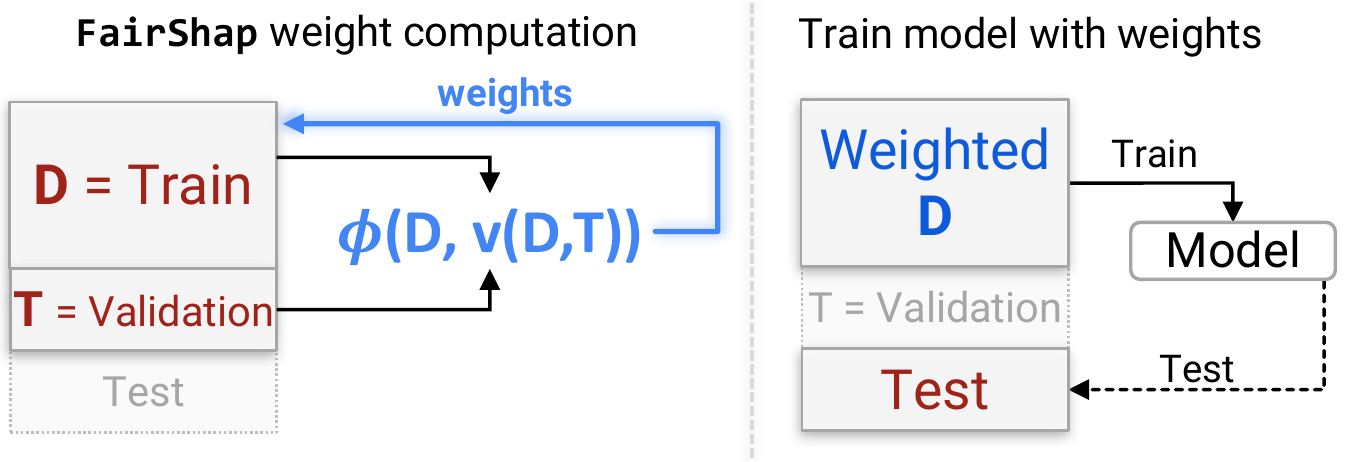}\label{fig:svAneqYpipeline}
}

\subfloat[]{
\includegraphics[width=0.49\textwidth, trim=0mm 0mm 2.2mm 0mm, clip]{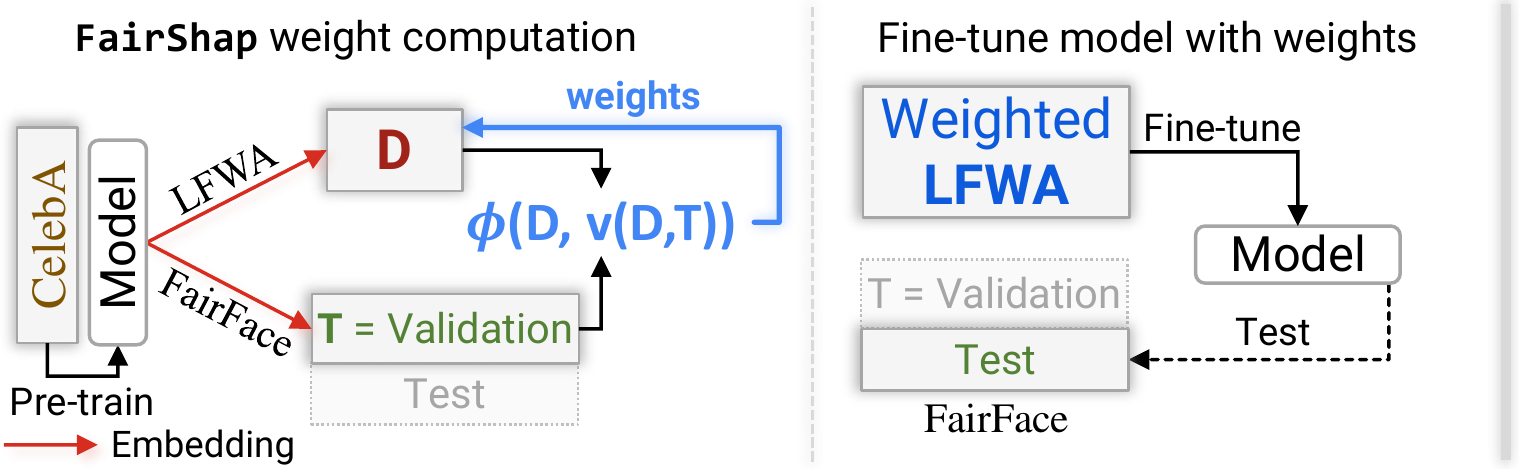}\label{fig:aeqypipeline}
}
\caption{Pipelines of the experiments described in \cref{sec:expAneqY} (a) and \cref{sec:expAeqY} (b). (a) Tabular data. $A\neq Y$ and $\T$ is biased. (b) Image experiment. $A=Y$ and $\T$ is fair.}
\label{fig:spipelines}
\end{figure}

\subsection{Instance-level data re-weighing on tabular datasets} \label{sec:expAneqY}

\noindent In this section, we consider a common real-life scenario on benchmark tabular datasets, where the models predict a target variable $Y\neq A$. Also, a single biased dataset $\D$ is used for training, validation, and testing. Thus, the validation ($\T$) and test sets are obtained as a partition of $\D$ according to the pipeline illustrated in \cref{fig:svAneqYpipeline}.

\textbf{Datasets.} We test \texttt{FairShap} on three commonly used datasets in the algorithmic fairness literature: 
(1) the German Credit \citep{kamiran2009german} dataset (German) with a binary target variable corresponding to an individual's good or bad \emph{credit risk}, and protected attributes \emph{age} and \emph{sex}; 
(2) the Adult Income dataset \citep{kohavi1996adult} (Adult) where the task is to predict whether the \emph{income} of a person is more than 50k per year, and \emph{sex} and \emph{race} are the protected attributes; and 
(3) the COMPAS \citep{angwin2016machine} dataset with binary target variable \emph{recidivism} and protected attributes \emph{sex} and \emph{race}. Appendix D.8 %
summarize the statistics of each of these datasets.

\textbf{Pipeline.} The model in all experiments is a Gradient Boosting Classifier (GBC)~\citep{friedman2001greedy}, known for its competitive performance on tabular data and interpretability properties. The pipeline in this set of experiments is depicted in Figure \ref{fig:svAneqYpipeline}. As seen in the Figure, the reference dataset $\T$ is the validation set of $\D$. 
The reported results correspond to the average values of running the experiment 50 times with random splits stratified by sensitive group and label: 70\% of the original dataset used for training ($\D$), 15\% for the reference set ($\T$) and 15\% for the test set. 
Train, reference and test set are stratified by $A$ and $Y$ such that they have the same percentage of $A$ - $Y$ samples as in the original dataset.

\textbf{FairShap re-weighting.} Given that $A\neq Y$, \texttt{FairShap} considers two different group fairness-based valuation functions: $\bm{\phi(\EOp)}$ and $\bm{\phi(\EOds)}$ as per \cref{eq:sveop} and \cref{eq:sveodds}, respectively. Note that in this case the weights assigned to each data point, $w_i$, are obtained by normalizing $\bm{\phi(\EOp)}$ and $\bm{\phi(\EOds)}$ following a methodology similar to that described in \citep{chai2022adaptative}:  
$w_i=\frac{\phi_i}{\sum_i \phi'_i} |D|$ where $\phi'_i=\frac{\phi_i-\min(\bm{\phi})}{\max(\bm{\phi})-\min(\bm{\phi})}
$.

\addtolength{\tabcolsep}{-0.pt} %
\begin{table*}[!t]
\caption{Performance of GBC without and with data re-weighting on benchmark datasets with different sensitive attributes. The \best{first} and \second{second} are color-coded. Statistically significant differences with the best performing model are denoted by $^\ddagger$ for $p<0.01$ and  $^\dagger$ for $p<0.05$.}
\label{tab:fairnessresult}
\begin{center}
\begin{small}
\begin{tabular}{lrrrrrrrr}
\toprule
&  \multicolumn{4}{c}{{\footnotesize Sex}} & \multicolumn{4}{c}{\footnotesize Age}\\
\cmidrule(lr){2-5}\cmidrule(lr){6-9}
\footnotesize{German} & Accuracy $\uparrow$ & M-F1 $\uparrow$ & EOp $\downarrow$ & EOdds $\downarrow$ & Accuracy $\uparrow$ & M-F1 $\uparrow$ & EOp $\downarrow$ & EOdds $\downarrow$ \\
\midrule
GBC & \second{.697}\tiny{$\pm$.006} &    \second{.519}\tiny{$\pm$.010} &  $^\ddagger$.107\tiny{$\pm$.020} &          $^\ddagger$.185\tiny{$\pm$.020} &
$^\dagger$\second{.704}\tiny{$\pm$.005} &    \best{.524}\tiny{$\pm$.010} &  $^\ddagger$.224\tiny{$\pm$.032} &  $^\ddagger$.345\tiny{$\pm$.030} \\
Group RW  & .695\tiny{$\pm$.006} & .514\tiny{$\pm$.010} &  $^\ddagger$.062\tiny{$\pm$.019} &          $^\ddagger$.123\tiny{$\pm$.025} &
$^\ddagger$.684\tiny{$\pm$.004} &  $^\ddagger$.396\tiny{$\pm$.041} &  $^\ddagger$.040\tiny{$\pm$.025} &  $^\ddagger$.029\tiny{$\pm$.026} \\
Postpro            &  $^\ddagger$.691\tiny{$\pm$.005} &  $^\ddagger$.366\tiny{$\pm$.055} &    \second{.013}\tiny{$\pm$.014} &  $^\dagger$\second{.036}\tiny{$\pm$.015} &
$^\ddagger$.686\tiny{$\pm$.005} &  $^\ddagger$.255\tiny{$\pm$.063} &  $^\ddagger$.044\tiny{$\pm$.022} &  $^\ddagger$.047\tiny{$\pm$.019} \\
LabelBias          &             .695\tiny{$\pm$.006} &  $^\ddagger$.465\tiny{$\pm$.035} &  $^\ddagger$.051\tiny{$\pm$.019} &          $^\ddagger$.092\tiny{$\pm$.026} &
$^\ddagger$.690\tiny{$\pm$.004} &  $^\ddagger$.354\tiny{$\pm$.053} &  $^\ddagger$.052\tiny{$\pm$.029} &  $^\ddagger$.052\tiny{$\pm$.035} \\
OptPrep            &             .694\tiny{$\pm$.006} &    \best{.521}\tiny{$\pm$.010} &  $^\ddagger$.104\tiny{$\pm$.022} &          $^\ddagger$.174\tiny{$\pm$.021} & $^\ddagger$.693\tiny{$\pm$.007} &  $^\ddagger$.487\tiny{$\pm$.030} &  $^\ddagger$.130\tiny{$\pm$.031} &  $^\ddagger$.204\tiny{$\pm$.039} \\
IF                 &    \second{.697}\tiny{$\pm$.006} &    \second{.519}\tiny{$\pm$.010} &  $^\ddagger$.107\tiny{$\pm$.020} &          $^\ddagger$.185\tiny{$\pm$.020} &
$^\dagger$\second{.704}\tiny{$\pm$.005} &    \best{.524}\tiny{$\pm$.010} &  $^\ddagger$.224\tiny{$\pm$.032} &  $^\ddagger$.345\tiny{$\pm$.030} \\
$\phi(\Acc)$       &    \best{.700}\tiny{$\pm$.005} &   $^\dagger$.507\tiny{$\pm$.009} &  $^\ddagger$.097\tiny{$\pm$.018} &          $^\ddagger$.184\tiny{$\pm$.018} &
\best{.706}\tiny{$\pm$.005} &  $^\ddagger$.517\tiny{$\pm$.010} &  $^\ddagger$.193\tiny{$\pm$.025} &  $^\ddagger$.313\tiny{$\pm$.025} \\
$\bm{\phi(\EOp)}$  &  $^\ddagger$.683\tiny{$\pm$.006} &  $^\ddagger$.460\tiny{$\pm$.034} &             .029\tiny{$\pm$.026} &           $^\dagger$.049\tiny{$\pm$.033} &
$^\ddagger$.685\tiny{$\pm$.004} &  $^\ddagger$.373\tiny{$\pm$.046} &    \second{.024}\tiny{$\pm$.023} &    \second{.007}\tiny{$\pm$.021} \\
$\bm{\phi(\EOds)}$ &  $^\ddagger$.686\tiny{$\pm$.006} &  $^\ddagger$.477\tiny{$\pm$.009} &    \best{.002}\tiny{$\pm$.025} &            \best{.002}\tiny{$\pm$.031} 
& $^\ddagger$.681\tiny{$\pm$.005} &  $^\ddagger$.353\tiny{$\pm$.049} &    \best{.019}\tiny{$\pm$.020} &    \best{.003}\tiny{$\pm$.013}\\
\bottomrule
&  \multicolumn{4}{c}{{\footnotesize Sex}} & \multicolumn{4}{c}{\footnotesize Race}\\
\cmidrule(lr){2-5}\cmidrule(lr){6-9}
\footnotesize{Adult} & Accuracy $\uparrow$ & M-F1 $\uparrow$ & EOp $\downarrow$ & EOdds $\downarrow$ & Accuracy $\uparrow$ & M-F1 $\uparrow$ & EOp $\downarrow$ & EOdds $\downarrow$ \\
\midrule
GBC                &    \second{.803}\tiny{$\pm$.001} &  $^\ddagger$.680\tiny{$\pm$.002} &  $^\ddagger$.451\tiny{$\pm$.004} &  $^\ddagger$.278\tiny{$\pm$.003} & 
\best{.803}\tiny{$\pm$.001} &   $^\ddagger$.682\tiny{$\pm$.002} &  $^\ddagger$.164\tiny{$\pm$.010} &  $^\ddagger$.106\tiny{$\pm$.006} \\
Group RW            &  $^\ddagger$.790\tiny{$\pm$.001} &    \best{.684}\tiny{$\pm$.002} &    \second{.002}\tiny{$\pm$.009} &    \second{.001}\tiny{$\pm$.005} & 
\best{.803}\tiny{$\pm$.001} &   $^\ddagger$.683\tiny{$\pm$.002} &             .010\tiny{$\pm$.009} &             .010\tiny{$\pm$.005} \\
Postpro            &  $^\ddagger$.791\tiny{$\pm$.001} &    $^\dagger$.679\tiny{$\pm$.004} &  $^\ddagger$.056\tiny{$\pm$.013} &  $^\ddagger$.034\tiny{$\pm$.007} & 
.802\tiny{$\pm$.001} &    \best{.688}\tiny{$\pm$.002}  &  $^\ddagger$.061\tiny{$\pm$.011} &  $^\ddagger$.042\tiny{$\pm$.006} \\
LabelBias          &  $^\ddagger$.781\tiny{$\pm$.001} &     $^\ddagger$.681\tiny{$\pm$.002} &  $^\ddagger$.065\tiny{$\pm$.011} &  $^\ddagger$.049\tiny{$\pm$.006} & 
$^\ddagger$.800\tiny{$\pm$.001} &    \second{.686}\tiny{$\pm$.002}  &  $^\ddagger$.118\tiny{$\pm$.013} &  $^\ddagger$.074\tiny{$\pm$.007}\\
OptPrep            &  $^\ddagger$.789\tiny{$\pm$.001} &    $^\ddagger$.676\tiny{$\pm$.004} &  $^\ddagger$.064\tiny{$\pm$.029} &  $^\ddagger$.037\tiny{$\pm$.017} & 
$^\ddagger$.800\tiny{$\pm$.001} &   $^\dagger$.685\tiny{$\pm$.002}  &  $^\ddagger$.044\tiny{$\pm$.015} &  $^\ddagger$.029\tiny{$\pm$.009} \\
IF                 &  $^\ddagger$.787\tiny{$\pm$.002} &     $^\dagger$.681\tiny{$\pm$.003} &  $^\ddagger$.159\tiny{$\pm$.037} &  $^\ddagger$.092\tiny{$\pm$.022} & 
$^\ddagger$.797\tiny{$\pm$.002} &   $^\dagger$.685\tiny{$\pm$.002}  &  $^\ddagger$.042\tiny{$\pm$.020} &  $^\ddagger$.031\tiny{$\pm$.012} \\
$\phi(\Acc)$       &    \best{.804}\tiny{$\pm$.001} &   $^\ddagger$.681\tiny{$\pm$.002} &  $^\ddagger$.452\tiny{$\pm$.005} &  $^\ddagger$.279\tiny{$\pm$.003} & \best{.803}\tiny{$\pm$.001} &   $^\ddagger$.681\tiny{$\pm$.002} &  $^\ddagger$.161\tiny{$\pm$.011} &  $^\ddagger$.104\tiny{$\pm$.007} \\
$\bm{\phi(\EOp)}$  &  $^\ddagger$.790\tiny{$\pm$.001} &      \best{.684}\tiny{$\pm$.002} &    \second{.002}\tiny{$\pm$.009} &    \best{3e-4}\tiny{$\pm$.005} & 
.802\tiny{$\pm$.001} &   $^\ddagger$.683\tiny{$\pm$.002} &    \second{.009}\tiny{$\pm$.010} &    \second{.009}\tiny{$\pm$.005} \\
$\bm{\phi(\EOds)}$ &  $^\ddagger$.790\tiny{$\pm$.001} &               .683\tiny{$\pm$.002} &    \best{8e-4}\tiny{$\pm$.009} &    \second{.001}\tiny{$\pm$.005} & 
.802\tiny{$\pm$.001} &   $^\ddagger$.683\tiny{$\pm$.002} &    \best{.007}\tiny{$\pm$.009} &    \best{.007}\tiny{$\pm$.005} \\
\bottomrule
&  \multicolumn{4}{c}{{\footnotesize Sex}} & \multicolumn{4}{c}{\footnotesize Race}\\
\cmidrule(lr){2-5}\cmidrule(lr){6-9}
\footnotesize{COMPAS} & Accuracy $\uparrow$ & M-F1 $\uparrow$ & EOp $\downarrow$ & EOdds $\downarrow$ & Accuracy $\uparrow$ & M-F1 $\uparrow$ & EOp $\downarrow$ & EOdds $\downarrow$ \\
\midrule
GBC                &    \second{.666}\tiny{$\pm$.004} &    \best{.662}\tiny{$\pm$.004} &  $^\ddagger$.158\tiny{$\pm$.014} &  $^\ddagger$.199\tiny{$\pm$.014} &
\best{.663}\tiny{$\pm$.004} &    \best{.658}\tiny{$\pm$.004} &           $^\ddagger$.184\tiny{$\pm$.013} &  $^\ddagger$.218\tiny{$\pm$.013}\\
Group RW            &             .664\tiny{$\pm$.004} &             .660\tiny{$\pm$.004} &             .020\tiny{$\pm$.016} &  $^\ddagger$.038\tiny{$\pm$.014} &
$^\ddagger$.649\tiny{$\pm$.004} &  $^\ddagger$.646\tiny{$\pm$.004} &           $^\ddagger$.028\tiny{$\pm$.015} &             .007\tiny{$\pm$.016} \\
Postpro            &  $^\ddagger$.660\tiny{$\pm$.003} &  $^\ddagger$.655\tiny{$\pm$.003} &             .017\tiny{$\pm$.017} &   $^\dagger$.030\tiny{$\pm$.015} &
$^\ddagger$.647\tiny{$\pm$.005} &  $^\ddagger$.642\tiny{$\pm$.005} &             \best{1e-4}\tiny{$\pm$.015} &  $^\ddagger$.026\tiny{$\pm$.016} \\
LabelBias          &  $^\ddagger$.639\tiny{$\pm$.005} &  $^\ddagger$.612\tiny{$\pm$.007} &    \best{.006}\tiny{$\pm$.013} &    \second{.010}\tiny{$\pm$.014} &
$^\ddagger$.645\tiny{$\pm$.004} &  $^\ddagger$.627\tiny{$\pm$.005} &           $^\ddagger$.030\tiny{$\pm$.011} &  $^\ddagger$.045\tiny{$\pm$.014} \\
OptPrep            &             .664\tiny{$\pm$.003} &             .660\tiny{$\pm$.003} &  $^\ddagger$.045\tiny{$\pm$.020} &  $^\ddagger$.065\tiny{$\pm$.019} &
$^\ddagger$.655\tiny{$\pm$.004} &   $^\dagger$.651\tiny{$\pm$.004} &           $^\ddagger$.044\tiny{$\pm$.020} &  $^\ddagger$.078\tiny{$\pm$.020}\\
IF                 &             .663\tiny{$\pm$.003} &             .658\tiny{$\pm$.003} &  $^\ddagger$.129\tiny{$\pm$.016} &  $^\ddagger$.161\tiny{$\pm$.015} &
.660\tiny{$\pm$.004} &             .655\tiny{$\pm$.004} &           $^\ddagger$.165\tiny{$\pm$.017} &  $^\ddagger$.198\tiny{$\pm$.015}\\
$\phi(\Acc)$       &    \best{.667}\tiny{$\pm$.004} &    \best{.662}\tiny{$\pm$.004} &  $^\ddagger$.156\tiny{$\pm$.014} &  $^\ddagger$.198\tiny{$\pm$.013} &
\best{.663}\tiny{$\pm$.004} &    \second{.657}\tiny{$\pm$.004} &           $^\ddagger$.184\tiny{$\pm$.013} &  $^\ddagger$.218\tiny{$\pm$.013} \\
$\bm{\phi(\EOp)}$  &   $^\dagger$.661\tiny{$\pm$.003} &   $^\dagger$.658\tiny{$\pm$.004} &    \second{.013}\tiny{$\pm$.024} &    \best{.007}\tiny{$\pm$.021} &
$^\ddagger$.650\tiny{$\pm$.004} &  $^\ddagger$.647\tiny{$\pm$.004} &  $^\ddagger$\second{.027}\tiny{$\pm$.016} &    \best{.004}\tiny{$\pm$.017} \\
$\bm{\phi(\EOds)}$ &             .663\tiny{$\pm$.004} &             .659\tiny{$\pm$.004} &             .019\tiny{$\pm$.021} &   $^\dagger$.036\tiny{$\pm$.020} &
$^\ddagger$.648\tiny{$\pm$.004} &  $^\ddagger$.646\tiny{$\pm$.004} &           $^\ddagger$.036\tiny{$\pm$.017} &    \best{.004}\tiny{$\pm$.018} \\
\bottomrule
\end{tabular}

\end{small}
\end{center}
\end{table*}
\addtolength{\tabcolsep}{0.pt}

\begin{figure*}[ht]
\centering
\subfloat[]{\includegraphics[width=0.3\textwidth]{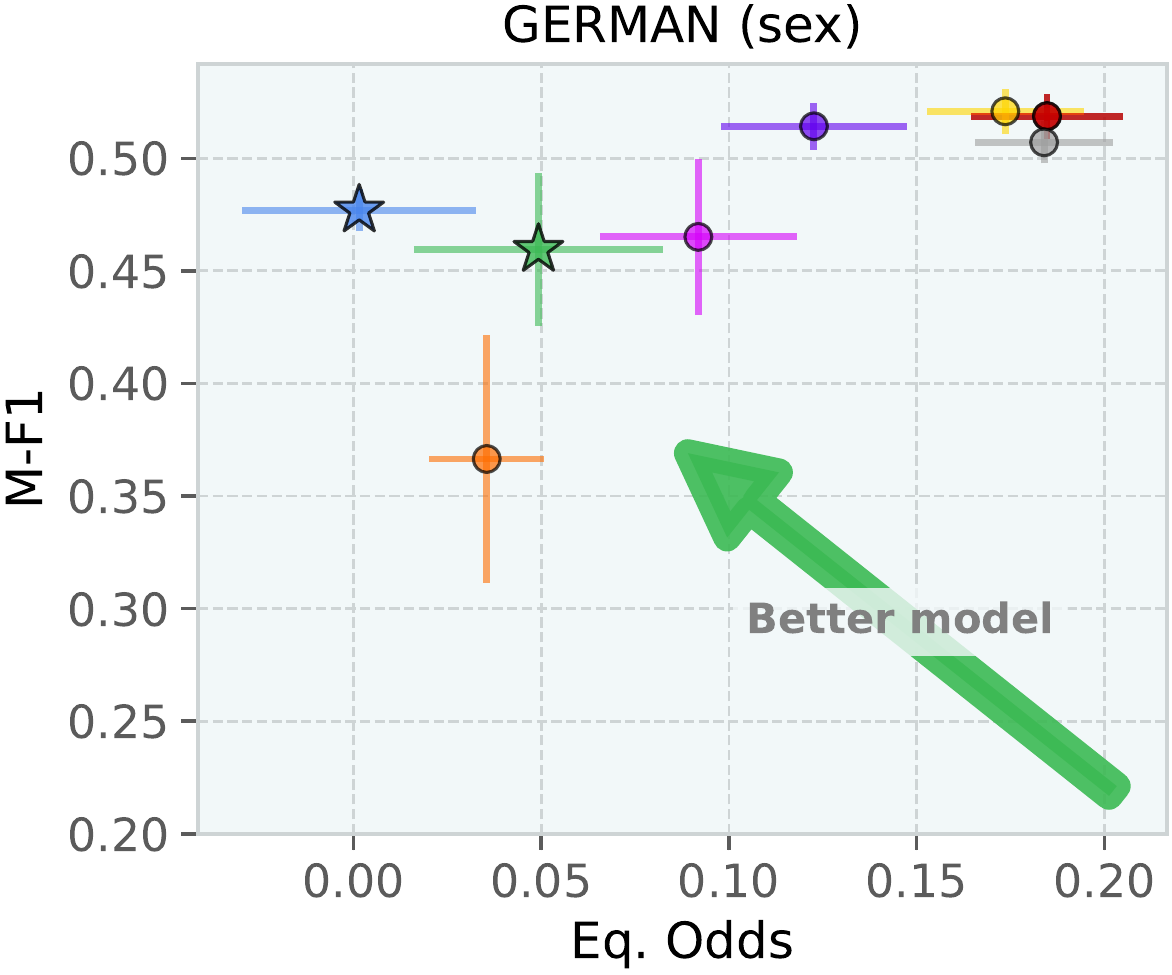}
}
\hfill
\subfloat[]{\includegraphics[width=0.29\textwidth]{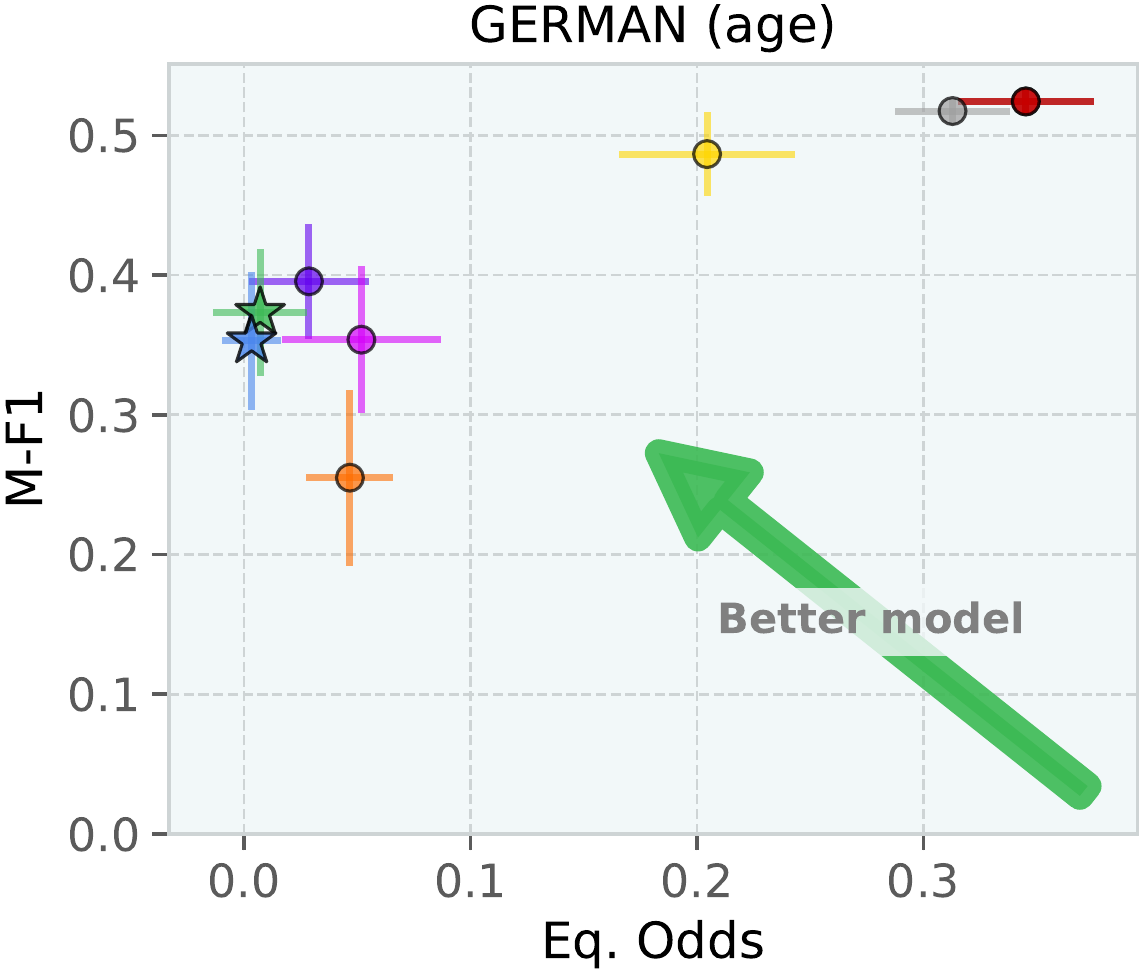}
}
\hfill
\subfloat[]{\includegraphics[width=0.3\textwidth]{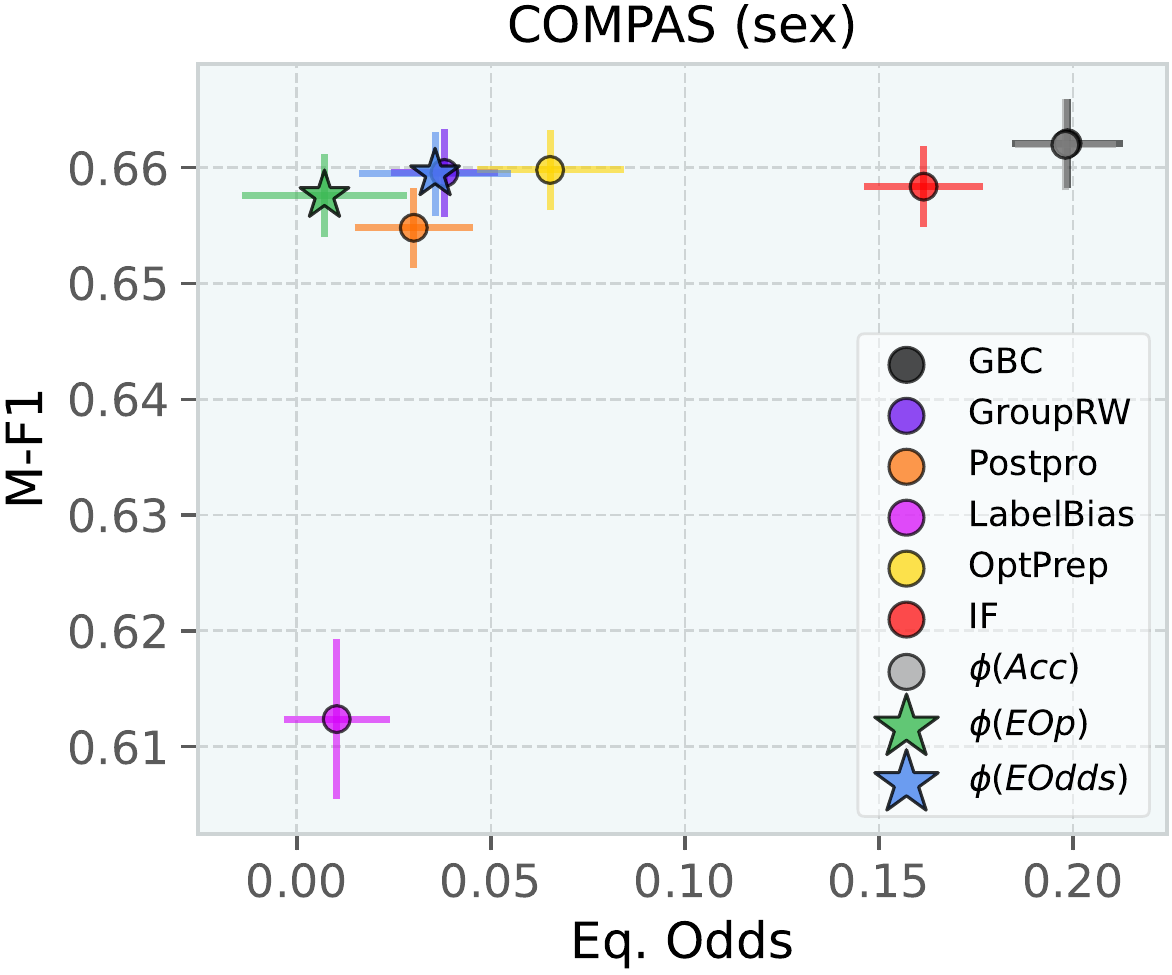}
}
\caption{Accuracy (M-F1) vs fairness analysis. The models trained with data re-weighting via \texttt{FairShap} (depicted as stars in the graphs) improve in fairness while maintaining competitive levels of accuracy when compared to the baselines.}
\label{fig:pareto}
\end{figure*}

\textbf{Baselines.}
To the best of our knowledge, \texttt{FairShap} is the only interpretable, instance-level model-agnostic data re-weighting and data valuation based approach for group algorithmic fairness (see Table 4 in the Appendix). %
Thus, we compare its performance with 6 state-of-the-art algorithmic fairness approaches that only \emph{partially} satisfy \texttt{FairShap}'s properties:

\emph{1. Group RW}: A group-based re-weighting method that assigns the same weights to all samples from the same category or group according to the protected attribute~\citep{kamiran2012data}. Thus, this is not an instance-level re-weighting approach; 

\emph{2. Post-pro:} A post-processing algorithmic fairness method that does not fulfill any of the desiderata, but it is broadly used in the community \citep{hardt2016equality}; 

\emph{3. LabelBias:} A model that learns the weights in a in-processing manner and therefore it is neither a pre-processing nor a model-agnostic approach~\citep{jiang2020identifying}; 

\emph{4. Opt-Pre:} A model-agnostic pre-processing approach for algorithmic fairness based on feature and label transformations which does not assign any weights to the data~\citep{calmon2017optimized}; 

\emph{5. IFs:} An Influence Function (IF)-based approach whis is an in-processing re-training approach, since the weights are computed from the Hessian of a pretrained model. We use the same hyper-parameters reported by the authors for each of the datasets \citep{li2022individualreweighing}; and

\emph{6. $\bm{\phi(\Acc)}$}: A method based on data re-weighting by means of an accuracy-based valuation function without any fairness considerations~\citep{ghorbani2019data}. The weights assigned to each data point are obtained according to the same normalization as that used in \texttt{FairShap}'s case. 

An extended explanation of the methods and the hyperparameters used in the experiments can be found in Appendix~D.2.%

\textbf{Experimental setup.}
We adopt the experimental setup that is commonly followed in the ML community: the weights, the influence functions and the thresholds required by the different methods are computed on the validation set. Furthermore, all the reported results correspond to the mean values of running 50 experiments on each dataset with random stratified train, validation and test set splits in each experiment, as previously described. Note that some previous works in the algorithmic fairness literature do not perform label-group stratification on the splits, or compute the weights or thresholds using the test set instead of the validation set. Hence, reported performances are not directly comparable to ours.

\textbf{Results.} The metrics used for evaluation are accuracy (Acc); Macro-F1 (M-F1), which is an extension of the F1 score that addresses class imbalances, as it is the case in our benchmark datasets (see Appendix~D.7 %
for a definition of M-F1); EOp and EOdds, all of them between groups according to the sensitive attribute. \cref{tab:fairnessresult} summarizes the results, highlighting in \textbf{bold} the best-performing method. The arrows indicate if the optimal result is 0 ($\downarrow$) or 1 ($\uparrow$).

As shown in the \cref{tab:fairnessresult} and Figure \ref{fig:pareto}, data re-weighting with \texttt{FairShap} ($\bm{\phi}(\EOds)$ and $\bm{\phi}(\EOp)$) generally yields significantly better results in the fairness metrics than the baselines while keeping competitive levels of accuracy. This improvement is notable when compared to the performance of the model built without data re-weighing (GBC).
For example, in the German dataset with sex as protected attribute, the model's Equalized Odds metric is \textbf{93x} smaller (better) when re-weighting via \texttt{FairShap} ($\bm{\phi}(\EOds)$) than the baseline model (GBC) and \textbf{18x} better than the most competitive baseline (PostPro).

From the results, we draw several observations. First, the variance in the accuracy of the post-processing approach (PostPro) is significantly larger than that of other methods. Second, a simple method such as Group RW delivers very competitive results, even better than more sophisticated, recent approaches. Finally, accuracy is not an appropriate metric of the performance of the classifier due to the imbalance of the datasets.

\begin{figure}[ht]
\centering
\subfloat[]{\includegraphics[width=0.24\textwidth]
{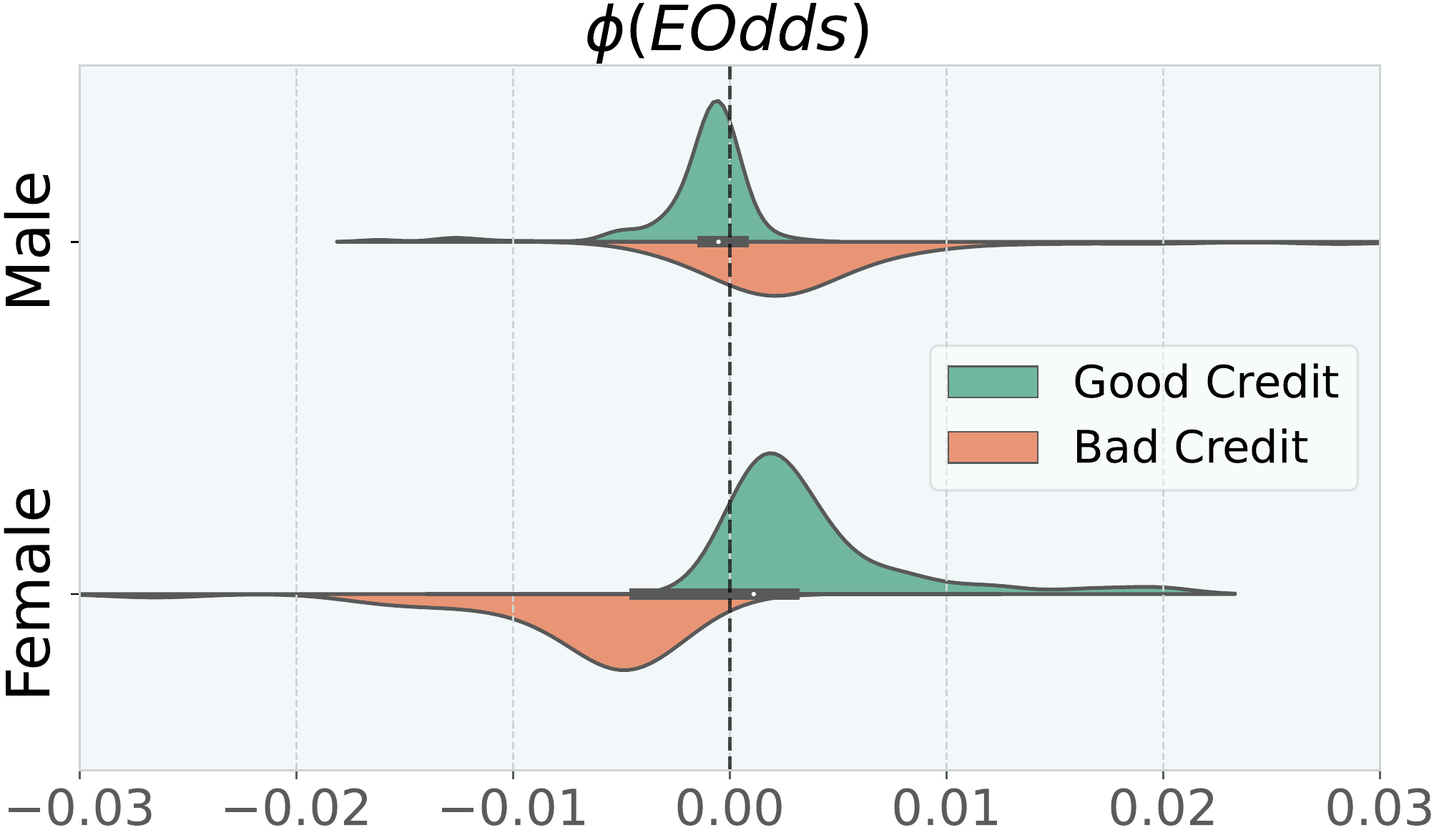}
}
\subfloat[]{\includegraphics[width=0.24\textwidth]{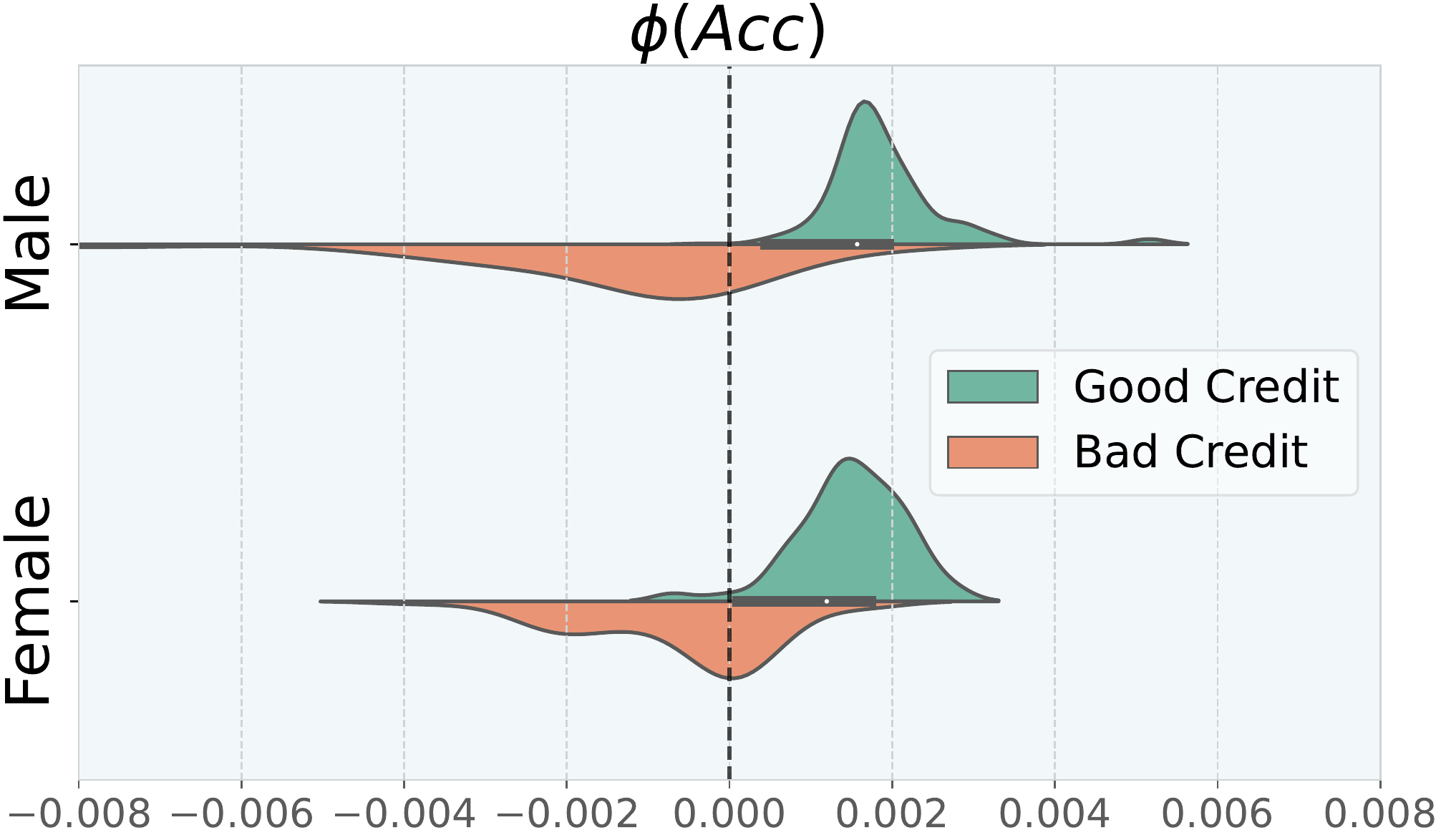}
}
\caption{(a) $\phi_i(\EOds)$ and (b) $\phi_i(\Acc)$ for the German Credit dataset with $A=\text{sex}$.}
\label{fig:histogramsgermansex}
\end{figure}

To shed further light on the behavior of \texttt{FairShap}, \cref{fig:histogramsgermansex} depicts the histograms of $\bm{\phi}(\EOds)$ and $\bm{\phi}(\Acc)$ on the German Credit dataset with sex as protected attribute. Note how the distribution of $\bm{\phi}(\Acc)$ is similar for males and females, even though the dataset is highly imbalanced: examples with good credit, irrespective of their sex, receive larger weights than those with bad credit. Conversely, the $\bm{\phi}(\EOds)$ values are larger for female applicants with good credit than to their male counterparts. In addition, $\bm{\phi}(\EOds)$ are larger for male applicants with bad credit than to their female counterparts. These distributions of $\bm{\phi}(\EOds)$ compensate for the imbalances in the raw dataset (both in terms of sex and credit risk), yielding fairer classifiers, as reflected in the results reported in \cref{tab:fairnessresult}. 

Finally, an ablation study of the impact of the size of the reference dataset $\T$ on the performance can be found in Appendix D.6. %

\subsection{Accuracy vs fairness}

\noindent  Shapley Values provide a way to assign weights to individual data points based on their contributions to a particular outcome --such as the model's prediction or fairness. While Shapley Values can be used to re-weight data points to improve fairness according to a specific fairness metric, the impact on accuracy is not guaranteed to be consistent across all datasets and scenarios. However, in many scenarios, particularly when there are a large number of examples in the majority group, optimizing TPR and FPR via data re-weighting does not necessarily lead to a decrease in accuracy for the majority group while improving the model's fairness for the disadvantaged group. We observe this behavior in all our experiments.

\begin{figure*}[t]
\centering
\subfloat[]{\includegraphics[width=.24\linewidth]{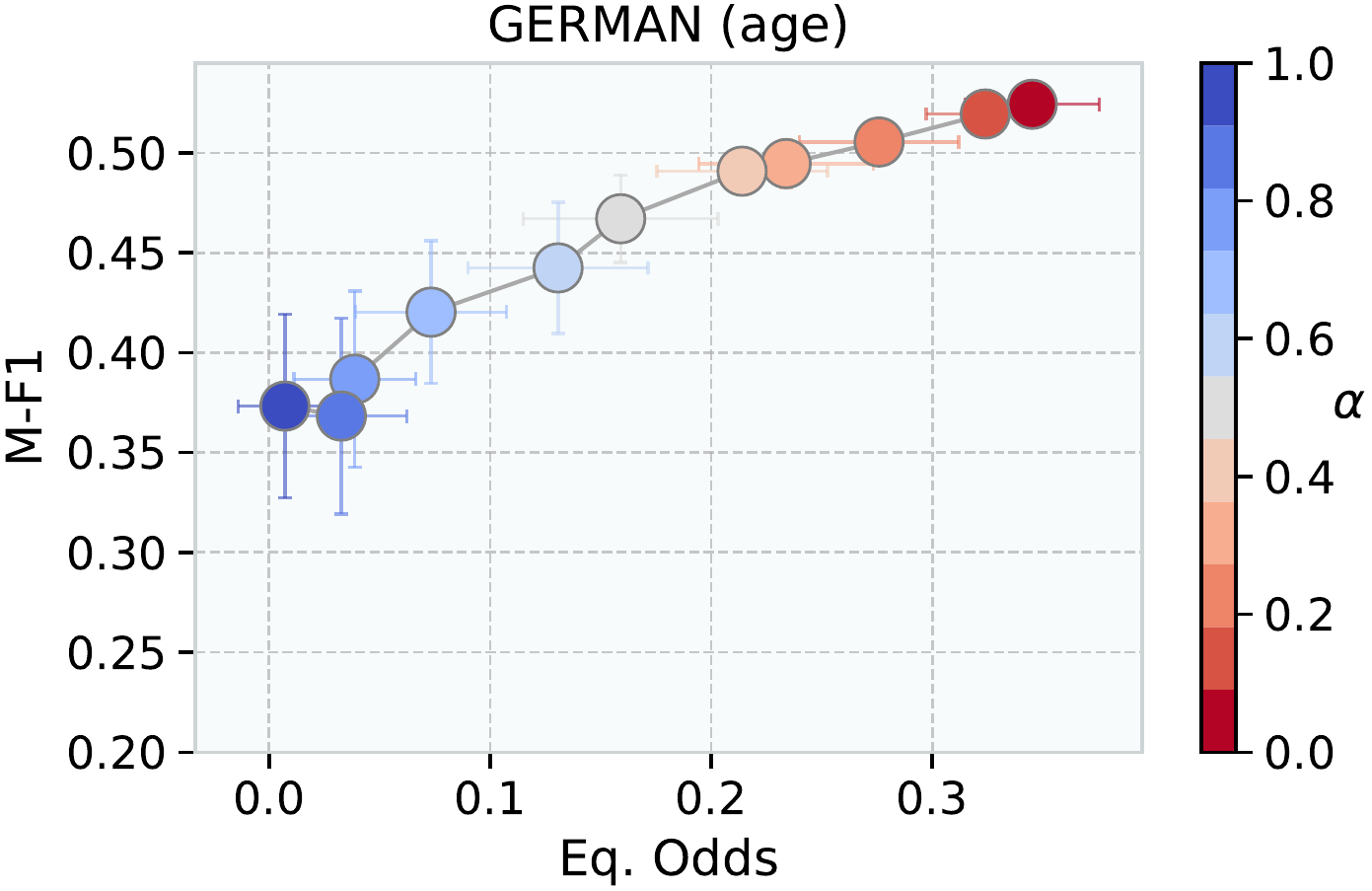}
}
\hfill
\subfloat[]{\includegraphics[width=.24\linewidth]{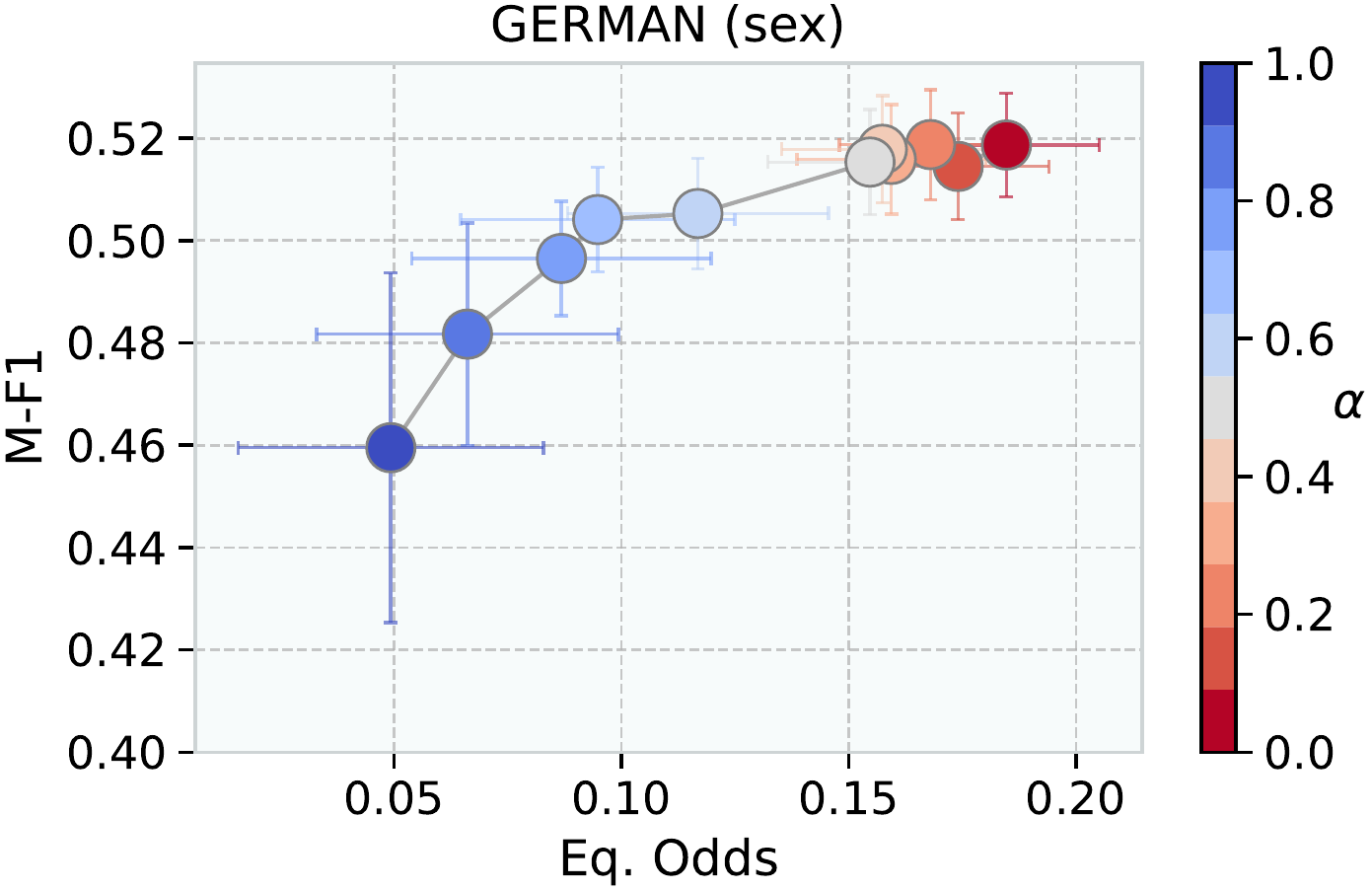}
}
\hfill
\subfloat[]{\includegraphics[width=.24\linewidth]{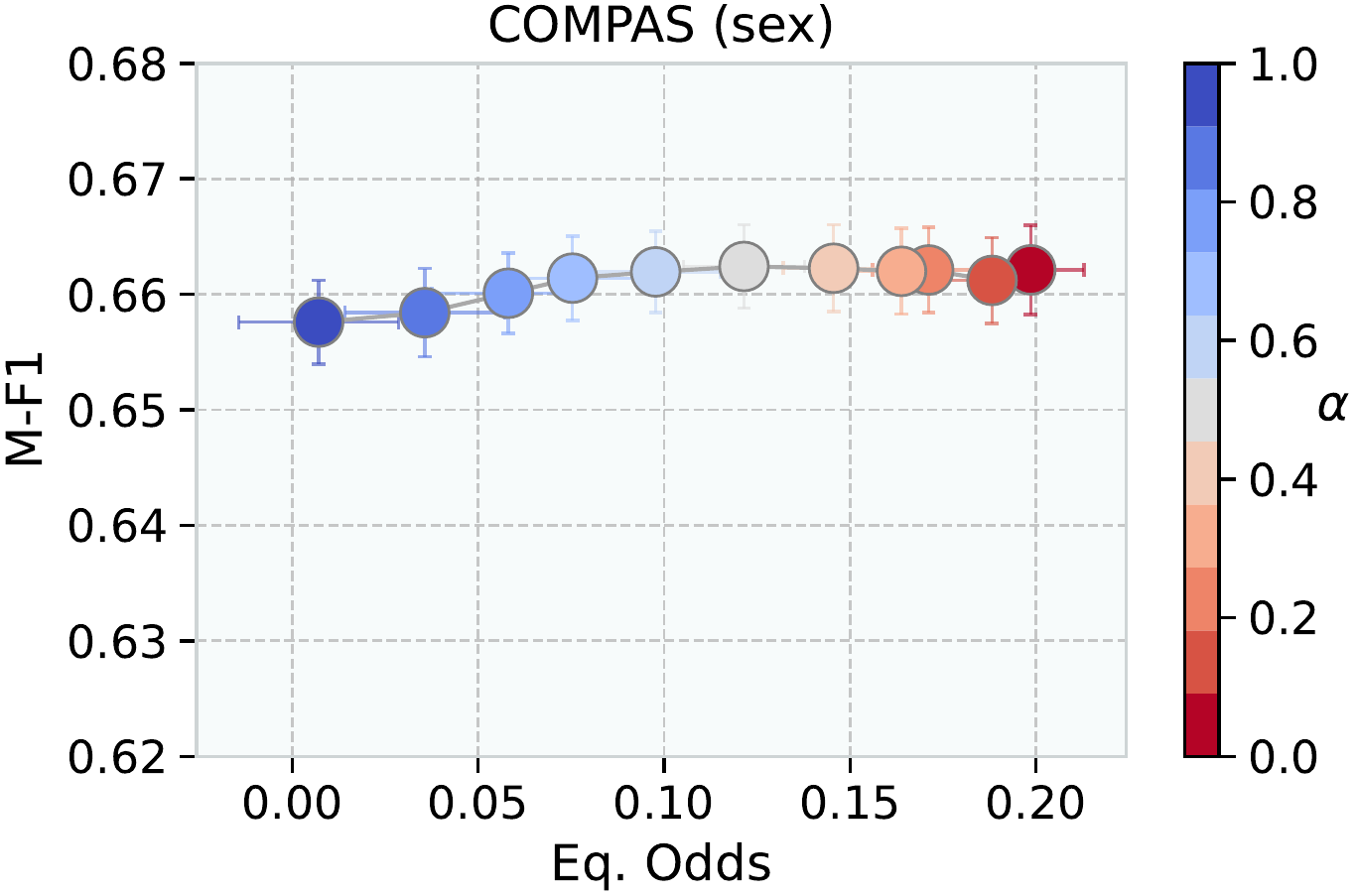}
}
\hfill
\subfloat[]{\includegraphics[width=.24\linewidth]{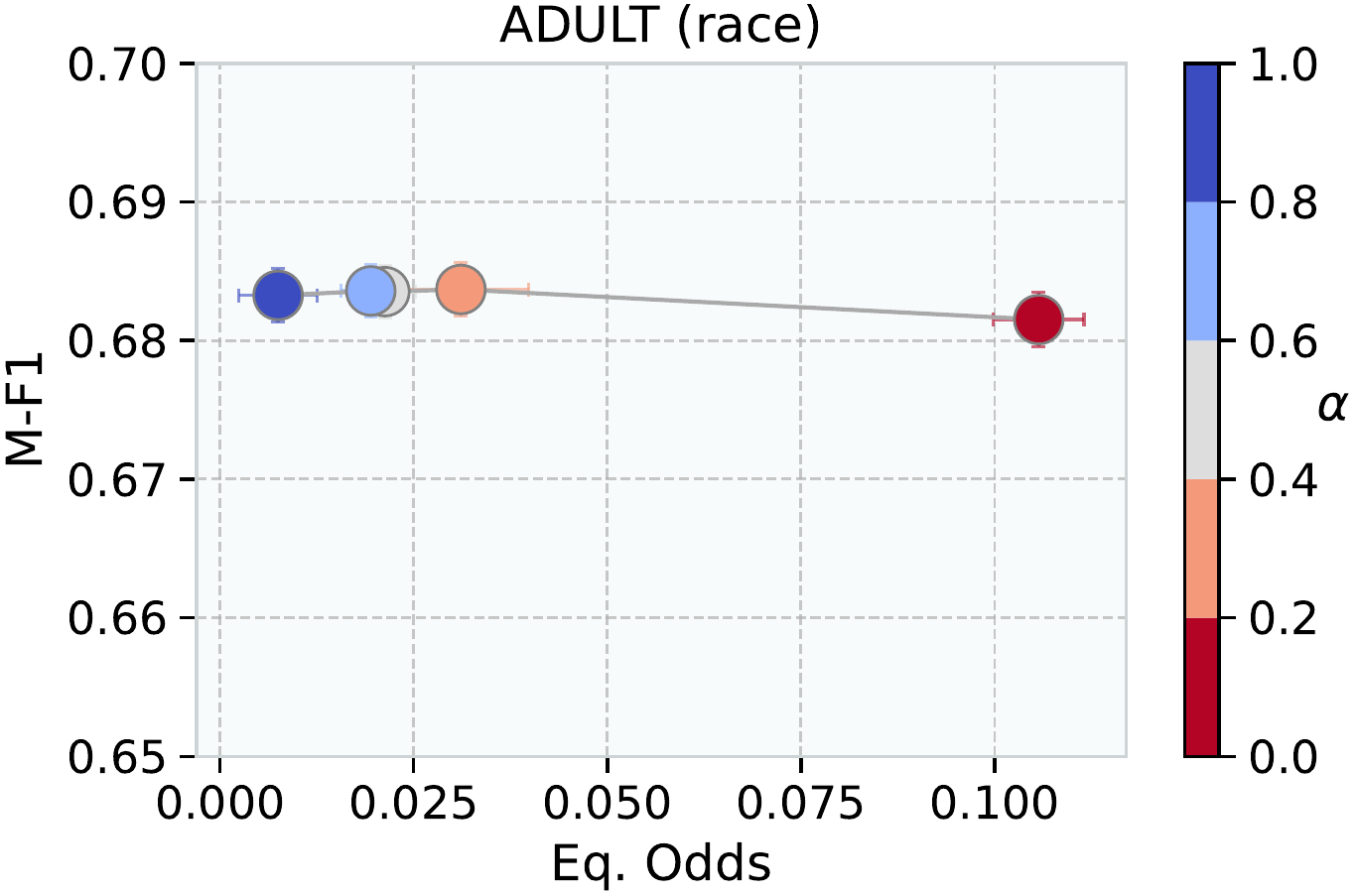}
}

\subfloat[]{\includegraphics[width=.23\linewidth]{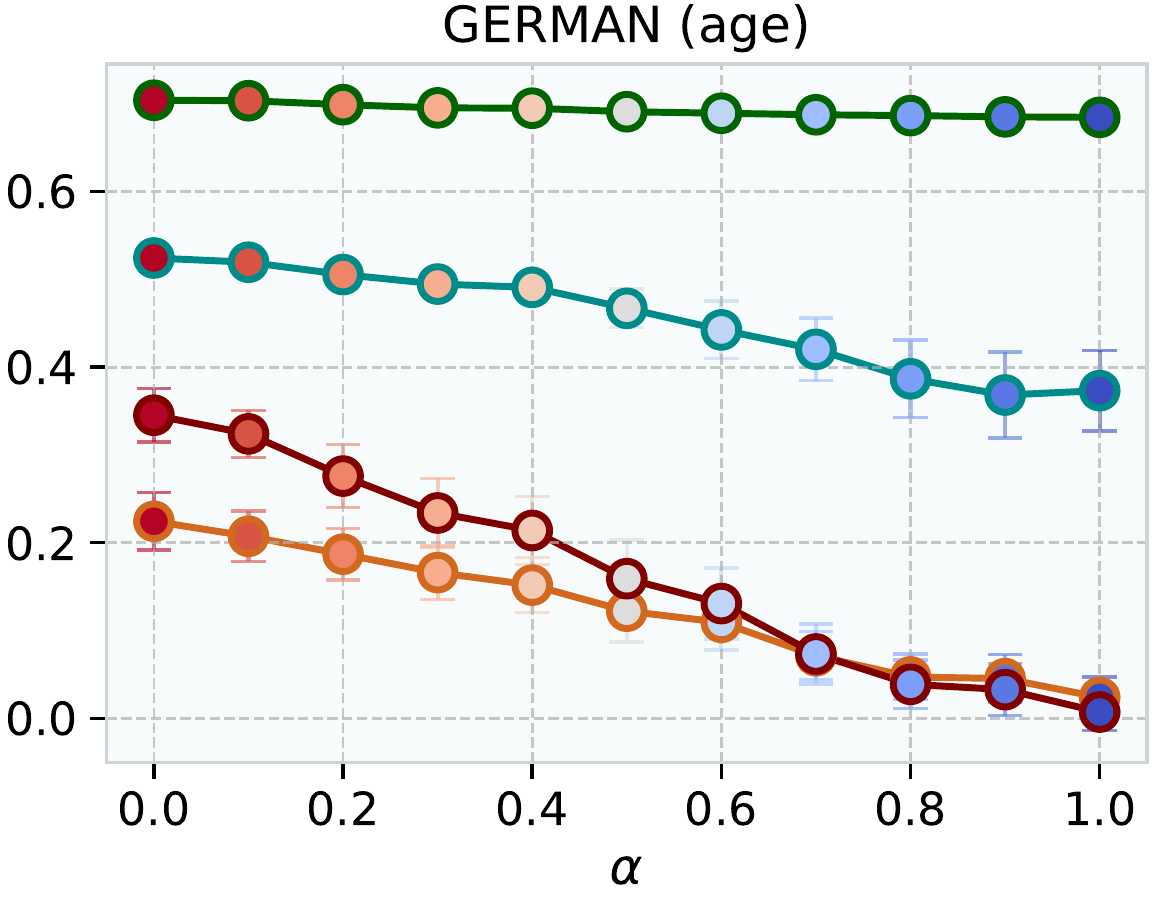}
}
\hfill
\subfloat[]{\includegraphics[width=.23\linewidth]{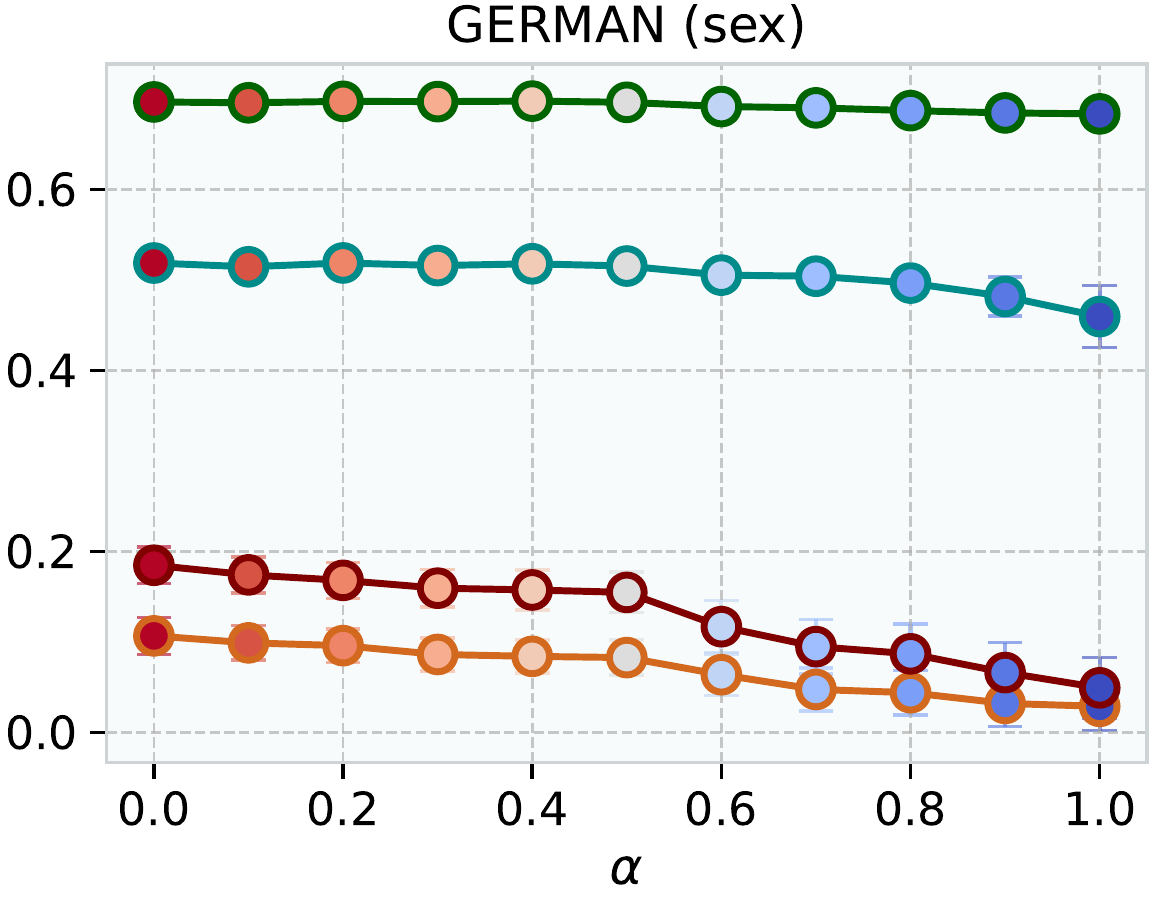}
}
\hfill
\subfloat[]{\includegraphics[width=.23\linewidth]{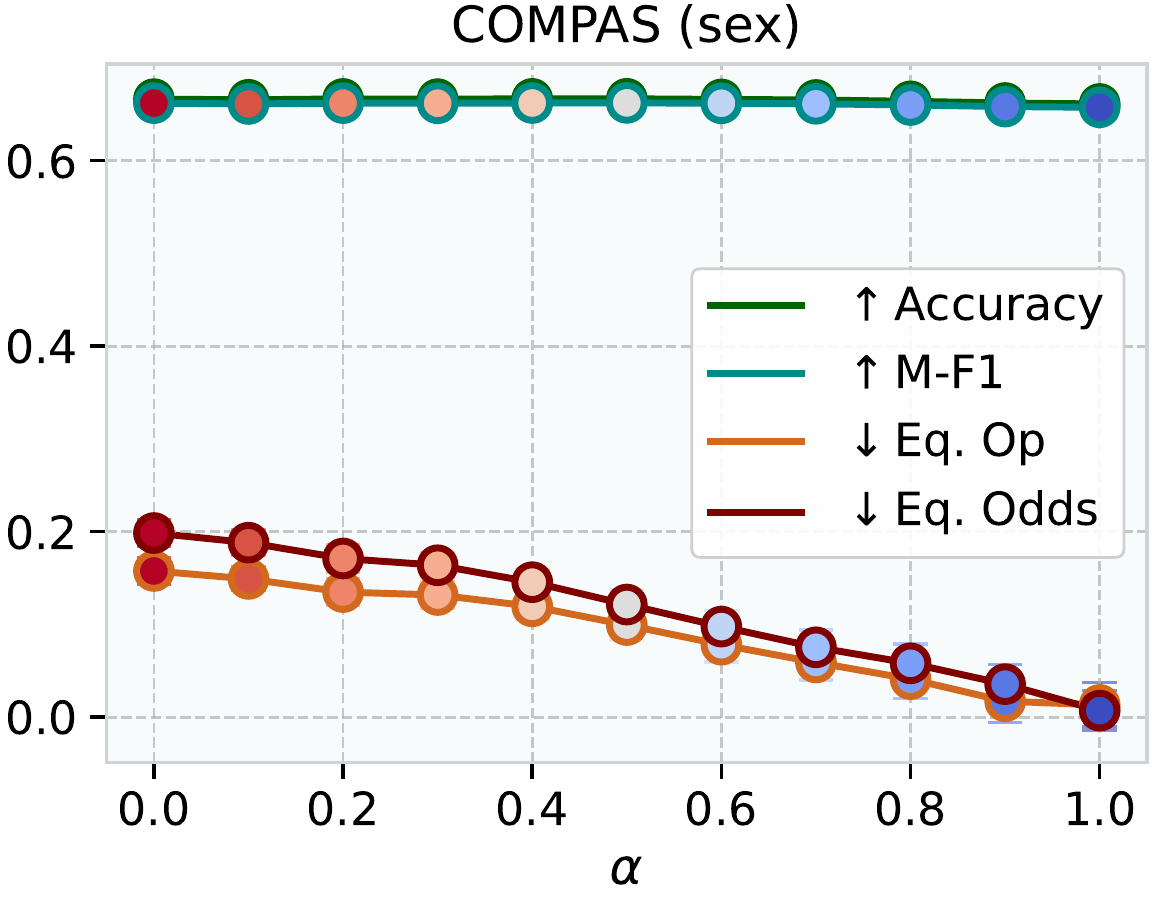}
}
\hfill
\subfloat[]{\includegraphics[width=.23\linewidth]{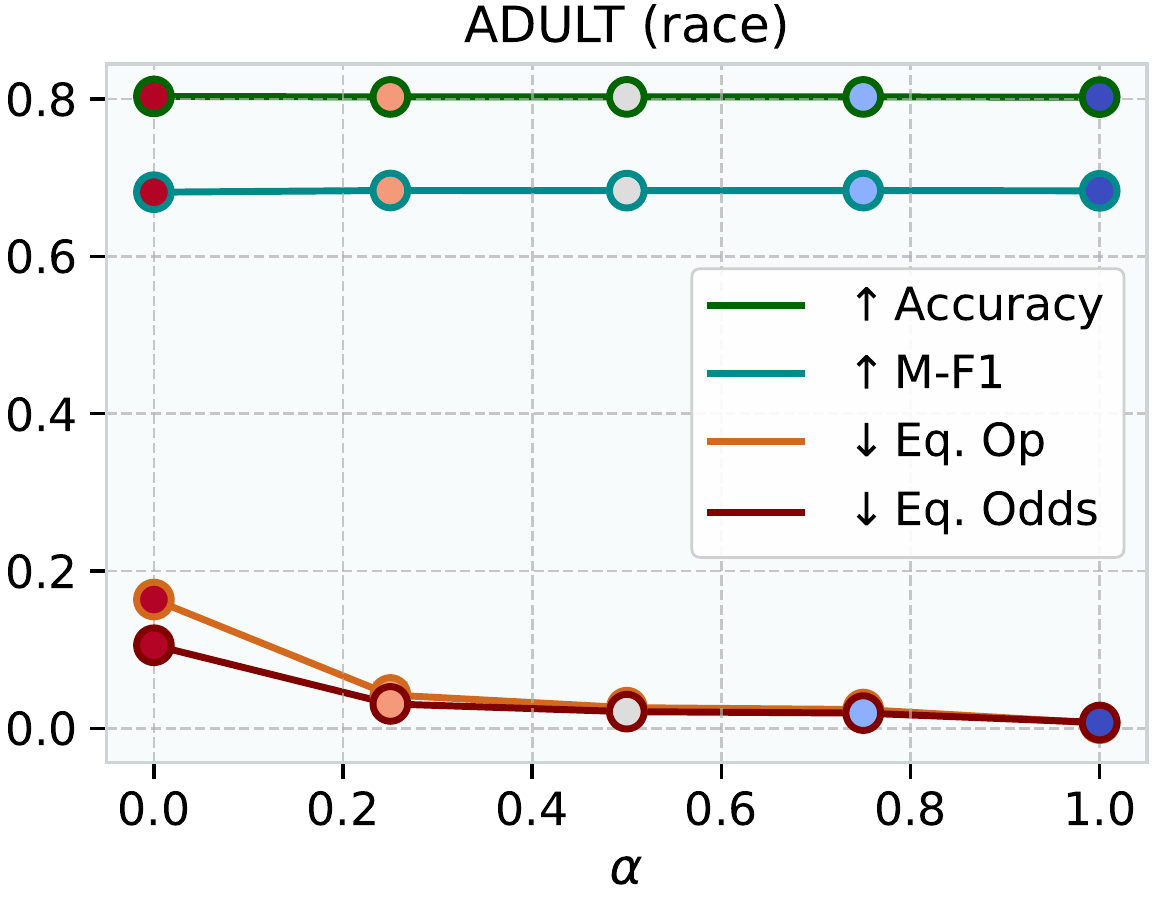}
}
\caption{Accuracy vs fairness trade-off for different values of $\alpha$, where $\alpha=0=$ means no data re-weighting and $\alpha=1$ means data re-weighting according to \texttt{FairShap}. Results show the mean and 95\% CI over 50 random iterations for three different datasets, different accuracy and fairness metrics. $\bm{\Phi(\EOp)}$ is used to re-weight the German and COMPAS datasets, and $\bm{\Phi(\EOds)}$ to re-weight in the Adult dataset. Top graphs (a-d) show the accuracy-fairness Pareto front --where accuracy is measured using M-F1 and fairness by Equalized Odds. The bottom graphs (e-h) illustrate the standard Accuracy, M-F1, Equal Opportunity and Equalized Odds for increasing values of $\alpha$. Observe how data re-weighting by means of \texttt{FairShap} ($\alpha=1$) delivers the fairest models while keeping competitive levels of accuracy on the COMPAS and ADULT datasets.}
\label{fig:tradeoff}
\end{figure*}

\textbf{Experiment.} To further illustrate the impact of \texttt{FairShap}'s data re-weighting on the model's accuracy and fairness, \cref{fig:tradeoff} depicts the utility-fairness curves on the three benchmark datasets (German, Adult and COMPAS). We define a parameter $\alpha$ that controls the contribution to the weights of each data point according to \texttt{FairShap}, ranging from $\alpha$ = 0 (no data re-weighting) to $\alpha$ = 1 (weights as given by \texttt{FairShap}). Thus, the weights of each data point $i$ are computed as $w_i' = (1-\alpha)\mathbf{1}_{|\D|}+\alpha w_i$ where $\mathbf{1}_{n}=(1,1,\dotso, 1)\in \mathbb{R}^{n}$ is the constant vector and $w_i$ are the weights according to \texttt{FairShap}.

As shown in the Figure, the larger the $\alpha$, i.e. the larger the importance of \texttt{FairShap}'s weights, the better the model's fairness. In some scenarios, such as on the German dataset, we observe a utility-fairness Pareto front where the fairest models correspond to $\alpha$ = 1 and the best performing models correspond to $\alpha$ = 0. Conversely, on the COMPAS (sex) and Adult (race) datasets, larger values of $\alpha$ significantly increase the fairness of the model while keeping similar levels of utility (M-F1 and Accuracy).

\subsection{Data pruning or selection}
\label{sec:pruning}
\noindent Data valuation functions can also be used to guide policies for data selection. In this section, we evaluate the effectiveness of \texttt{FairShap} in a data pruning task.

The goal is to identify and remove samples with low value (negative influence) on fairness metrics such that the model trained with the resulting data would maintain good levels of accuracy while being fairer than a model trained with the entire dataset. To achieve such a goal, we first we computed the value of each training point. Then we ranked the points from lowest to highest value, prioritizing those with a negative impact on fairness. We sequentially removed data points in this order, training and testing a model with the remaining points. At each step, we removed 0.5\% of the points and continued until a total of 15\% had been removed. This process was repeated for 50 random training ($\D$), validation ($\T$), and test splits for each method.

We compared five different data valuation approaches: \textit{Rand}, which randomly removes data points; the same \textit{IF}-based method used as a baseline in the previously presented experiments; $\phi(\Acc)$, and the two \texttt{FairShap} valuations: $\phi(\EOp)$ and $\phi(\EOds)$. A total of 45,000 models were trained for this experiment: 30 models for each pruning experiment, repeated over 50 random splits, for each of the 5 methods and three 3 datasets with 2 protected attributes each.

\begin{figure*}[t]
\centering
\subfloat[]{\includegraphics[width=.49\textwidth,trim=0mm 0mm 80mm 10mm, clip]{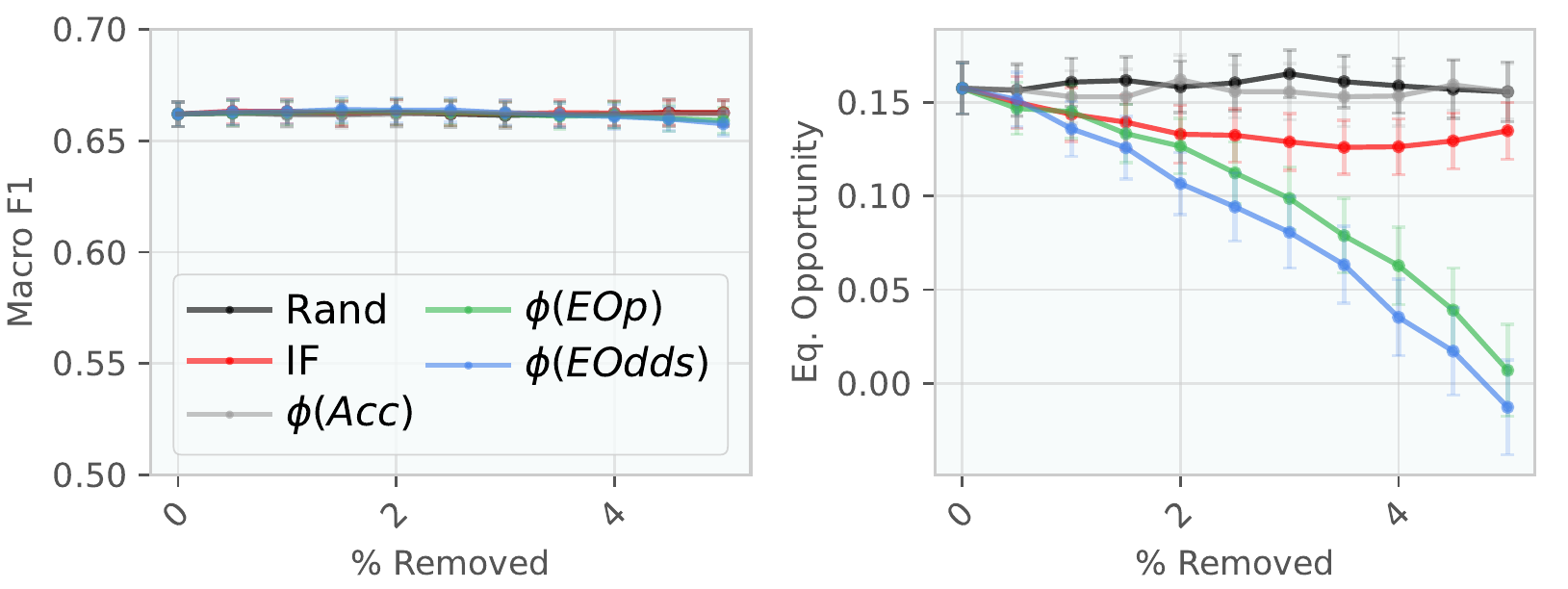}
}
\subfloat[]{\includegraphics[width=.49\textwidth,trim=0mm 0mm 80mm 10mm, clip]{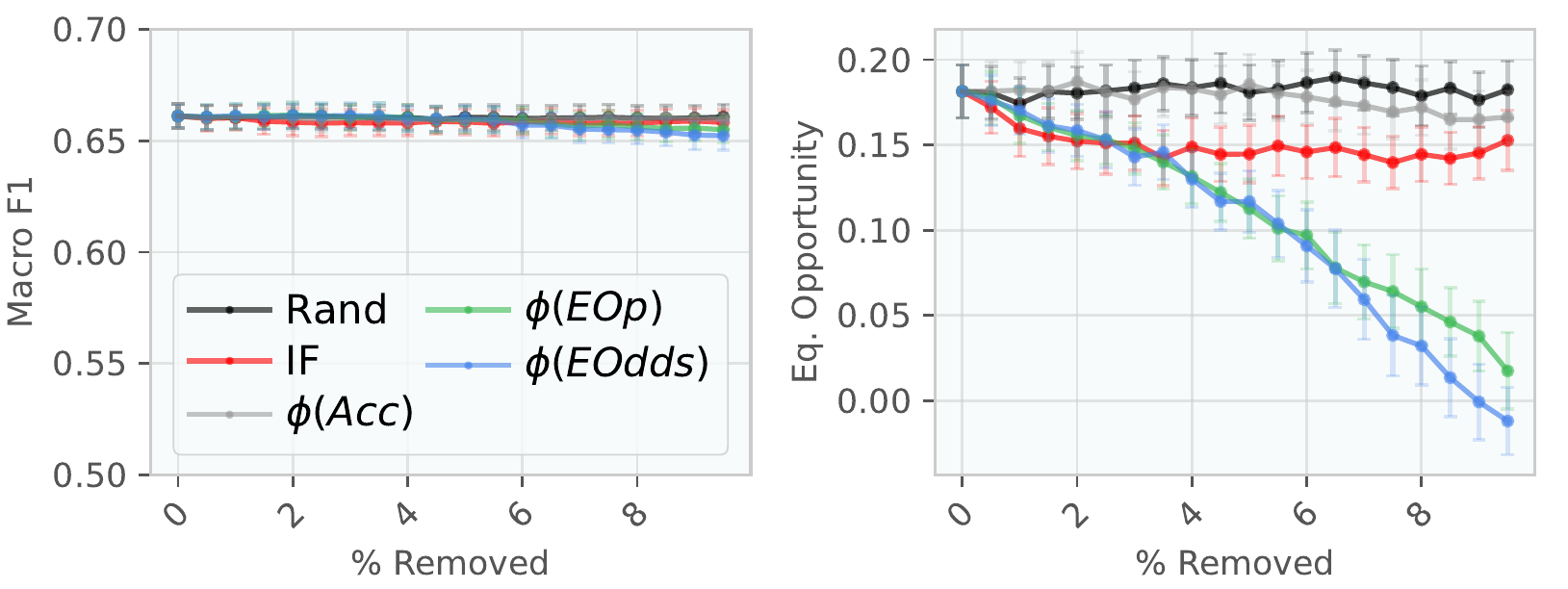}
}

\subfloat[]{\includegraphics[width=.49\textwidth,trim=0mm 0mm 80mm 10mm, clip]{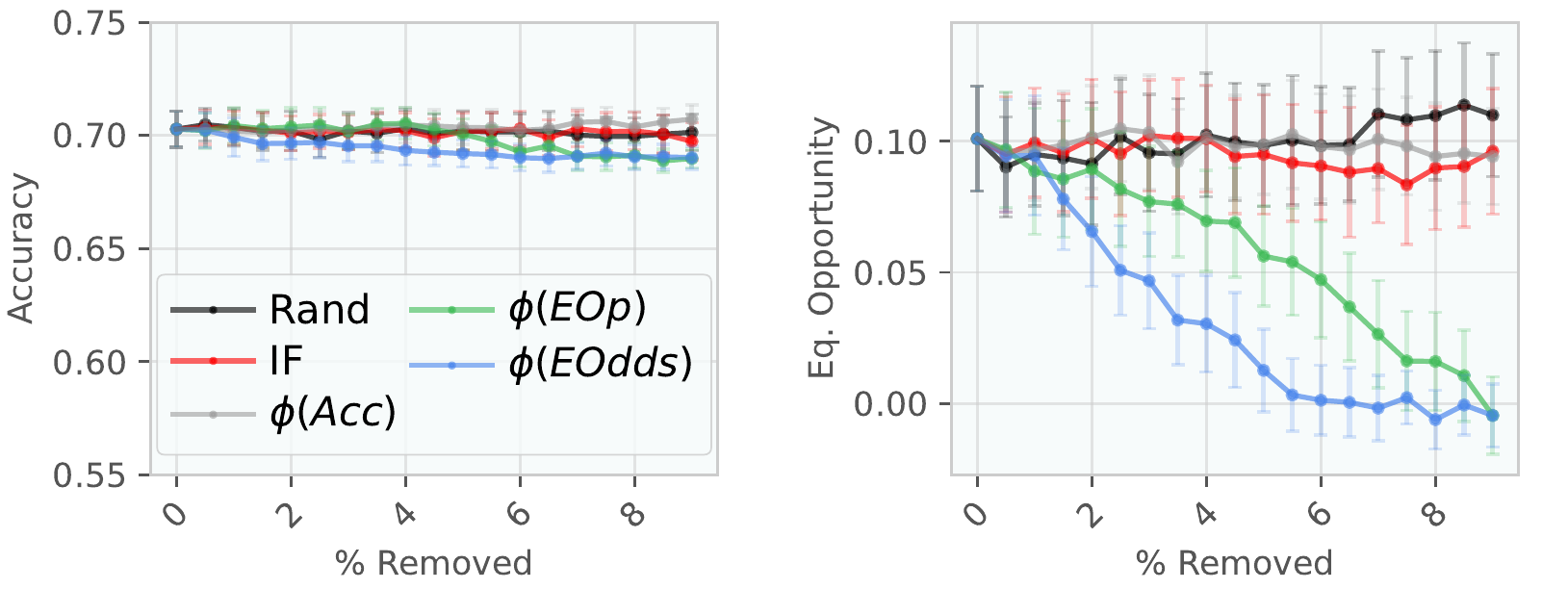}
}
\subfloat[]{\includegraphics[width=.49\textwidth,trim=0mm 0mm 80mm 10mm, clip]{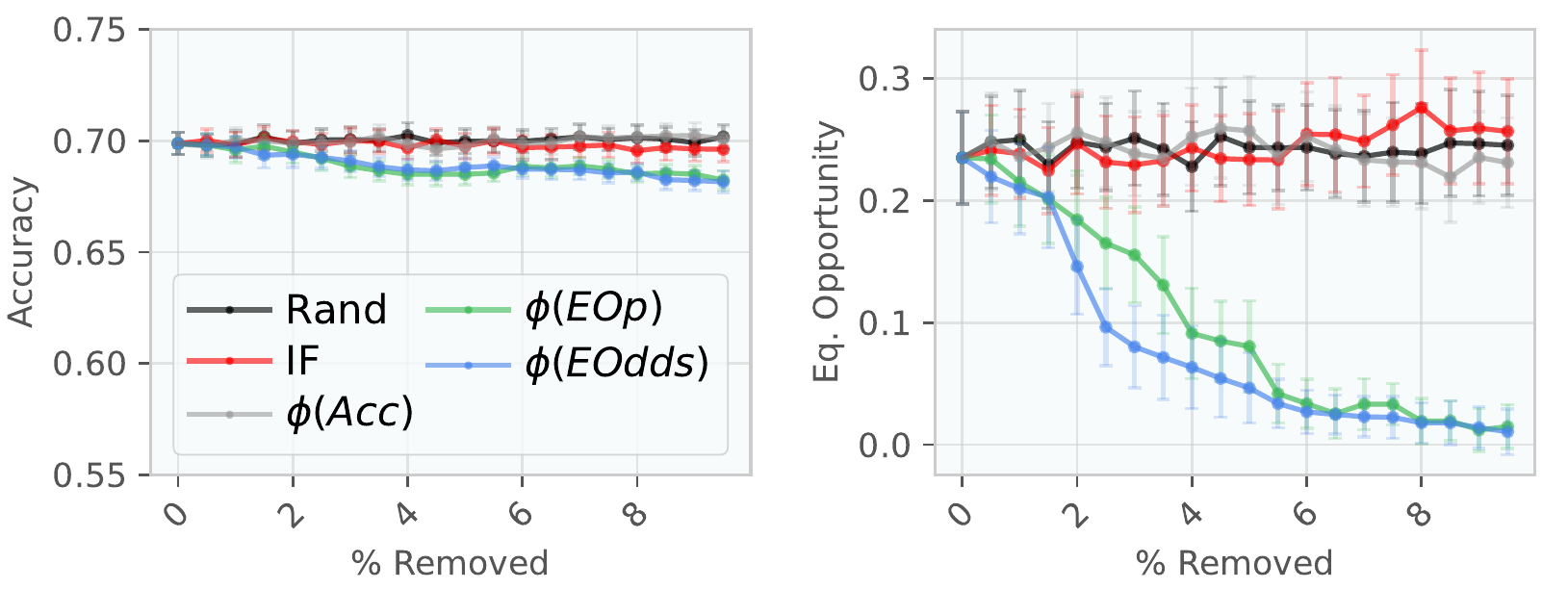}
}

\subfloat[]{\includegraphics[width=.49\textwidth,trim=0mm 0mm 80mm 10mm, clip]{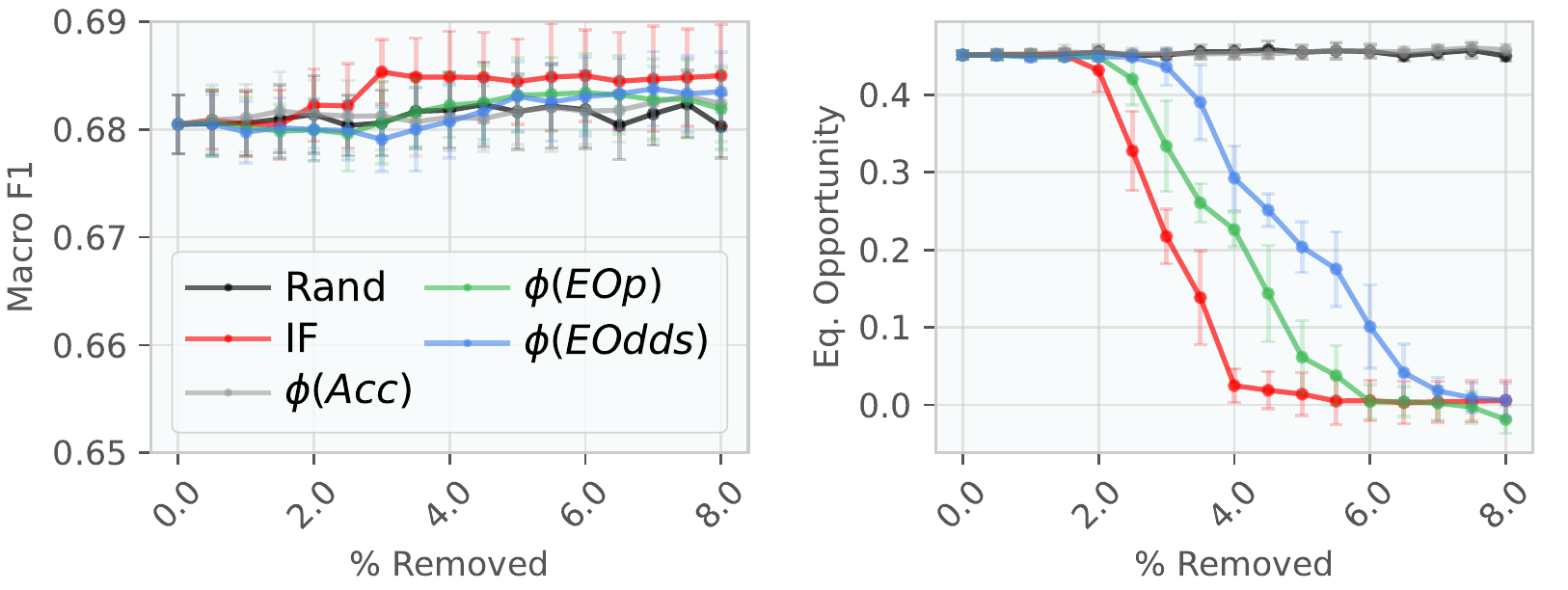}
}
\subfloat[]{\includegraphics[width=.49\textwidth,trim=0mm 0mm 82mm 10mm, clip]{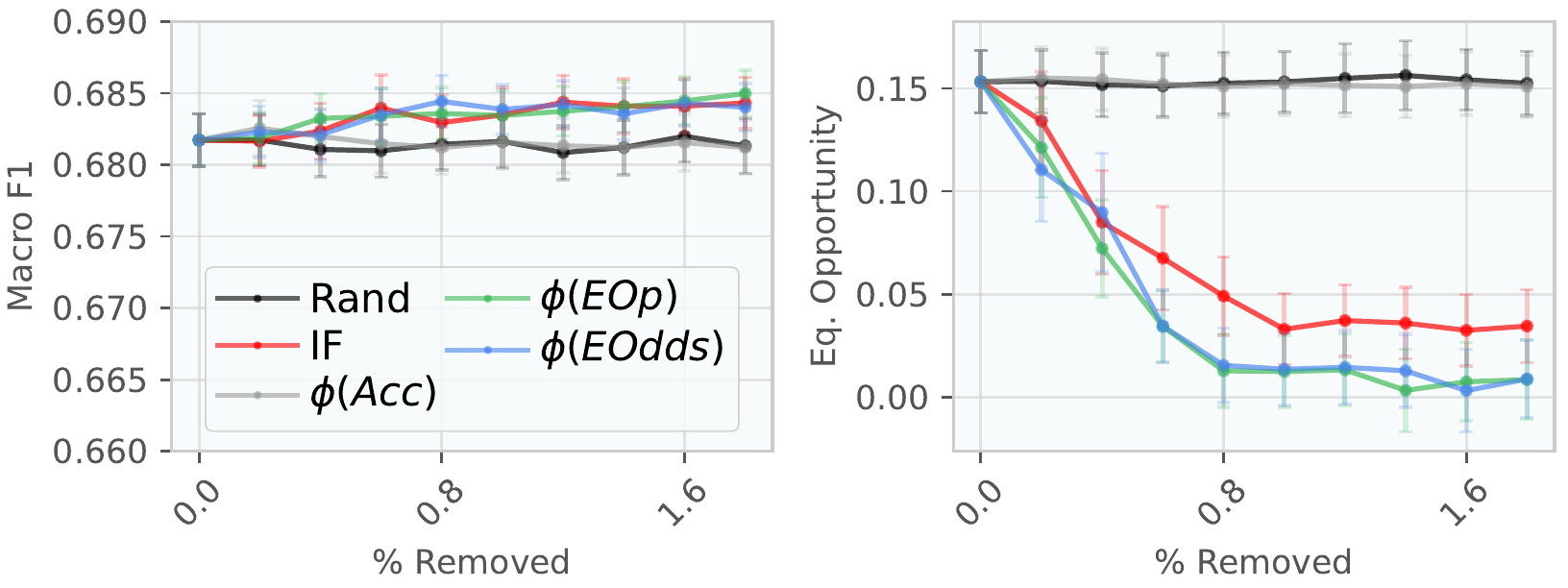}
}
\caption{Performance and fairness metrics as a function of the percentage of data removed according to different valuation functions. Observe how data pruning according to \texttt{FairShap} is able to significantly increase the fairness of the resulting model while maintaining competitive levels of accuracy, particularly on the COMPAS and German datasets. Subfigures depict results for (a) COMPAS - Sex, (b) COMPAS - Race, (c) German - Sex, (d) German - Age, (e) Adult - Sex and (f) Adult - Race.}
\label{fig:pruning-main}
\end{figure*}

\cref{fig:pruning-main} depicts the results. As seen in the Figure, data removal according to \texttt{FairShap} significantly improves the fairness of the resulting classifier even when removing a small percentage of samples (about 5\% in the COMPAS and 10\% in the German datasets). It is also more consistent in improving fairness than Influence Functions which perform well on the Adult dataset but exhibit poor performance on the COMPAS and German datasets. Additional results for all metrics and datasets are included in Appendix D.5. %

\subsection{Instance-level data re-weighting on image data}
\label{sec:expAeqY}
\noindent In this scenario, we focus on a computer vision task to illustrate the versatility of data re-weighting via \texttt{FairShap}. In this case, the goal is to predict the sensitive attribute, i.e. $A=Y$. 
Furthermore, this scenario explores the benefits of leveraging a fair external reference dataset $\T$. The task consists of automatic sex classification from facial images by means of a deep convolutional network (Inception Resnet V1) using \texttt{FairShap} for data re-weighting. Sex (male/female) is therefore both the protected attribute ($A$) and the target variable ($Y$). The pipeline of this scenario is illustrated in \cref{fig:aeqypipeline}.

\textbf{Datasets.} We leverage three publicly available face datasets: CelebA, LFWA~\citep{liu2015imgdatasets} and FairFace~\citep{karkkainen2021fairface}, where LFWA is the training set $\D$ (large-scale and biased) and FairFace is the reference dataset $\T$ (small but fair). 
The test split in the FairFace dataset is used for testing. 
 CelebA is used to pre-train the Inception Resnet V1 model~\citep{szegedy2017inception} to obtain the LFWA and FairFace embeddings that are needed to compute the Shapley Values efficiently by means of a k-NN approximation in the embedding space. In the three datasets, sex is a binary variable with two values: male, female. 

 \begin{figure*}[t]
\centering
\subfloat[]{
\includegraphics[width=.62\linewidth]{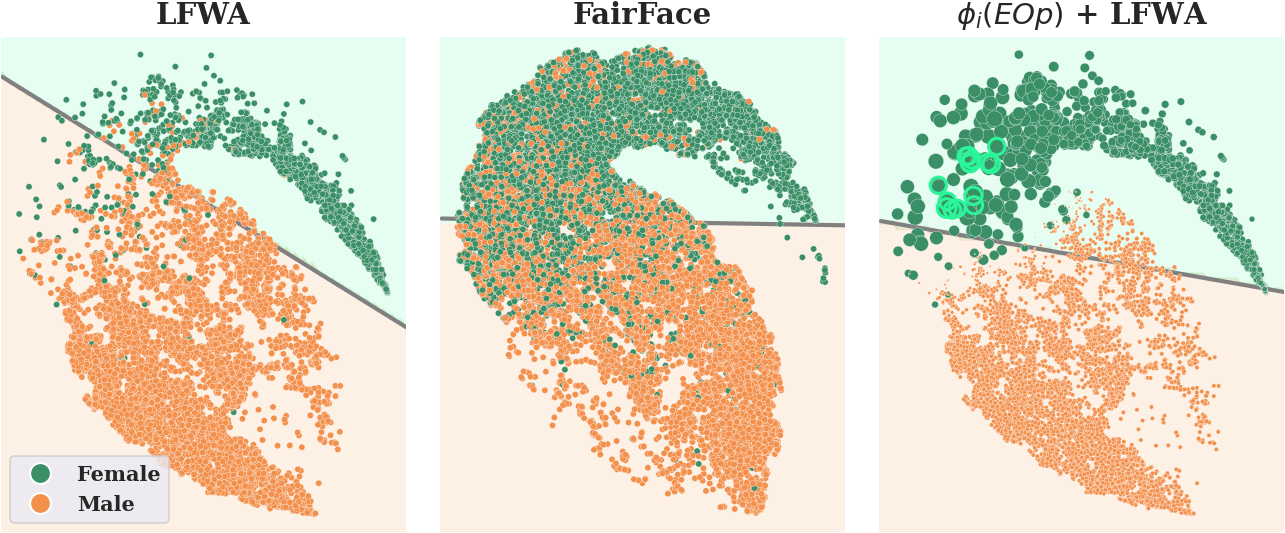}
    \label{fig:imglatent}
}
\hfill
\subfloat[]{\includegraphics[width=.34\linewidth]{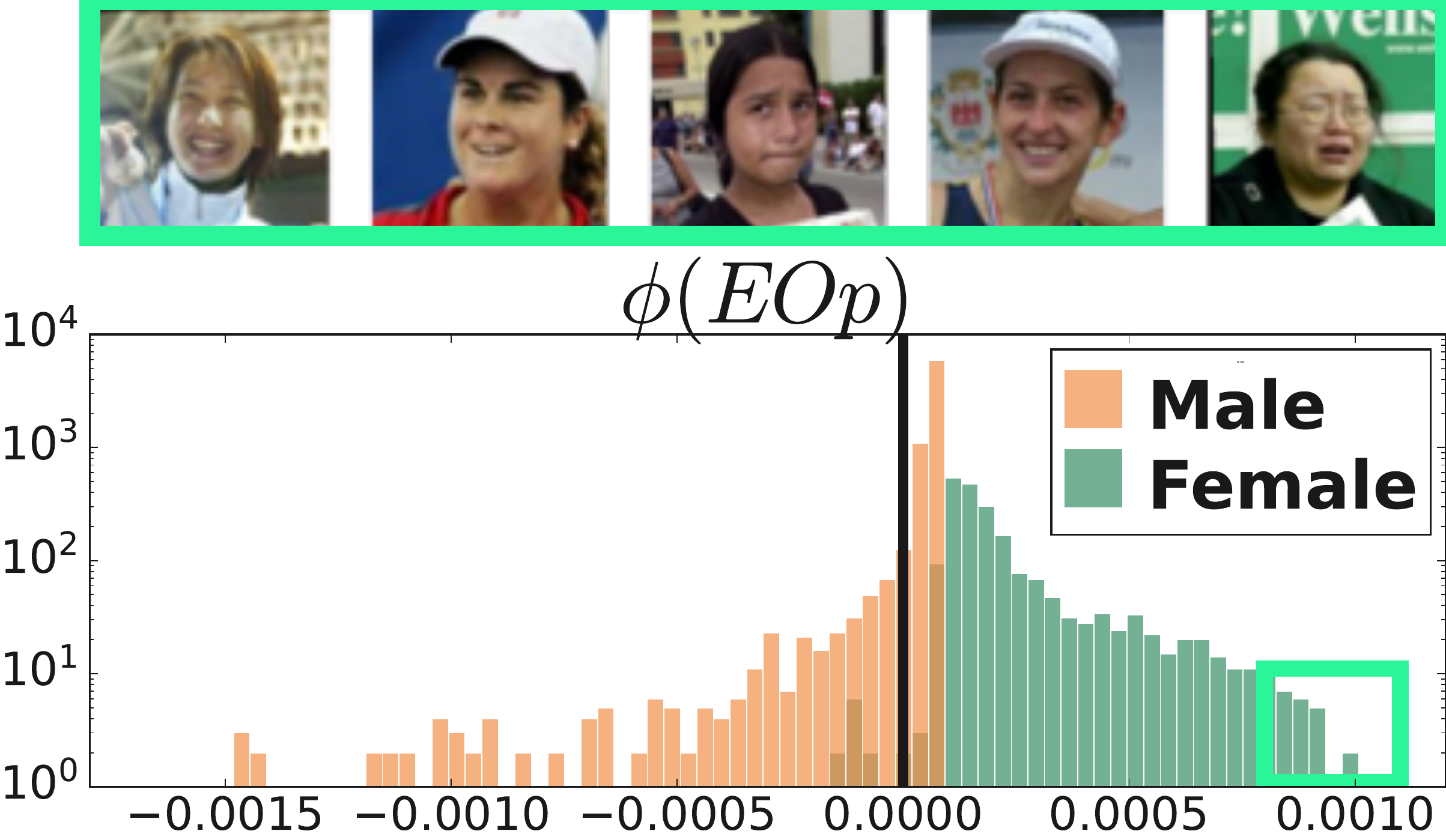}
    \label{fig:histLFWAsveopTop5}
}
\caption{(a) Left: LFWA embeddings. (a) Middle: FairFace embeddings. (a) Right: LFWA embeddings with data point sizes $\propto \phi_i(\EOp)$. Points with the largest $\phi_i(\EOp)$ are highlighted in green. Note how they all correspond to examples of females near the decision boundary of the original LFWA model. As a result of the data-reweighting, the decision boundary has been shifted, yielding a fairer model. (b) 5 Images with the largest $\phi_i(\EOp)$ and histogram of $\phi_i(\EOp)$ on the LFWA dataset.}
\label{fig:SVIMGS}
\end{figure*}

\textbf{Pipeline.}
The pipeline to obtain the \texttt{FairShap}'s weights in this scenario is depicted in \cref{fig:aeqypipeline} and 
proceeds as follows: 
(1) Pre-train an Inception Resnet V1 model with the CelebA dataset; (2) Use this model to obtain the embeddings of the LFWA and FairFace datasets; (3) Compute the weights on the LFWA training set ($\D$) using as reference dataset ($\T$) the FairFace validation partition. 
(4) Fine-tune the pretrained model using the re-weighted data in the LFWA training set according to $\bm{\phi}$; and (5) Test the resulting model on the test partition of the FairFace dataset. The experiment's training details and hyper-parameter setting are described in Appendix D.3.%

\textbf{FairShap Re-weighting.} In this case, the group fairness metrics are equivalent and thus we report results using $\phi_i(\EOp)$: $\phi_i(\EOp)$ quantifies the contribution of the $i$th data point (image) in LFWA to the fairness metric (Equal Opportunity) of the model tested on the FairFace dataset. 

\textbf{Baselines.} The purpose of this experiment is to illustrate the versatility of \texttt{FairShap} in different scenarios rather than to perform an exhaustive comparison with other methods, as previously done with tabular data. Nonetheless, we compare \texttt{FairShap} with three baselines: the pre-trained model using CelebA; the fine-tuned model using LFWA without re-weighting; and a data re-weighting approach using $\bm{\phi(\Acc)}$ from \citep{ghorbani2019data}.
We report two performance metrics: the accuracy of the models in correctly classifying the sex in the images (\textsc{Acc}) and the Equal Opportunity (\textsc{EOp}), measured as $\TPR_M-\TPR_W$ where $W$ is the disadvantaged group (females in thfis case). We also report the specific TPR for males and females. A summary of the experimental setup for this scenario is depicted in~\cref{fig:aeqypipeline}. 

\textbf{Results.} The results of this experiment are summarized in \cref{tab:aeqyexp}. Note how both re-weighting approaches ($\bm{\phi(\Acc)}$ and \texttt{FairShap}) significantly improve the fairness metrics while \emph{increasing the accuracy} of the model. \texttt{FairShap} yields the best results \textbf{both in fairness and accuracy}. Regarding \textsc{EOp}, the model trained with data re-weighted according to \texttt{FairShap} yields improvements of \textbf{88\%} and \textbf{66\%}  when compared to the model trained without re-weighting (LFWA) and the model trained with weights according to $\bm{\phi(\Acc)}$, respectively. In sum, data re-weighting with \texttt{FairShap} is able to leverage complex models trained on biased datasets and improve both their fairness and accuracy.

\begin{table}[th]
\caption{Performance of the Inception Resnet V1 model tested on the FairFace dataset without and with re-weighting and with binary protected attribute $A$=$Y$=sex. The arrows next to the metrics' name indicate if the optimal result of the metric is 0 ($\downarrow$) or 1 ($\uparrow$).}
\label{tab:aeqyexp}
\begin{center}
\begin{small}
\begin{tabular}{lccc}
\midrule
Training Set & Acc$\uparrow$ & TPR$_W$ $\mid$ TPR$_M$ & EOp $\downarrow$ \\
\midrule
FairFace            & 0.909               & 0.906 $\mid$ 0.913              & 0.007\\ %
\midrule
CelebA              & 0.759              & 0.580 $\mid$ 0.918          & 0.34 \\
LFWA                & 0.772              & 0.635 $\mid$ 0.896          & 0.26 \\
$\bm{\phi}(\Acc)$   & 0.793              & 0.742 $\mid$ 0.839                        & 0.09 \\
\texttt{FairShap} - $\bm{\phi}(\EOp)$   & \textbf{0.799}             & 0.782 $\mid$ 0.813   & \textbf{0.03}\\ %
\bottomrule
\end{tabular}
\end{small}
\end{center}
\end{table}

To gain a better understanding of the behavior of \texttt{FairShap} in this scenario, \cref{fig:histLFWAsveopTop5} (bottom) depicts a histogram of the $\bm{\phi(\EOp)}$ values on the LFWA training dataset. As seen in the Figure, $\bm{\phi_i(\EOp)}$ are mostly positive for the examples labeled as \textit{female} (green) and mostly zero or negative for the examples labeled as \textit{male} (orange). This result makes intuitive sense given that the original model is biased against females, i.e. the probability of misclassification is significantly higher for the images labeled as female than for those labeled as male. \cref{fig:histLFWAsveopTop5} (top) depicts the five images with the largest $\phi_i(\EOp)$: they all belong to the female category and depict faces with a variety of poses, different facial expressions and from diverse races. 

Note that in this case \texttt{FairShap} behaves like a distribution shift method.
\cref{fig:imglatent} shows how $\phi_i(\EOp)$ shifts the distribution of $\D$ (LFWA) to be as similar as possible to the distribution of the reference dataset $\T$ (FairFace). Therefore, biased datasets (such as $\D$) may be debiased by re-weighting their data according to $\phi_i(\EOp)$, yielding models with competitive performance both in terms of accuracy and fairness. \cref{fig:imglatent} illustrates how the group fairness metrics impact individual data points: critical data points are those near the decision boundary. This finding is consistent with recent work that has proposed using Shapley Values to identify counterfactual samples~\citep{albini2022counterfactual}.

\subsection{Computational cost} 

\begin{figure}[t]
\centering
\subfloat[]{\includegraphics[width=.7\linewidth]{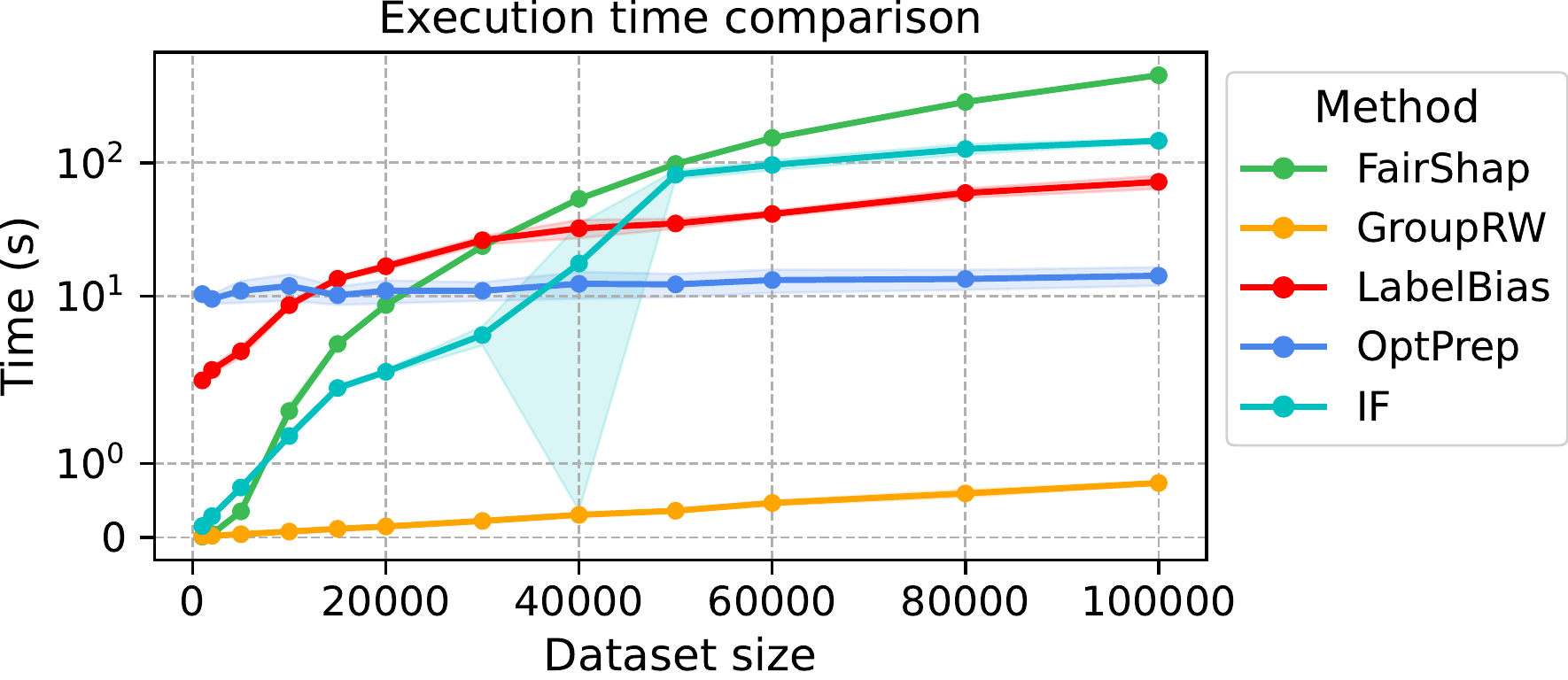}
}

\subfloat[]{\includegraphics[width=.7\linewidth]{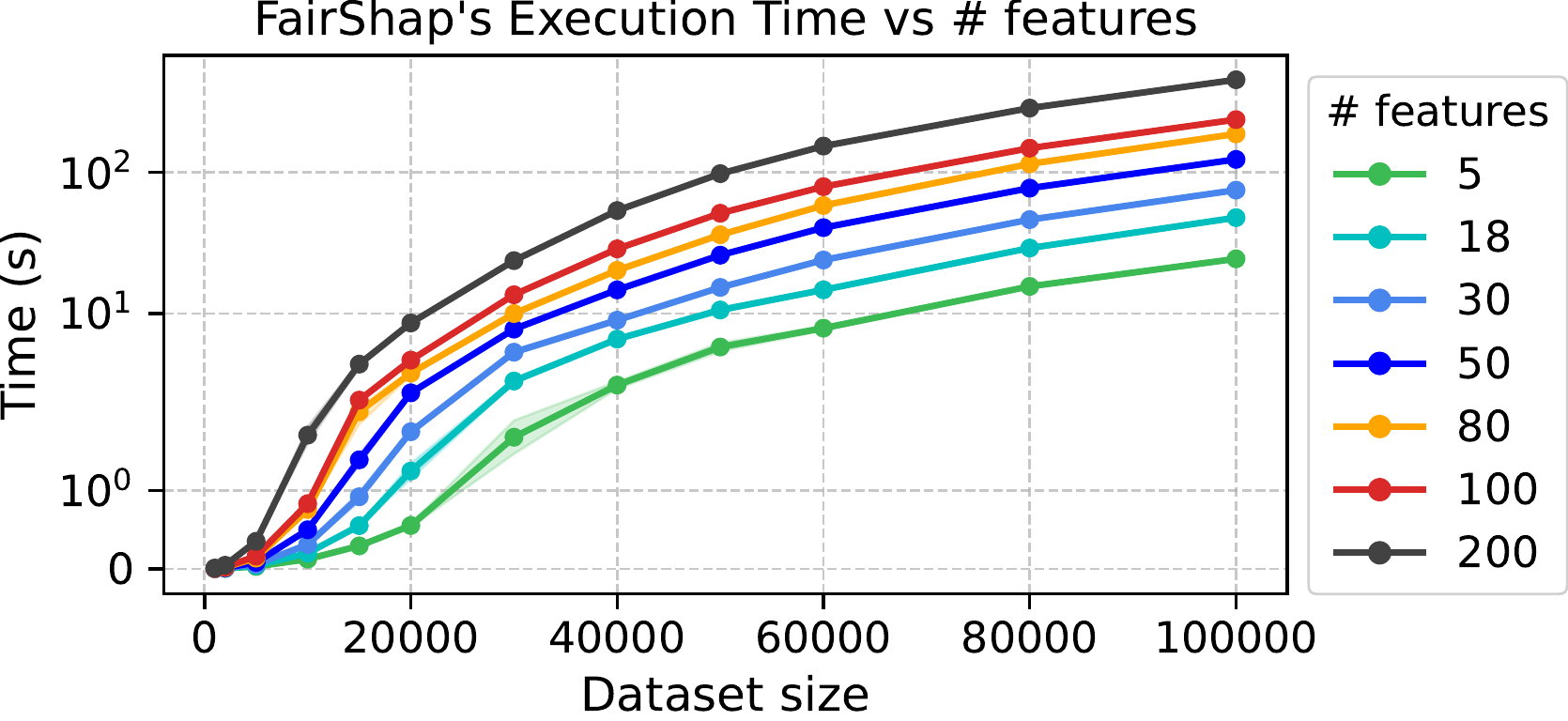}
}

\subfloat[]{\includegraphics[width=.85\linewidth]{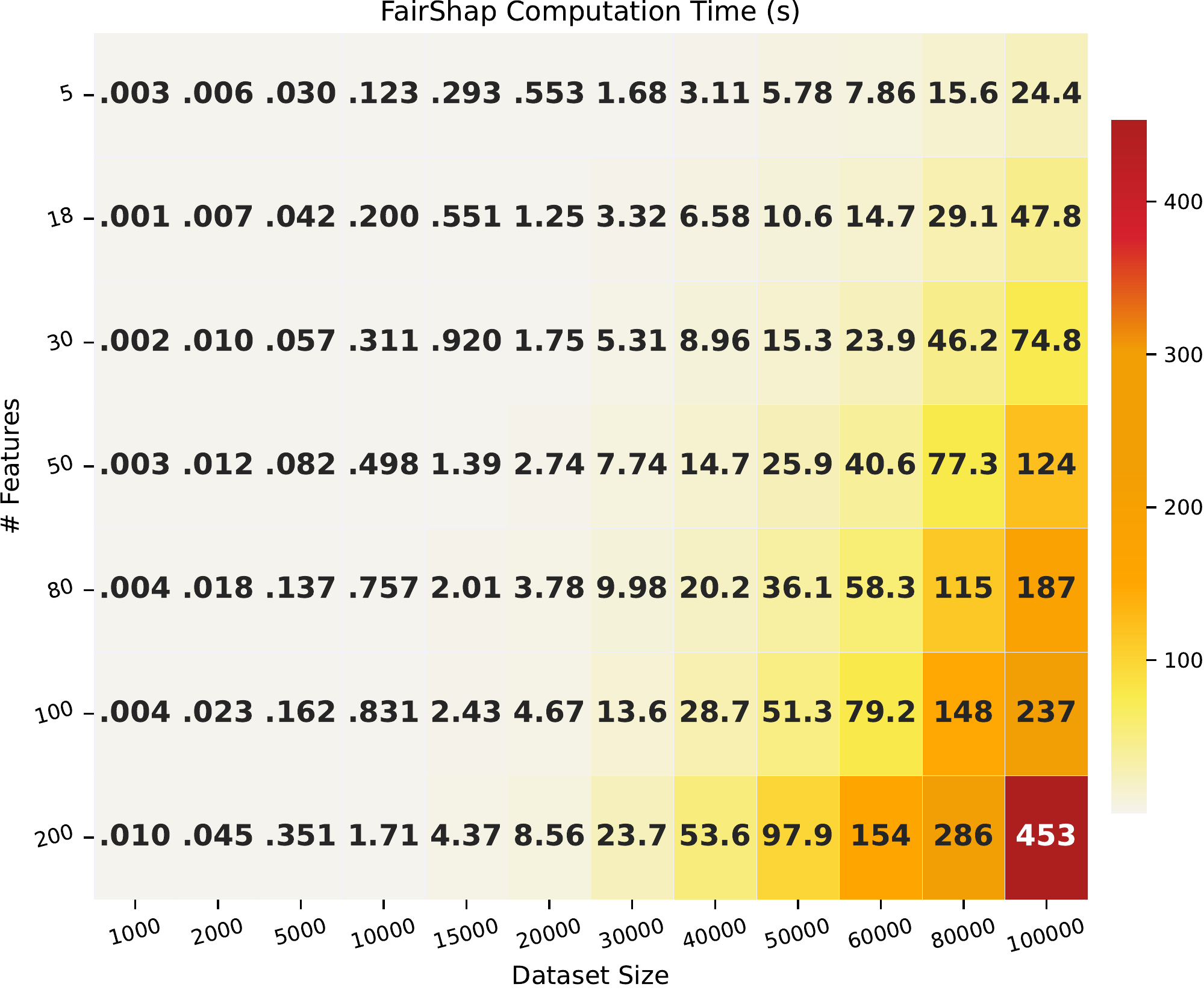}
}
\caption{Run time comparison of re-weighting via \texttt{FairShap} and baselines with respect to data set size (a) and number of features (b). Datasets are split in 80\% as $D$ and 20\% as $T$. We report mean and std run times for 10 iterations. (c) \texttt{FairShap}'s execution times (in seconds) on datasets with increasing numbers of features and sizes. In all experiments, the CPU is an Intel i7-1185G7 3.00GHz.}
    \label{fig:runningtime}
\end{figure}

\noindent  We describe experiments to illustrate \texttt{FairShap}'s computational cost relative to other approaches by applying data re-weighting on synthetic datasets of varying sizes, ranging from 1k to 100k data points, each comprising 200 features. We compare the run time (in seconds) of computing the weights using \texttt{FairShap}, Group Re-weighting~\citep{kamiran2012data}, OptPrep~\citep{calmon2017optimized}, LabelBias~\citep{jiang2020identifying} and IFs~\citep{li2022individualreweighing}. We leave the post-processing~\citep{hardt2016equality} approach out of the comparison since its based on tweaking thresholds after a model is trained, such that the running time heavily depends on the training time of the model of choice. In these experiments, we allocate 80\% of the data for training and 20\% for validation. With 10 iterations for each configuration, we compute the mean and standard deviation of the run time on an Intel i7-1185G7 3.00GHz CPU. Results are reported in~\cref{fig:runningtime}.

As seen in the Figure, instance level re-weighting via \texttt{FairShap} is computationally competitive for datasets with up to 30k data points. Group Re-weighting and LabelBias are computationally more efficient than \texttt{FairShap} on datasets with >30k data points.  

Note that OptPrep~\citep{calmon2017optimized} and IFs~\citep{li2022individualreweighing} require a hyperparameter search for each model and each dataset, yielding a significant increase on the computation time. In our experiments, we used the hyperparameters provided by the authors and hence did not have to tune them. Consequently, the actual running time for these methods would significantly increase depending on the number of hyperparameter configurations to be tested. For example, OptPrep consistently requires $\approx10$ seconds regardless of the dataset's size. However, a hyperparameter grid-search scenario with 20 different hyperparameter settings and 10-fold cross-validation, would increase the run time to 2,000 seconds (i.e., 10s/it x 20 x 10) or 20,000s for IFs on a dataset size of 60,000 samples (i.e., 100s/it x 20 x 10). These run times are significantly larger than those required to compute \texttt{FairShap}'s weights.

The Figure also depicts \texttt{FairShap}'s execution times (in seconds) with different numbers of features in datasets of increasing sizes. As seen in the Figure, three datasets with 60k, 80k, and 100k instances and feature dimension of 18, have runtimes of 14.7s, 29.1s, and 47.8s. 

Finally, we provide an overview of \texttt{FairShap}'s run times for the experiments described in \cref{sec:expAneqY}. On the German dataset, \texttt{FairShap} has an average execution time of 0.001s$\pm$0.002, where $|\D|=700$, $|\T|=150$, and there are 11 features. In the case of the Adult dataset, the execution time remains consistent at 12.7s$\pm$3, for a dataset with $|\D|=34,189$, $|\T|=7,326$, and 18 features. Finally, for the COMPAS dataset, the run time is 0.063s$\pm$0.004 on a dataset with $|\D|=3,694$, $|\T|=792$, and 10 features. These numbers are consistent with the run times reported in \cref{fig:runningtime}.

\section{Conclusion, Discussion and Future Work}
\noindent In this paper, we have proposed \texttt{FairShap}, an instance-level, model-agnostic data re-weighting approach to achieve group fairness via data valuation using Shapley Values. It is based on the proposed formalization of the pairwise contribution of a training point to the correct classification of a reference point, $\phi_{ij}$. We have empirically validated \texttt{FairShap} with several state-of-the-art datasets in different scenarios and using two different types of models (GBCs and deep neural networks). In our experimental results, the models trained with data re-weighted according to \texttt{FairShap} delivered competitive accuracy and fairness results. Our experiments also highlight the value of using fair reference datasets ($\T$) for data valuation. 
We have illustrated the interpretability of \texttt{FairShap} by means of histograms and a latent space visualization. 
We have also studied the accuracy vs fairness trade-off, the impact of the size of the reference dataset and \texttt{FairShap}'s computational cost when compared to baseline models.  From our experiments, we conclude that data re-weighting by means of \texttt{FairShap} could be a valuable approach to achieve algorithmic fairness. Furthermore, from a practical perspective, \texttt{FairShap} satisfies interpretability desiderata proposed by legal stakeholders and upcoming regulations. 

In future work, we plan to explore the use of more efficient and more robust Shapley Value computations~\citep{wang2023data}. In addition, we plan to assess \texttt{FairShap}'s potential to generate counterfactual explanations and its potential to guide data generation processes. We will also explore sub-group re-weighting instead of instance-level re-weighting.

\section{Acknowledgments}
Funded by the European Union. Views and opinions expressed are however those of the author(s) only and do not necessarily reflect those of the European Union or the European Health and Digital Executive Agency (HaDEA). EU - HE ELIAS -- Grant Agreement 101120237. Funded also by Intel corporation, a nominal grant received at the ELLIS Unit Alicante Foundation from the Regional Government of Valencia in Spain (Convenio Singular signed with Generalitat Valenciana, Conselleria de Innovación, Industria, Comercio y Turismo, Dirección General de Innovación) and a grant by the Banc Sabadell Foundation.

\bibliographystyle{IEEEtran}
\bibliography{refs}

\begin{thebibliography}{10}
\providecommand{\url}[1]{#1}
\csname url@samestyle\endcsname
\providecommand{\newblock}{\relax}
\providecommand{\bibinfo}[2]{#2}
\providecommand{\BIBentrySTDinterwordspacing}{\spaceskip=0pt\relax}
\providecommand{\BIBentryALTinterwordstretchfactor}{4}
\providecommand{\BIBentryALTinterwordspacing}{\spaceskip=\fontdimen2\font plus
\BIBentryALTinterwordstretchfactor\fontdimen3\font minus
  \fontdimen4\font\relax}
\providecommand{\BIBforeignlanguage}[2]{{%
\expandafter\ifx\csname l@#1\endcsname\relax
\typeout{** WARNING: IEEEtran.bst: No hyphenation pattern has been}%
\typeout{** loaded for the language `#1'. Using the pattern for}%
\typeout{** the default language instead.}%
\else
\language=\csname l@#1\endcsname
\fi
#2}}
\providecommand{\BIBdecl}{\relax}
\BIBdecl

\bibitem{barocas2017fairness}
S.~Barocas, M.~Hardt, and A.~Narayanan, \emph{Fairness and Machine Learning:
  Limitations and Opportunities}.\hskip 1em plus 0.5em minus 0.4em\relax The
  MIT Press, 2019.

\bibitem{smuha2019ethics}
N.~Smuha, ``{Ethics guidelines for trustworthy AI},'' in \emph{AI \& Ethics,
  Date: 2019/05/28-2019/05/28, Brussels, Belgium}.\hskip 1em plus 0.5em minus
  0.4em\relax {European Commission}, 2019.

\bibitem{oliver2022srip}
N.~Oliver, ``{Artificial intelligence for social good - The way forward},'' in
  \emph{{Science, Research and Innovation performance of the EU 2022
  report}}.\hskip 1em plus 0.5em minus 0.4em\relax {European Commission}, 2022,
  ch.~11, pp. 604--707.

\bibitem{carey2022statistical}
A.~N. Carey and X.~Wu, ``The statistical fairness field guide: perspectives
  from social and formal sciences,'' \emph{AI and Ethics}, pp. 1--23, 2022.

\bibitem{hardt2016equality}
M.~Hardt, E.~Price, and N.~Srebro, ``Equality of opportunity in supervised
  learning,'' \emph{Advances in Neural Information Processing Systems},
  vol.~29, 2016.

\bibitem{gomez2017eodds}
M.~B. Zafar, I.~Valera, M.~Gomez~Rodriguez, and K.~P. Gummadi, ``Fairness
  beyond disparate treatment and disparate impact: Learning classification
  without disparate mistreatment,'' in \emph{International Conference on World
  Wide Web}, 2017, p. 1171–1180.

\bibitem{dwork2012fairness}
C.~Dwork, M.~Hardt, T.~Pitassi, O.~Reingold, and R.~Zemel, ``Fairness through
  awareness,'' in \emph{Proceedings of the 3rd innovations in theoretical
  computer science conference}, 2012, pp. 214--226.

\bibitem{mehrabi2021survey}
N.~Mehrabi, F.~Morstatter, N.~Saxena, K.~Lerman, and A.~Galstyan, ``A survey on
  bias and fairness in machine learning,'' \emph{ACM Computing Surveys (CSUR)},
  vol.~54, no.~6, pp. 1--35, 2021.

\bibitem{kamiran2012data}
F.~Kamiran and T.~Calders, ``Data preprocessing techniques for classification
  without discrimination,'' \emph{Knowledge and information systems}, vol.~33,
  no.~1, pp. 1--33, 2012.

\bibitem{zemel2013learning}
R.~Zemel, Y.~Wu, K.~Swersky, T.~Pitassi, and C.~Dwork, ``Learning fair
  representations,'' in \emph{International Conference on Machine
  Learning}.\hskip 1em plus 0.5em minus 0.4em\relax PMLR, 2013, pp. 325--333.

\bibitem{zhang2018mitigating}
B.~H. Zhang, B.~Lemoine, and M.~Mitchell, ``Mitigating unwanted biases with
  adversarial learning,'' in \emph{Proceedings of the 2018 AAAI/ACM Conference
  on AI, Ethics, and Society}, 2018, pp. 335--340.

\bibitem{kamishima2012fairness}
T.~Kamishima, S.~Akaho, H.~Asoh, and J.~Sakuma, ``Fairness-aware classifier
  with prejudice remover regularizer,'' in \emph{Joint European conference on
  machine learning and knowledge discovery in databases}.\hskip 1em plus 0.5em
  minus 0.4em\relax Springer, 2012, pp. 35--50.

\bibitem{feldman2015certifying}
M.~Feldman, S.~A. Friedler, J.~Moeller, C.~Scheidegger, and
  S.~Venkatasubramanian, ``Certifying and removing disparate impact,'' in
  \emph{Proceedings of the 21th ACM SIGKDD International Conference on
  Knowledge Discovery and Data Mining}, 2015, pp. 259--268.

\bibitem{hacker2022varieties}
P.~Hacker and J.-H. Passoth, ``Varieties of ai explanations under the law. from
  the gdpr to the aia, and beyond,'' in \emph{International Workshop on
  Extending Explainable AI Beyond Deep Models and Classifiers}.\hskip 1em plus
  0.5em minus 0.4em\relax Springer, 2022, pp. 343--373.

\bibitem{madaio2022assessing}
M.~Madaio, L.~Egede, H.~Subramonyam, J.~Wortman~Vaughan, and H.~Wallach,
  ``Assessing the fairness of ai systems: Ai practitioners' processes,
  challenges, and needs for support,'' \emph{Proceedings of the ACM on
  Human-Computer Interaction}, vol.~6, no. CSCW1, pp. 1--26, 2022.

\bibitem{gebru2021datasheets}
T.~Gebru, J.~Morgenstern, B.~Vecchione, J.~W. Vaughan, H.~Wallach, H.~D. Iii,
  and K.~Crawford, ``Datasheets for datasets,'' \emph{Communications of the
  ACM}, vol.~64, no.~12, pp. 86--92, 2021.

\bibitem{hagendorff2020ethics}
T.~Hagendorff, ``The ethics of ai ethics: An evaluation of guidelines,''
  \emph{Minds and Machines}, vol.~30, no.~1, pp. 99--120, 2020.

\bibitem{chouldechova2020snapshot}
A.~Chouldechova and A.~Roth, ``A snapshot of the frontiers of fairness in
  machine learning,'' \emph{Communications of the ACM}, vol.~63, no.~5, pp.
  82--89, 2020.

\bibitem{ghorbani2019data}
A.~Ghorbani and J.~Zou, ``{Data shapley: Equitable valuation of data for
  machine learning},'' in \emph{International Conference on Machine
  Learning}.\hskip 1em plus 0.5em minus 0.4em\relax PMLR, 2019, pp. 2242--2251.

\bibitem{shapley1953value}
L.~S. Shapley, ``A value for n-person games,'' \emph{Contributions to the
  Theory of Games}, vol.~2, pp. 307--317, 1953.

\bibitem{chouldechova2017fair}
A.~Chouldechova, ``Fair prediction with disparate impact: A study of bias in
  recidivism prediction instruments,'' \emph{Big data}, vol.~5, no.~2, pp.
  153--163, 2017.

\bibitem{barocas2016big}
S.~Barocas and A.~D. Selbst, ``Big data's disparate impact,'' \emph{California
  law review}, pp. 671--732, 2016.

\bibitem{krasanakis2018adaptive}
E.~Krasanakis, E.~Spyromitros-Xioufis, S.~Papadopoulos, and Y.~Kompatsiaris,
  ``Adaptive sensitive reweighting to mitigate bias in fairness-aware
  classification,'' in \emph{Proceedings of the 2018 world wide web
  conference}, 2018, pp. 853--862.

\bibitem{jiang2020identifying}
H.~Jiang and O.~Nachum, ``Identifying and correcting label bias in machine
  learning,'' in \emph{International Conference on Artificial Intelligence and
  Statistics}.\hskip 1em plus 0.5em minus 0.4em\relax PMLR, 2020, pp. 702--712.

\bibitem{chai2022adaptative}
J.~Chai and X.~Wang, ``Fairness with adaptive weights,'' in \emph{International
  Conference on Machine Learning}, ser. Proceedings of Machine Learning
  Research, vol. 162.\hskip 1em plus 0.5em minus 0.4em\relax PMLR, 17--23 Jul
  2022, pp. 2853--2866.

\bibitem{jung2023reweighting}
S.~Jung, T.~Park, S.~Chun, and T.~Moon, ``Re-weighting based group fairness
  regularization via classwise robust optimization,'' in \emph{The Eleventh
  International Conference on Learning Representations}, 2023.

\bibitem{caton2020fairness}
S.~Caton and C.~Haas, ``Fairness in machine learning: A survey,'' \emph{ACM
  Comput. Surv.}, August 2023.

\bibitem{ali2021accounting}
J.~Ali, P.~Lahoti, and K.~P. Gummadi, ``Accounting for model uncertainty in
  algorithmic discrimination,'' in \emph{Proceedings of the 2021 AAAI/ACM
  Conference on AI, Ethics, and Society}, 2021, pp. 336--345.

\bibitem{sim2022data}
R.~H.~L. Sim, X.~Xu, and B.~K.~H. Low, ``Data valuation in machine
  learning:“ingredients”, strategies, and open challenges,'' in \emph{Proc.
  IJCAI}, 2022, pp. 5607--5614.

\bibitem{koh2017understanding}
P.~W. Koh and P.~Liang, ``Understanding black-box predictions via influence
  functions,'' in \emph{International Conference on Machine Learning}.\hskip
  1em plus 0.5em minus 0.4em\relax PMLR, 2017, pp. 1885--1894.

\bibitem{pruthi2022datainfluence}
G.~Pruthi, F.~Liu, S.~Kale, and M.~Sundararajan, ``Estimating training data
  influence by tracing gradient descent,'' in \emph{Advances in Neural
  Information Processing Systems}, vol.~33, 2020, pp. 19\,920--19\,930.

\bibitem{paul2021dataimportance}
M.~Paul, S.~Ganguli, and G.~K. Dziugaite, ``Deep learning on a data diet:
  Finding important examples early in training,'' in \emph{Advances in Neural
  Information Processing Systems}, 2021.

\bibitem{Sundararajan2017axiomatic}
M.~Sundararajan, A.~Taly, and Q.~Yan, ``Axiomatic attribution for deep
  networks,'' in \emph{International Conference on Machine Learning}, ser.
  Proceedings of Machine Learning Research, vol.~70.\hskip 1em plus 0.5em minus
  0.4em\relax PMLR, 06--11 Aug 2017, pp. 3319--3328.

\bibitem{black2021loofairness}
E.~Black and M.~Fredrikson, ``Leave-one-out unfairness,'' in \emph{ACM
  Conference on Fairness, Accountability, and Transparency}, 2021, p.
  285–295.

\bibitem{wang2022instancelevel}
J.~Wang, X.~E. Wang, and Y.~Liu, ``{Understanding Instance-Level Impact of
  Fairness Constraints},'' in \emph{International Conference on Machine
  Learning}, vol. 162.\hskip 1em plus 0.5em minus 0.4em\relax PMLR, 17--23 Jul
  2022, pp. 23\,114--23\,130.

\bibitem{li2022individualreweighing}
P.~Li and H.~Liu, ``Achieving fairness at no utility cost via data reweighing
  with influence,'' in \emph{International Conference on Machine Learning},
  vol. 162.\hskip 1em plus 0.5em minus 0.4em\relax PMLR, 17--23 Jul 2022, pp.
  12\,917--12\,930.

\bibitem{basu2021influence}
S.~Basu, P.~Pope, and S.~Feizi, ``Influence functions in deep learning are
  fragile,'' in \emph{International Conference on Learning Representations},
  2021.

\bibitem{kwon2022betashap}
Y.~Kwon and J.~Zou, ``Beta shapley: a unified and noise-reduced data valuation
  framework for machine learning,'' in \emph{International Conference on
  Artificial Intelligence and Statistics}.\hskip 1em plus 0.5em minus
  0.4em\relax PMLR, 28--30 Mar 2022, pp. 8780--8802.

\bibitem{hammoudeh2024training}
Z.~Hammoudeh and D.~Lowd, ``Training data influence analysis and estimation: A
  survey,'' \emph{Machine Learning}, vol. 113, no.~5, pp. 2351--2403, 2024.

\bibitem{wang2023data}
J.~T. Wang and R.~Jia, ``Data banzhaf: A robust data valuation framework for
  machine learning,'' in \emph{International Conference on Artificial
  Intelligence and Statistics}.\hskip 1em plus 0.5em minus 0.4em\relax PMLR,
  2023, pp. 6388--6421.

\bibitem{wu2022davinz}
Z.~Wu, Y.~Shu, and B.~K.~H. Low, ``{DAVINZ}: Data valuation using deep neural
  networks at initialization,'' in \emph{International Conference on Machine
  Learning}, 2022.

\bibitem{gillies1959core}
D.~B. Gillies, ``Solutions to general non-zero-sum games,'' \emph{Contributions
  to the Theory of Games}, vol.~4, pp. 47--85, 1959.

\bibitem{wang2019measure}
G.~Wang, C.~X. Dang, and Z.~Zhou, ``Measure contribution of participants in
  federated learning,'' in \emph{2019 IEEE international conference on big data
  (Big Data)}.\hskip 1em plus 0.5em minus 0.4em\relax IEEE, 2019, pp.
  2597--2604.

\bibitem{brophy2020exit}
J.~Brophy, ``Exit through the training data: A look into instance-attribution
  explanations and efficient data deletion in machine learning,''
  \emph{Technical report Oregon University}, 2020.

\bibitem{schoch2022csshapley}
S.~Schoch, H.~Xu, and Y.~Ji, ``{CS}-shapley: Class-wise shapley values for data
  valuation in classification,'' in \emph{Advances in Neural Information
  Processing Systems}, 2022.

\bibitem{fern2021text}
X.~Fern and Q.~Pope, ``Text counterfactuals via latent optimization and
  shapley-guided search,'' in \emph{Proceedings of the 2021 Conference on
  Empirical Methods in Natural Language Processing}, 2021, pp. 5578--5593.

\bibitem{albini2022counterfactual}
E.~Albini, J.~Long, D.~Dervovic, and D.~Magazzeni, ``Counterfactual shapley
  additive explanations,'' in \emph{2022 ACM Conference on Fairness,
  Accountability, and Transparency}, 2022, pp. 1054--1070.

\bibitem{molnar2020interpretable}
C.~Molnar, \emph{Interpretable machine learning}.\hskip 1em plus 0.5em minus
  0.4em\relax Lulu. com, 2020.

\bibitem{lundberg2017unified}
S.~M. Lundberg and S.-I. Lee, ``A unified approach to interpreting model
  predictions,'' \emph{Advances in Neural Information Processing Systems},
  vol.~30, 2017.

\bibitem{gultchin2022beyond}
L.~Gultchin, V.~Cohen-Addad, S.~Giffard-Roisin, V.~Kanade, and
  F.~Mallmann-Trenn, ``Beyond impossibility: Balancing sufficiency, separation
  and accuracy,'' \emph{In NeurIPS Workshop on Algorithmic Fairness through the
  Lens of Causality and Privacy}, 2022.

\bibitem{jia2019knn}
R.~Jia, D.~Dao, B.~Wang, F.~A. Hubis, N.~M. Gurel, B.~Li, C.~Zhang, C.~Spanos,
  and D.~Song, ``Efficient task-specific data valuation for nearest neighbor
  algorithms,'' \emph{Proc. VLDB Endow.}, vol.~12, no.~11, p. 1610–1623,
  2019.

\bibitem{jiang2023opendataval}
K.~F. Jiang, W.~Liang, J.~Zou, and Y.~Kwon, ``{OpenDataVal: a Unified Benchmark
  for Data Valuation},'' in \emph{Advances in Neural Information Processing
  Systems Datasets and Benchmarks Track}, 2023.

\bibitem{berk2021fairness}
R.~Berk, H.~Heidari, S.~Jabbari, M.~Kearns, and A.~Roth, ``Fairness in criminal
  justice risk assessments: The state of the art,'' \emph{Sociological Methods
  \& Research}, vol.~50, no.~1, pp. 3--44, 2021.

\bibitem{kamiran2009german}
F.~Kamiran and T.~Calders, ``Classifying without discriminating,'' in
  \emph{2009 2nd international conference on computer, control and
  communication}.\hskip 1em plus 0.5em minus 0.4em\relax IEEE, 2009, pp. 1--6.

\bibitem{kohavi1996adult}
R.~Kohavi \emph{et~al.}, ``Scaling up the accuracy of naive-bayes classifiers:
  A decision-tree hybrid.'' in \emph{Kdd}, vol.~96, 1996, pp. 202--207.

\bibitem{angwin2016machine}
J.~Angwin, J.~Larson, S.~Mattu, and L.~Kirchner, ``Machine bias: There’s
  software used across the country to predict future criminals. and it’s
  biased against blacks. propublica, may 23,'' 2016.

\bibitem{friedman2001greedy}
J.~H. Friedman, ``Greedy function approximation: a gradient boosting machine,''
  \emph{Annals of statistics}, pp. 1189--1232, 2001.

\bibitem{calmon2017optimized}
F.~Calmon, D.~Wei, B.~Vinzamuri, K.~Natesan~Ramamurthy, and K.~R. Varshney,
  ``Optimized pre-processing for discrimination prevention,'' in \emph{Advances
  in Neural Information Processing Systems}, vol.~30, 2017.

\bibitem{liu2015imgdatasets}
Z.~Liu, P.~Luo, X.~Wang, and X.~Tang, ``Deep learning face attributes in the
  wild,'' in \emph{Proceedings of the IEEE international conference on computer
  vision}, 2015, pp. 3730--3738.

\bibitem{karkkainen2021fairface}
K.~Karkkainen and J.~Joo, ``Fairface: Face attribute dataset for balanced race,
  gender, and age for bias measurement and mitigation,'' in \emph{Proceedings
  of the IEEE/CVF Winter Conference on Applications of Computer Vision}, 2021,
  pp. 1548--1558.

\bibitem{szegedy2017inception}
C.~Szegedy, S.~Ioffe, V.~Vanhoucke, and A.~A. Alemi, ``Inception-v4,
  inception-resnet and the impact of residual connections on learning,'' in
  \emph{Thirty-first AAAI conference on artificial intelligence}, 2017.

\bibitem{aif360}
R.~K. Bellamy, K.~Dey, M.~Hind, S.~C. Hoffman, S.~Houde, K.~Kannan, P.~Lohia,
  J.~Martino, S.~Mehta, A.~Mojsilovi{\'c} \emph{et~al.}, ``Ai fairness 360: An
  extensible toolkit for detecting and mitigating algorithmic bias,'' \emph{IBM
  Journal of Research and Development}, vol.~63, no. 4/5, pp. 4--1, 2019.

\bibitem{cruz2023unprocessing}
A.~F. Cruz and M.~Hardt, ``Unprocessing seven years of algorithmic fairness,''
  in \emph{The Twelfth International Conference on Learning Representations},
  2024.

\end{thebibliography}

\clearpage
\newpage
\onecolumn
\appendices

\section{Preliminaries}
\subsection{Notation}

\begin{table*}[ht]
\caption{Notation.}
\label{tab:notation}
\centering
{
\small
\begin{tabular}{lp{4in}} 
\toprule
Symbol                  & Description \\
\toprule
$\D=\{(x_i,y_i)\}_{i=0}^n$                 & Training dataset \\
$\T=\{(x_j,y_j)\}_{i=0}^m$                 & Reference dataset\\
$S\subseteq \D$                 & Subset of a dataset $\D$ \\
\midrule
$A$    & Set of variables that are protected attributes. \\
$\text{TPR}_{A=a}$ & True positive rate for test points with values in the protected attribute equal to $a$. Also $\TPR_{a}$ if the protected attribute is known. The same logic applies to FPR, TNR, and FNR.\\
\midrule
$p(y|x,\D)$ & Predictive distribution of data point x when trained with $\D$.\\
$p(y=y_j|x_j,\D)$ & Likelihood of correct classification of data point $x$ when trained with $\D$.\\
$\phi_i(\D,v)$       & Shapley Value for data point $i$ in the 
training dataset $\D$ according to the performance function $v$ \\
$\bm{\phi}(\D,v)$    & Vector with all the SVs of the entire dataset $\in \R^{|\D|}$.\\
$v(S,T)$            & Value of dataset $S$ w.r.t a reference dataset $\T$. E.g., the accuracy of a model trained with $S$ tested on $\T$ ($v=\Acc$) or the value of Equal Opportunity of of a model trained with $S$ tested on $\T$ ($v=\text{EOp}$)\\
$\bm{\Phi}\in \R^{|\D|\times |\T|}$           & Matrix where $\Phi_{i,j}$ is the contribution of the training point $i \in \D$ to the correct classification of $j \in \T$ according to \ref{appsubsec:knnsv}\\
$\overline{\bm{\Phi}}_{i,:}$    & Mean of row $i$ \\
$\overline{\bm{\Phi}}_{i,:|A=a}$    & Mean of row $i$ conditioned to columns where $A=a$\\
$\mathbf{1}$    & Vector of ones := $[1,1,...,1]$\\
\bottomrule
\end{tabular}
}
\end{table*}

\subsection{Desiderata}

\begin{table*}[htbp]
  \caption{Comparative properties of related algorithmic fairness methods}
    \label{tab:desiderates}
  \centering
  \begin{small}
  \begin{tabular}{lcccccc}
    \toprule
    Method & {\bf D1} & {\bf D2} & {\bf D3} & {\bf D4} & {\bf D5} & {\bf D6} \\  
    & {\scriptsize Data Val.} & {\scriptsize Interpretable} & {\scriptsize Pre-processing} & {\scriptsize Model agnostic} & {\scriptsize Data RW} & {\scriptsize Instance-level} \\
    \midrule
    \textbf{\texttt{FairShap}} & \tick & \tick & \tick & \tick & \tick & \tick  \\
    Group-RW & \cross & \tick & \tick & \tick & \tick & \cross  \\
    Influence Functions  & \tick & \cross & \cross & \cross & \tick & \cross  \\
    Inpro-RW (LabelBias) & \cross & \tick & \cross & \cross & \tick & \cross  \\
    Massaging (OptPre)  & \cross & \tick & \tick & \tick & \cross & \cross  \\
    Post-pro & \cross & \tick & \cross & \tick & \cross & \cross  \\
    \bottomrule
  \end{tabular}
  \end{small}
\end{table*}

As previously noted in \cref{sec:related}, the closest methods to \texttt{FairShap} in the literature are 
\textit{Influence Functions}~\citep{wang2022instancelevel, li2022individualreweighing},
\textit{In-processing reweighting} (e.g. LabelBias) ~\citep{krasanakis2018adaptive, jiang2020identifying, chai2022adaptative},
\textit{Group reweighting}~\citep{kamiran2012data}
and \textit{Massaging} (e.g. OptPre)~\citep{feldman2015certifying, calmon2017optimized}.

\textbf{D1 - Data valuation method.} Our aim is to propose a novel fairness-aware data valuation approach. Thus, the first desired property concerns whether the method performs data valuation or not~\citep{hammoudeh2024training}. Data valuation methods compute the contribution or influence of a given data point to a target function, typically by analyzing the interactions between points (LOO, pair-wise or all the subsets in the data powerset).

\textbf{D2 - Interpretable.} The method should be easy to understand by a broad set of technical and non-technical stakeholders when applied to a variety of scenarios and purposes, including for data minimization, data acquisition policies, data selection for transfer learning, active learning, data sharing, mislabeled example detection and federated learning.

\textbf{D3 - Pre-processing.} The method should provide data insights that can be applicable to train a wide variety of ML learning methods.

\textbf{D4 - Model agnostic.} The computation of model-weights, data valuation values, data insights or data transformations should not rely on learning a model iteratively, to enhance flexibility, computational efficiency, interpretability and mitigate uncertainty. Therefore, this follows the guidelines to make data valuation models data-driven~\citep{sim2022data}.

\textbf{D5 - Data Re-weighting.} The data insights drawn from applying the method should be in the form of weights to be applied to the data, which can be used to rebalance the dataset.

\textbf{D6 - Instance-Level.} Different insights or weights are given to each of the data points.

\subsection{Clarification of the concept of fairness}
Note that the concept of fairness in the definition of the Shapley Values (SVs) is different from algorithmic fairness. The former relates to the desired quality of the SVs to be proportional to how much each data point contributes to the model's performance. Formally, this translates to the SVs fulfilling certain properties (e.g. efficiency, symmetry, additivity...) to ensure a fair payout. The latter refers to the concept of fairness used in the machine learning literature, as described in the introduction. \texttt{FairShap} uses Shapley Values for data valuation in a pre-processing approach with the objective of  mitigating bias in machine learning models. As \texttt{FairShap} is based on the theory of Shapley Values, it also fulfills their four axiomatic properties. %

\subsection{Algorithmic Fairness Definitions}

As aforementioned, the fairness metrics used as valuation functions in \texttt{FairShap} depend on the \emph{conditioned} true/false negative/positive rates, depending on the protected attribute $A$:
\begin{align*}
\text{TPR}_{A=a} &:= \mathbb{P}[\hat{Y}=1|Y=1, A=a],\:\: \text{TNR}_{A=a}:= \mathbb{P}[\hat{Y}=0|Y=0, A=a]\\
\text{FPR}_{A=a} &:= \mathbb{P}[\hat{Y}=1|Y=0, A=a],\:\: \text{FNR}_{A=a}:= \mathbb{P}[\hat{Y}=0|Y=1, A=a]
\end{align*}

Note that different fairness metrics are defined by forcing the equality in true/false negative/positive rates between different protected groups. For instance, in a binary classification scenario with a binary sensitive attribute, Equal Opportunity (EOp) and Equalized Odds (EOdds) are defined as follows:
\begin{align*}
\text{EOp} &:= \mathbb{P}[\hat{y}=1|Y=1, A=a]=\mathbb{P}[\hat{Y}=1|Y=1]\\
\text{EOdds} &:= \mathbb{P}[\hat{y}=1|Y=i, A=a]=\mathbb{P}[\hat{Y}=1|Y=i], \forall i \in \{0,1\}
\end{align*}

In practical terms, the  metrics above are relaxed and computed as the difference for the different groups:
\begin{align*}
\text{EOp} &:= \text{TPR}_{A=a} - \text{TPR}_{A=b}, \:\: \text{EOdds}:= \frac{1}{2} ((\text{TPR}_{A=a} - \text{TPR}_{A=b}) +(\text{FPR}_{A=a} - \text{FPR}_{A=b}))
\end{align*}

The proposed Fair Shapley Values include as their valuation function these group fairness metrics.

\section{Shapley Values proposed in \texttt{FairShap}}
\label{app:sec:allfairshap}

\texttt{FairShap} proposes $\bm{\phi(\EOp)}$ and $\bm{\phi(\EOds)}$ as the data valuation functions to compute the Shapley Values of individual data points in the training set. These functions are computed from the $\bm{\phi(\TPR)}, \bm{\phi(\FPR)}, \bm{\phi(\TNR)}$ and $\bm{\phi(\FNR)}$ functions, leveraging the Efficiency axiom of the SVs, and the decomposability properties of fairness metrics. 

First, using our proposed formalization of the contribution of a training data point to the correct classification of a test datapoint defined as 
\begin{align}
    \Phi_{i,j} = \mathbb{E}_{S\sim\mathcal{P}(\D \backslash \{i\})}\left[p(y=y_j|x_j, S \cup \{i\}) - p(y=y_j|x_j, S)\right],
\end{align}
the Shapley value for accuracy \citep{jia2019knn} can be redefined as 
 $\phi_i(\Acc) := \frac{1}{|\T|} \sum_{j\in \T} \Phi_{i,j} = \mathbb{E}_{j \sim p(\T)}[\Phi_{i,j}]$.
In the following, we summarize all proposed valuation functions presented in this work:

\textbf{True/False Positive/Negative rates}:
\begin{align*}
&\phi_i(\TPR) := \mathop{\mathbb{E}}_{j \sim p(\T|Y=1)}\left[\mathop{\mathbb{E}}_{S\sim\mathcal{P}(\D\backslash \{i\})}\left[p(y=1|x_j, S \cup \{i\}) - p(y=1|x_j, S)\right]\right] = \mathop{\mathbb{E}}_{j \sim p(\T|Y=1)}[\Phi_{i,j}] = \frac{\sum_{j\in \T} \Phi_{i,j} \mathbb{I}[y_j=1]}{|\{x: x \in \T | y=1\}|}\nonumber\\
&\phi_i(\TNR) := \Exp_{j \sim p(\T|Y=0)}\left[\Exp_{S\sim\mathcal{P}(\D\backslash \{i\})}\left[p(y=0|x_j, S \cup \{i\}) - p(y=0|x_j, S)\right]\right] = \Exp_{j \sim p(\T|Y=0)}[\Phi_{i,j}]= \frac{\sum_{j\in \T} \Phi_{i,j} \mathbb{I}[y_j=0]}{|\{x: x \in \T | y=0\}|}\nonumber\\
&\phi_i(\FNR) := \Exp_{j \sim p(\T|Y=1)}\left[\Exp_{S\sim\mathcal{P}(\D\backslash \{i\})}\left[p(y=0|x_j, S \cup \{i\}) - p(y=0|x_j, S)\right]\right]= \frac{1}{|\D|}-\phi_i(\TPR)\\
&\phi_i(\FPR) := \Exp_{j \sim p(\T|Y=0)}\left[\Exp_{S\sim\mathcal{P}(\D\backslash \{i\})}\left[p(y=1|x_j, S \cup \{i\}) - p(y=1|x_j, S)\right]\right]= \frac{1}{|\D|}-\phi_i(\TNR)
\end{align*}

where to compute $\phi_i(\FNR)$ we use the following equality:
\begin{align*}
    &\TPR=1-\FNR \rightarrow  \sum \phi_i(\TPR) = 1 - \sum \phi_i(\FNR) \rightarrow  \sum \phi_i(\TPR) = \sum 1/n - \phi_i(\FNR) \\
    \rightarrow &\phi_i(\TPR) = 1/n - \phi_i(\FNR) \rightarrow \phi_i(\FNR) = 1/n - \phi_i(\TPR).
\end{align*}

\textbf{Conditioned True/False Positive/Negative rates}:

$\phi_i(\TPR_a) := \mathbb{E}_{j \sim p(\T|Y=1, A=a)}[\Phi_{i,j}] = \frac{\sum_{j\in \T} \Phi_{i,j} \mathbb{I}[y_j=1, A_j=a]}{|\{x: x \in \T | y=1, A=a\}|} = \overline{\bm{\Phi}}_{i,:|Y=1, A=a}$

$\phi_i(\TPR_b) := \mathbb{E}_{j \sim p(\T|Y=1, A=b)}[\Phi_{i,j}] = \frac{\sum_{j\in \T} \Phi_{i,j} \mathbb{I}[y_j=1, A_j=b]}{|\{x: x \in \T | y=1, A=b\}|} = \overline{\bm{\Phi}}_{i,:|Y=1, A=b}$

$\phi_i(\TNR_a) := \mathbb{E}_{j \sim p(\T|Y=0, A=a)}[\Phi_{i,j}] = \frac{\sum_{j\in \T} \Phi_{i,j} \mathbb{I}[y_j=0, A_j=a]}{|\{x: x \in \T | y=0, A=a\}|} = \overline{\bm{\Phi}}_{i,:|Y=0, A=a}$

$\phi_i(\TNR_b) := \mathbb{E}_{j \sim p(\T|Y=0, A=b)}[\Phi_{i,j}] = \frac{\sum_{j\in \T} \Phi_{i,j} \mathbb{I}[y_j=0, A_j=b]}{|\{x: x \in \T | y=0, A=b\}|} = \overline{\bm{\Phi}}_{i,:|Y=0, A=b}$

$\phi_i(\FPR_a) := \frac{1}{|\D|} - \phi_i(\TNR_a)$

$\phi_i(\FPR_b) := \frac{1}{|\D|} - \phi_i(\TNR_b)$

$\phi_i(\FNR_a) := \frac{1}{|\D|} - \phi_i(\TPR_a)$

$\phi_i(\FNR_b) := \frac{1}{|\D|} - \phi_i(\TPR_b)$

\vskip 0.1in
\textbf{When A=Y}:

$\phi_i(\EOp) := 
\phi_i(\EOp)=\phi_i(\TPR)+\phi_i(\TNR)-\frac{1}{|\D|}$

or its bounded version $\phi_i(\EOp)=\frac{\phi_i(\TPR)+\phi_i(\TNR)}{2}$. 

See \cref{sec:aeqymetrics} for more details on how to derive these formulas.

\vskip 0.1in
\textbf{When A$\neq$Y}:
\begin{align}
    \phi_i(\EOp) :=& \phi_i(\TPR_a) - \phi_i(\TPR_b) \nonumber\\
    =& \mathop{\mathbb{E}}_{j \sim p(\T|Y=1,A=a)}[\Phi_{i,j}]-\mathop{\mathbb{E}}_{j \sim p(\T|Y=1,A=b)}[\Phi_{i,j}]\\
    =&\mathop{\mathbb{E}}_{j \sim p(\T|Y=1,A=a)}\left[\mathop{\mathbb{E}}_{S\sim\mathcal{P}(\D\backslash \{i\})}\left[p(y=1|x_j, \D \cup \{i\}) - p(y=1|x_j, \D)\right]\right] 
    \nonumber\\
    -&\mathop{\mathbb{E}}_{j \sim p(\T|Y=1,A=b)}\left[\mathop{\mathbb{E}}_{S\sim\mathcal{P}(\D\backslash \{i\})}\left[p(y=1|x_j, \D \cup \{i\}) - p(y=1|x_j, \D)\right]\right]\nonumber
\end{align}
\begin{align}
    \phi_i(\EOds) :=& \frac{1}{2} \left((\phi_i(\FPR_a)-\phi_i(\FPR_b))+(\phi_i(\TPR_a)-\phi_i(\TPR_b))\right)\\
    =& \frac{1}{2} \left(\left(\left(\frac{1}{|\D|} - \phi_i(\TNR_a)\right)-\left(\frac{1}{|\D|} - \phi_i(\TNR_b)\right)\right)+\left(\Exp_{j \sim p(\T|Y=1,A=a)}[\Phi_{i,j}]-\Exp_{j \sim p(\T|Y=1,A=b)}[\Phi_{i,j}]\right)\right)\nonumber\\
    =&\frac{1}{2} \left(\left(\left(\frac{1}{|\D|} - \Exp_{j \sim p(\T|Y=0,A=a)}[\Phi_{i,j}]\right)-\left(\frac{1}{|\D|} - \Exp_{j \sim p(\T|Y=0,A=b)}[\Phi_{i,j}]\right)\right)\right.\nonumber\\
    +&\left.\left(\Exp_{j \sim p(\T|Y=1,A=a)}[\Phi_{i,j}]-\Exp_{j \sim p(\T|Y=1,A=b)}[\Phi_{i,j}]\right)\right)\nonumber
\end{align}

We refer to the reader to \cref{sec:aneqymetrics} for more details on how to obtain these formulas from the algorithmic fairness definitions.

\section{Methodology}

\subsection{Efficient \texorpdfstring{$k$}{k}-NN Shapley Value} \label{appsubsec:knnsv}

Jia et al \citep{jia2019knn} propose an efficient, exact calculation of the Shapley Values by means of a recursive $k$-NN algorithm with complexity $O(N \log N)$. The proposed method yields a matrix $\bm{\Phi} \in \R^{|\D|\times |\T|}$ with the contribution of each training point to the accuracy of each point in the reference data set $\T$. Therefore, $\Phi_{i,j}$ defines how much data point $i$ in the training set contributes to the probability of correct classification of data point $j$ in $\T$. The intuition behind is that $\Phi_{i,j}$ quantifies to which degree a training point $i$ helps in the correct classification of $j$. The $k$-NN-based recursive calculation is at follows.

For each $j$ in $\T$:
\begin{itemize}
    \item Order $i \in \D$ according to the distance to $j \in \T \rightarrow (x_1, x_2,...,x_N)$
    \item Calculate $\Phi_{i,j}$ recursively, starting from the furthest point:
    $$
    \Phi_{N,j}= \frac{I[y_{x_N}=y_j]}{N}
    $$
    $$
    \Phi_{i,j}= \Phi_{i+1, j} + \frac{I[y_i=y_{j}]-I[y_{i+1}=y_{j}]}{\max{K,i}}
    $$
    
    \item $\bm{\Phi}$ is a $|\D|\times|\T|$ matrix given by:

    $$
    \bm{\Phi} = 
    \begin{bmatrix}
    \Phi_{00} & \cdots & \Phi_{0|\T|}\\
    \vdots & \ddots & \vdots\\
    \Phi_{|\D|0} & \dots & \Phi_{|\D||\T|}\\
    \end{bmatrix} \in \R^{|\D|\times |\T|}
    $$
\end{itemize}

where $\Phi_{i,j}$ is the contribution of training point $i$ to the accuracy of the model on point $j$ in $\T$. Thus, the overall SV of a training point $i$ with respect to $\T$ is the average of all the values of row $i$ in the SV matrix:
$$
\phi_i(\Acc) = \frac{1}{m}\sum^{m}_{j=0} \Phi_{i,j} = \overline{\bm{\Phi}}_{i,:}\in \R 
$$

Note that the mean of a column $j$ in $\bm{\Phi}$ is the accuracy of the model on that test point. The vector with the SV of every training data point is computed as: 
$$
\bm{\phi}(\Acc) = [\phi_0, \cdots, \phi_n] \in \R^{|\D|}
$$

In addition, given the efficiency axiom of the Shapley Value, the sum of $\bm{\phi}$ is the accuracy of the model on the training set.
$$
    V(\D)=\sum^{n}_{i=0}\phi_i=\sum^{n}_{i=0}    \frac{1}{m}\sum^{m}_{j=0}\Phi_{i,j}=\text{Acc}
$$

Technically speaking, the process may be parallelized over all points in $\T$ (columns of the matrix) since the computation is independent, reducing the practical complexity from $O(N \log N)$ to $O(N)$. %

\subsection{Threshold independence} \label{appsubsec:thindep}
Computing $\bm{\phi}(\cdot)$ according to the original Shapley Value implementation (\cref{sec:sv}) entails evaluating the performance function $v(S)$ on each data point, which requires testing the model trained with $S$. As the group fairness metrics are based on different classification errors, they depend on the classification threshold $t$, such that $\text{TP}=|\{\hat Y >t | Y=1\}|$, $\text{TN}=|\{\hat Y <t | Y=0\}|$,  $\text{FP}=|\{\hat Y >t | Y=0\}|$ and $\text{FN}=|\{\hat Y <t | Y=1\}|$. %

However, the efficient method (\cref{appsubsec:knnsv}) is threshold independent since it calculates the accuracy as the average of the probability of correct classification for all test points, as shown in \cref{appsubsec:knnsv}. 

\subsection{\texorpdfstring{$\phi_i(\EOp)$}{sv(EOp)} derivation when \texorpdfstring{$A=Y$}{A=Y}} \label{sec:aeqymetrics}

When $A=Y$ in a binary classification task, TPR and TNR are the accuracies for each protected group, respectively. In this case, $\textrm{DP}$ collapses to $\mathbb{P}(\hat{Y}=1|A=a)\rightarrow \mathbb{P}(\hat{Y}=1|Y=a)$ . 
In this case, EOp measures the similarity of TPRs between groups.

As a result, when $A=Y$ in a binary classification scenario, the group fairness metrics measure the relationship between TPR, TNR, FPR and FNR not conditioned on the protected attribute $A$, since these metrics already depend on $Y$ and $A=Y$. As an example, Equal opportunity is defined in this case as 
$(\TPR+\TNR)/2 \in [0,1]$ \citep{hardt2016equality}:

\begin{align*}
\text{EOp} &= \frac{\text{TPR}-\text{FPR}+1}{2} = \frac{\text{TPR}-(1-\text{FNR})+1}{2} = \frac{\text{TPR}+\text{TNR}}{2} \in [0,1]
\end{align*}

Consequently, $\phi_i(\EOp)$ $\in [0,1]$ when $A=Y$ can be obtained as follows:
$$
\text{EOp} = \frac{\sum_{i \in \D} \phi_i(\TPR)+\sum_{i \in \D} \phi_i(\TNR)}{2} = \sum_{i \in \D}\frac{\phi_i(\TPR)}{2} + \sum_{i \in \D}\frac{\phi_i(\TNR)}{2}
$$

$$
\phi_i(\EOp) = \frac{\phi_i(\TPR)+\phi_i(\TNR)}{2}
$$

\subsection{\texorpdfstring{$\phi_i(\EOp)$}{SV(EOp)} and \texorpdfstring{$\phi_i(\EOds)$}{SV(EOds)} derivation when \texorpdfstring{$A\neq Y$}{A!=Y}} \label{sec:aneqymetrics}

We derive  $\bm{\phi}(\EOp)$ and $\bm{\phi}(\EOds)$ when $A\neq Y$ using the definitions for EOdds and EOp given by:
\begin{align*}
\text{EOp} &= \TPR_{A=a} - \TPR_{A=b}\\
\text{EOdds} &= \frac{1}{2} ((\TPR_{A=a} - \TPR_{A=b}) +(\FPR_{A=a} - \FPR_{A=b})) 
\end{align*}

Leveraging the Efficiency property of SVs, $\bm{\phi}(\EOp)$ is computed as:
\begin{align*}
\text{EOp} &= \sum_{i\in \D}\phi_i(\TPR_{A=a}) - \sum_{i\in \D}\phi_i(\TPR_{A=b}) \\
\text{EOp} &= \sum_{i\in \D}\left(\phi_i(\TPR_{A=a}) - \phi_i(\TPR_{A=b})\right) \rightarrow \phi_i(\EOp) = \phi_i(\TPR_{A=a}) - \phi_i(\TPR_{A=b}) 
\end{align*}

Similarly, $\bm{\phi}(\EOds)$ can be obtained as follows:
\begin{align*}
\text{EOdds} &= \frac{1}{2} ((\TPR_{A=a} - \TPR_{A=b}) +(\FPR_{A=a} - \FPR_{A=b})) \\
 &= \frac{(\sum_{i\in \D}\phi_i(\TPR_{A=a}) - \sum_{i\in \D}\phi_i(\TPR_{A=b})) +(\sum_{i\in \D}\phi_i(\FPR_{A=a}) - \sum_{i\in \D}\phi_i(\FPR_{A=b}))}{2} \\
 &= \frac{\sum_{i\in \D}\left(\phi_i(\TPR_{A=a}) - \phi_i(\TPR_{A=b})\right) + \sum_{i\in \D}\left(\phi_i(\FPR_{A=a}) - \sum_{i\in \D}\phi_i(\FPR_{A=b})\right)}{2} 
\\
 &= \frac{\sum_{i\in \D}\left(\left(\phi_i(\TPR_{A=a}) - \phi_i(\TPR_{A=b})\right) + \left(\phi_i(\FPR_{A=a}) - \phi_i(\FPR_{A=b})\right)\right)}{2} \rightarrow \\
\rightarrow \phi_i(\EOds) &= \frac{\left(\phi_i(\TPR_{A=a}) - \phi_i(\TPR_{A=b})\right) + \left(\phi_i(\FPR_{A=a}) - \phi_i(\FPR_{A=b})\right)}{2} 
\end{align*}

\subsection{Extension to multi-label and categorical sensitive attribute scenarios} \label{sec:app:multiclass}

As in the binary setting, the group fairness metrics are computed from TPR, TNR, FPR and FNR. Taking as an example TPR, the main change consists of replacing $y=1$ or $y=0$ for $y_j$=y:

\begin{align} 
\phi_i(\TPR_{\mid Y=y}) = \mathbb{E}_{j \sim p(\T\mid Y=y)}[\Phi_{i,j}] = \frac{\sum_{j\in \T} \Phi_{i,j} \mathbb{I}[y_j=y]}{|\{x: x \in \T | y=y\}|} = \overline{\bm{\Phi}}_{i,:|Y=y} \nonumber
\end{align}

The conditioned version $\phi_i(\TPR_a)$ may be obtained as: 
\begin{align}
 \phi_i(\TPR_{\mid Y=y, A=a}) = \mathbb{E}_{j \sim p(\T\mid Y=y, A=a)}[\Phi_{i,j}] = \frac{\sum_{j\in \T} \Phi_{i,j} \mathbb{I}[y_j=y, A_j=a]}{|\{x: x \in \T | y=y, A=a\}|}  = \overline{\bm{\Phi}}_{i,:|Y=y, A=a} \nonumber
\end{align}
where $y$ and $a$ can be categorical variables. In the scenario where $a$ is not a binary protected attribute, EOp is calculated as $\text{EOp}_a = |\text{TPR}-\text{TPR}_{A=a}|\: \forall a \in A$ , and then the maximum difference is selected as the unique EOp for the model $\text{EOp} = \max_{ \forall a \in A} \text{EOp}_a$, i.e. the EOp for the group that most differs from the TPR of the entire dataset. Therefore, $\phi_i(\EOp)$ for each data point is computed as
$\phi_i(\EOp) = \phi_i(\TPR_a) - \phi_i(\TPR)$ being $a$ the value of the protected attribute with maximum EOp. The same procedure applies to EOdds. In other words, $\phi_i(\EOp) = \mathbb{E}_{j \sim p(\T\mid Y=1, A=a)}[\Phi_{i,j}] - \mathbb{E}_{j \sim p(\T\mid Y=1)}[\Phi_{i,j}]$.

\section{Experiments}
The code for all the experiments described in this paper is publicly available at 
\url{https://anonymous.4open.science/r/fair-shap}.

\subsection{Experiments on synthetic datasets}
\label{app:synthetic}

\textbf{$\phi(\TPR)$ and $\phi(\TNR)$}
In this section, we present a visual analysis of $\phi(\TPR)$ and $\phi(\TNR)$ using a synthetic binary classification dataset featuring two Gaussian distributions.

\begin{figure*}[ht]
\begin{center}
\includegraphics[width=\textwidth]{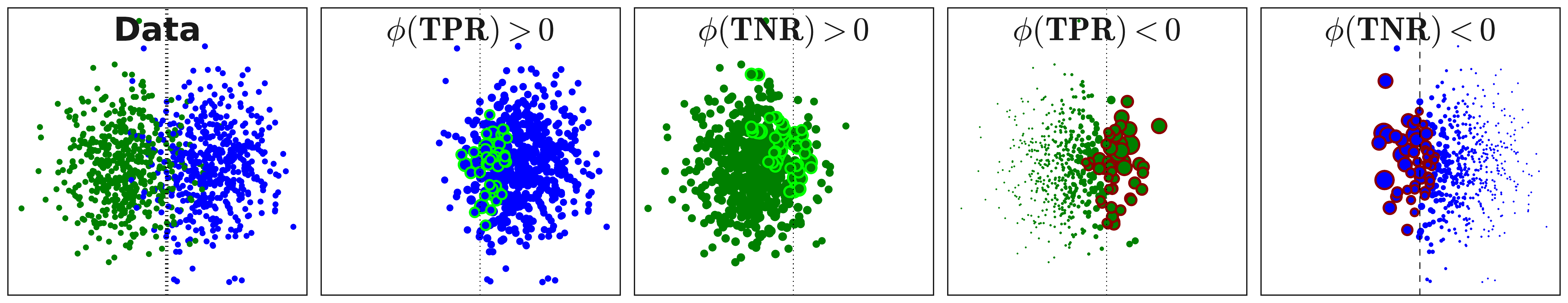}
\caption{Synthetic example with positive (Y=1, blue) and negative (Y=0,green) classes. Data points with the 50 largest (green) and smallest (red) $\phi_i$ are highlighted. Size $\propto|\phi_i|$.
}
\label{fig:synthetictpr}
\end{center}
\end{figure*}

\cref{fig:synthetictpr} illustrates the extent to which each data point contributes positively or negatively to the True Positive Rate (TPR) and True Negative Rate (TNR). Points with larger $\phi(\TPR)$ correspond to instances from the positive class located on the correct side near the decision boundary whereas points with smaller $\phi(\TPR)$ represent positive class points placed on the wrong side of the decision boundary. The same logic applies to $\phi(\TNR)$ with respect to the negative class points, providing intuitive insights related to the contributions to TPR or TNR of different data points.

\begin{figure}[ht]
\centering
\subfloat[]{\includegraphics[width=.22\linewidth]{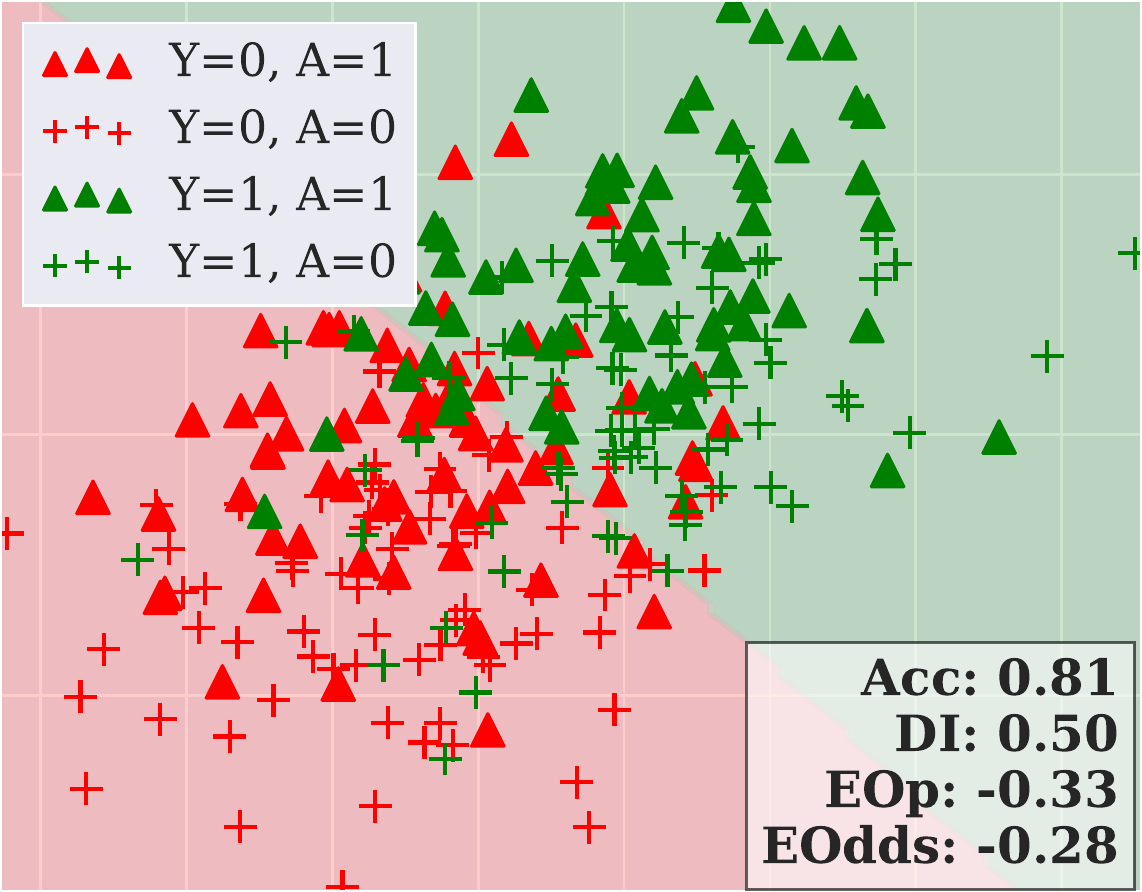}
}
\hfill
\subfloat[]{\includegraphics[width=.22\linewidth]{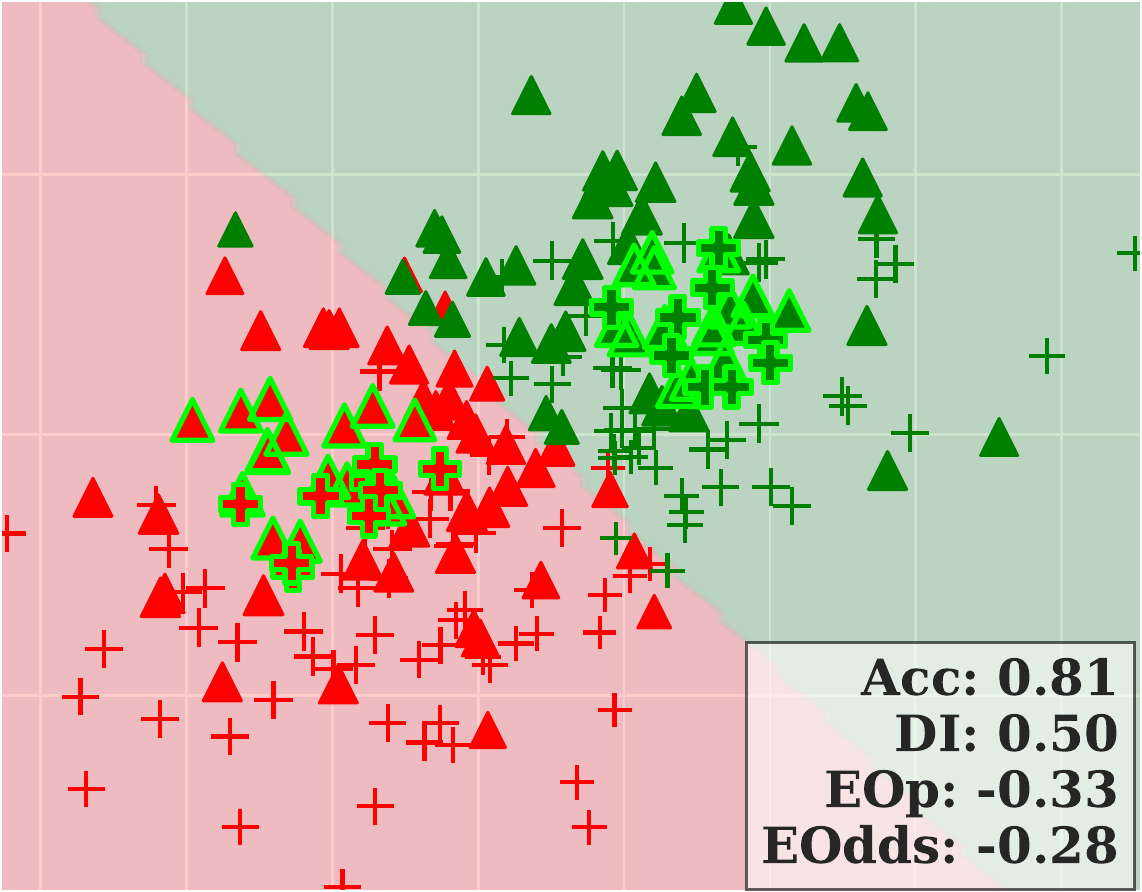}
}
\hfill
\subfloat[]{
\includegraphics[width=.22\linewidth]{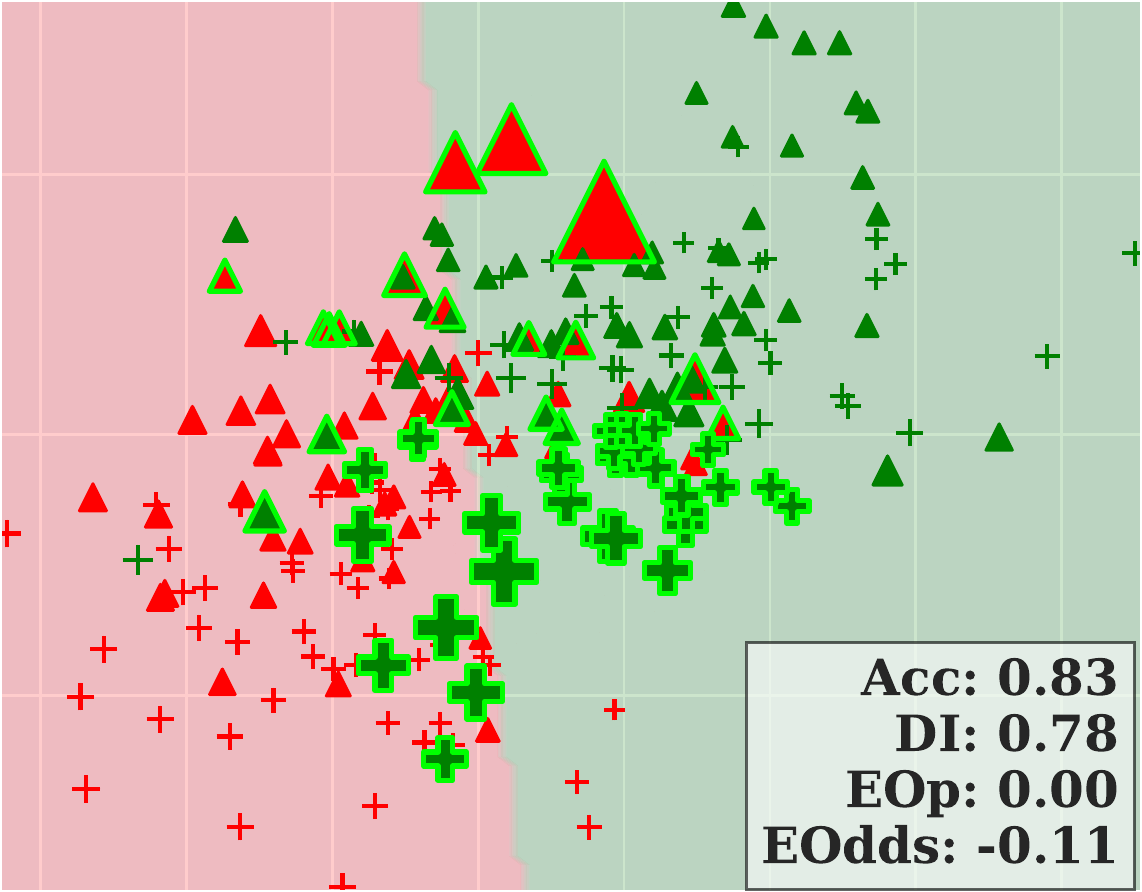}
}
\hfill
\subfloat[]{
    \includegraphics[width=.22\linewidth]{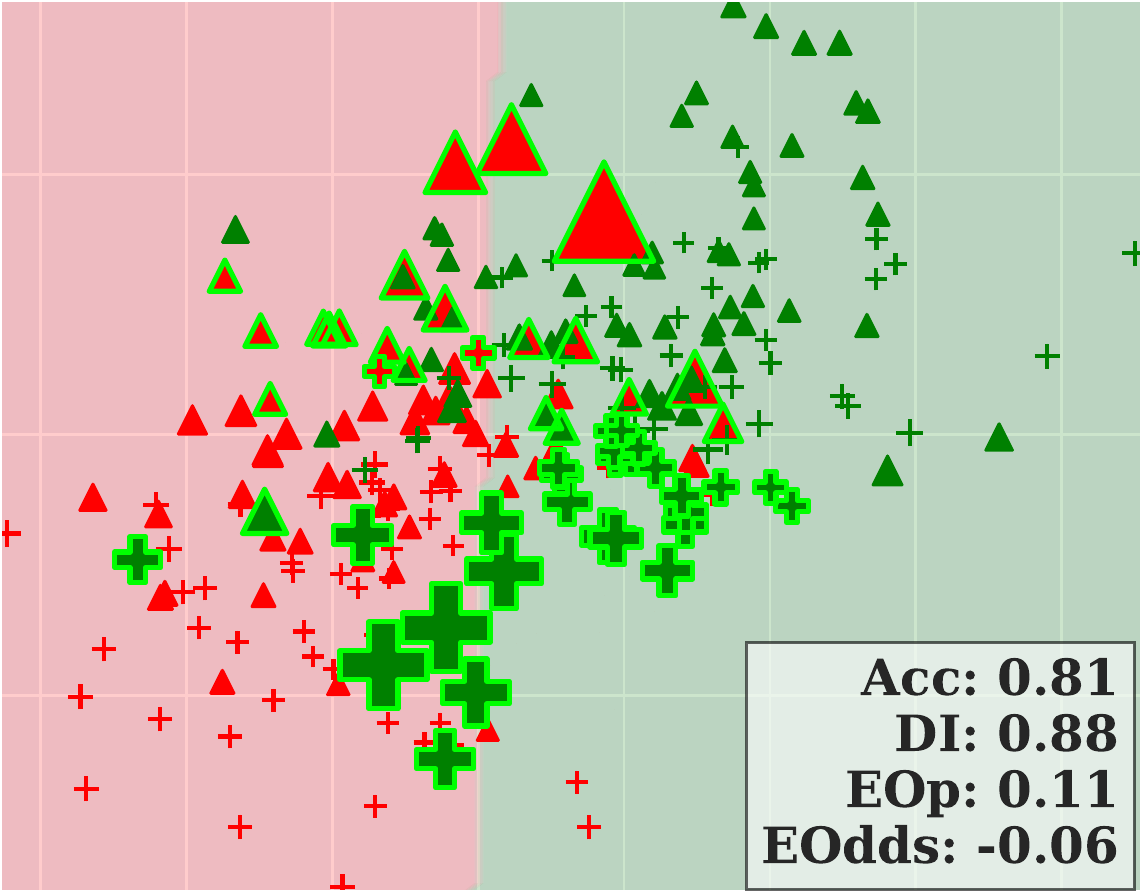}
}
\caption{Synthetic example where the group FPR  and the group FNR differences have different signs (Case I in \cite{gomez2017eodds}). (a) represents the original data and, in (b-d), the size of each data point is proportional to its (b) $\phi(\Acc)$, (c) $\phi(\EOp)$ and (d) $\phi(\EOds)$. The top-50 points, according to each $\mathbf{\phi}(\cdot)$, are highlighted in green. The label $Y=1$ corresponds to the favorable outcome (colored in \textbf{\textcolor{pastelgreen}{green}}), and the privileged group is defined by $A=1$ (represented as triangles~\triang). The label $Y=0$ corresponds to the unfavorable outcome (colored in \textbf{\textcolor{red}{red}}), and the disadvantaged group is defined by $A=0$ (represented as crosses~\crossb). Logistic Regression models are trained on different re-weighted versions of the data and evaluated using the same test split. Decision regions are shaded.}
\label{fig:synthFair}
\end{figure}

\textbf{A$\neq$Y} In this scenario, we generate synthetic data where the protected attribute $A$ and the label $Y$ are slightly correlated. Specifically, we employ Case I from \cite{gomez2017eodds} as a reference, where the disparity between the False Negative Rate (FNR) and the False Positive Rate (FPR) exhibits a distinct sign: larger FPR for the privileged group and larger FNR for the disadvantaged group. Consequently, the mean overlap occurs between the unfavorable-privileged and the favorable-disadvantaged classes.

\cref{fig:synthFair} visualizes data instances of this scenario as points, where the size of each point is proportional to its $\mathbf{|\phi}(\cdot)|$. Additionally, we highlight in green the top-50 points based on their $\mathbf{\phi}(\cdot)$. The label $Y=1$ corresponds to the favorable outcome (colored in green), and the privileged group is defined by $A=1$ (represented as triangles).
The label $Y=0$ corresponds to the unfavorable outcome (colored in red), and the disadvantaged group is defined by $A=0$ (represented as crosses). We train unconstrained Logistic Regression models on various versions of the data and assess their performance using the same test split.

The experimental results shown in \cref{fig:synthFair} illustrate significant changes in the decision boundaries of the models when trained using weights given by $\phi(\EOp)$ or $\phi(\EOds)$, yielding fairer models  while maintaining comparable or improved levels of accuracy. Moreover, the analysis reveals that both $\phi(\EOp)$ and $\phi(\EOds)$ predominantly prioritize instances in the unfavorable-privileged (red triangles) and favorable-disadvantaged groups (green crosses). 

\subsection{Description of the baselines used in the experiments \texorpdfstring{(\cref{sec:expAneqY})}{section}}
\label{app:baselineexplanation}

\textbf{Group RW~\citep{kamiran2012data}}: A group-based re-weighting method that assigns the same weights to all the samples from the same category or group according to the protected attribute. Weights are assigned to compensate that the expected probability if $A$ and $Y$ where independent on $\D$ is higher than the observed probability value.
$$
w_i(a_i,y_i)=\frac{|\{X\in\D|X(A)=a_i\}|\times|\{X\in\D|X(Y)=y_i\}|}{|\{\D\}|\times|\{X\in\D|X(A)=a_i, X(Y)=y_i\}|}
$$
Group RW does not require any additional parameters for its application. We use the implementation from \texttt{AIF360}~\citep{aif360}.

\textbf{Post-pro~\citep{hardt2016equality}}: A post-processing algorithmic fairness method that assigns different classification thresholds for different groups to equalize error rates. The method applies a threshold to the predicted scores to achieve this balance. 

In our experiments, we adopted the enhanced implementation of this method provided by the authors and based on the \texttt{error-parity} library~\citep{cruz2023unprocessing}.
This implementation makes its predictions using an ensemble of randomized classifiers instead of relying on a deterministic binary classifier. A randomized classifier  lies within the convex hull of the classifier's ROC curve at a specific target ROC point. This approach enhances the method's ability to satisfy the equality of error rates.

\textbf{LabelBias~\citep{jiang2020identifying}}: This model learns the weights in an iterative, in-processing manner based on the model's error. Consequently, this method is neither a pre-processing nor a model-agnostic approach. 

We used an implementation based on the \texttt{google-research/label\_bias}
repository, which is the official implementation of the original work. We applied the settings described in \citep{jiang2020identifying} and use a learning rate of $\mu=1$ with a fixed number of 100 iterations.

\textbf{Opt-Pre~\citep{calmon2017optimized}}: A model-agnostic pre-processing approach for algorithmic fairness based on feature and label transformations solving a a convex optimization. 

We used the pre-defined hyperparameters provided by both the authors~\citep[see][Supplementary~4.1-4.3]{calmon2017optimized} and the \texttt{AIF360} library: the discrimination parameter $\epsilon=0.05$; distortion constraints of [0.99, 1.99, 2.99], which are distance thresholds for individual distortions; and probability bounds of~[.1, 0.05, 0] for each threshold in the distortion constraints~\citep[Eq.~5]{calmon2017optimized}. We used the implementation from \texttt{AIF360}~\citep{aif360}.

\textbf{IFs~\citep{li2022individualreweighing}}: An Influence Function (IF)-based approach, where the influence of each training sample is modeled with regard to a fairness-related quantity and predictive utility. This is an in-processing, and re-training approach, as follows. First, a model is trained. Second, the Hessian vector product is computed for every sample $\mathbf{H}^{-1}_{\hat{\theta}(\mathbf{1})}\nabla^2_{\theta}\ell(x_i;\hat{\theta}(\mathbf{1})$, where the Hessian is defined as $\mathbf{H}_{\hat{\theta}(\mathbf{1})} = \sum_{i=1} \nabla^2_{\theta}\ell(x_i;\hat{\theta}(\mathbf{1}))$ and $\hat{\theta}(\mathbf{1})$ is the empirical risk minimization with equal sample weights. Third, the influence functions for every sample are obtained based on the vector products. Fourth, a linear problem based on these influence functions is solved to compute the weights. Finally, the model is re-trained with the new weights.
Notably, this method exhibits behavior resembling hard removal re-weighing --as observed in our experiments-- where the weights are either 0 or 1 for all samples, with no in-between values. This pattern aligns with the observations made by the authors themselves. While the method is theoretically categorized as individual re-weighting, in practice, it works as a sampling method.

We set the hyperparameters to the values reported by the authors for each dataset. Namely, for the German dataset: $\alpha=1$, $\beta=0$, $\gamma=0$ and $\mathrm{l2reg}=5.85$. For the Adult dataset: $\alpha=1$, $\beta=0.5$, $\gamma=0.2$ and $\mathrm{l2reg}=2.25$.
Finally, for the COMPAS dataset: $\alpha=1$, $\beta=0.2$, $\gamma=0.1$ and $\mathrm{l2reg}=37$. 
We use an implementation from the \texttt{influence-fairness} repository by Brandeis ML, which needs the request and installation of a Gurobi license.

\textbf{$\bm{\phi(\Acc)}$~\citep{ghorbani2019data}}: A method based on data re-weighting by means of an accuracy-based data valuation function without any fairness considerations. This method is explained in detail on \cref{appsubsec:knnsv}. We used our own efficient implementation using the Numba python library.

\subsection{Training set-up on the computer vision task}
\label{app:trainsetup}

In the experiment described in Section \ref{sec:experiments}, the Inception Resnet V1 model was initially pre-trained on the CelebA dataset and subsequently fine-tuned on LFWA. Binary cross-entropy loss and the Adam optimizer were used in both training phases. The learning rate was set to 0.001 for pre-training and reduced to 0.0005 for fine-tuning, each lasting 100 epochs. Training batches consisted of 128 images with an input shape of (160x160), and a patience parameter of 30 was employed for early stopping, saving the model with the highest accuracy on the validation set. The classification threshold for this model was set at 0.5.

\subsection{Impact of biased datasets on the models' evaluation}
\label{app:biasedeffects}

It is crucial to be aware that models trained on biased datasets may perform well in terms of accuracy and fairness when tested against themselves. However, when evaluated against fair datasets, their performance can deteriorate significantly. It is widely recognized that biased datasets can lead to biased machine learning models, which can perpetuate and exacerbate societal inequities. These models can reinforce existing biases and stereotypes, leading to unfair and discriminatory outcomes for certain groups, especially underrepresented or marginalized communities. Therefore, it is essential to develop fair reference datasets to ensure that machine learning models are tested in a way that accounts for the potential impact of bias and promotes fairness. In light of this, we present here the results of our experiments that illustrate the performance and fairness of a model trained and tested on three different dataset combinations: large yet biased datasets (LFWA and CelebA) and a smaller and unbiased dataset (FairFace). As illustrated in \cref{tab:bias-problem}, the performance of the models trained on biased datasets (LFWA and CelebA) and tested on fair datasets (FairFace) is significantly worse than when tested on the biased datasets.

\begin{table}[ht]
\caption{Sex classification results reported as Accuracy $\uparrow |$ Accuracy Disparity $\downarrow$ for an Inception Resnet V1 model trained and tested on different datasets. The protected attribute $A$ is sex. Note the degradation in performance when training on a biased dataset and evaluating on a fair dataset (marked in red font in the Table).}
\label{tab:bias-problem}
\begin{center}
\begin{small}
\begin{tabular}{lccc}
\toprule
Train $\backslash$ Test & FairFace & LFWA & CelebA \\
\midrule
FairFace   & 90.9 $|$ 0.01 &  95.7 $|$ 0.03 & 96.7 $|$ 0.09 \\
LFWA        & {\color{red}77.2 $|$ 0.49} &  96.6 $|$ 0.08 & 98.3 $|$ 0.02  \\
CelebA      & {\color{red}76.1 $|$ 0.61} &  96.9 $|$ 0.09 & 98.2 $|$ 0.01 \\
\bottomrule
\end{tabular}
\end{small}
\end{center} 
\end{table}

\subsection{Additional data pruning results}
\label{apx:sec:pruning}
\begin{figure}
\centering
\subfloat[]{
    \includegraphics[width=.82\textwidth,trim=0mm 0mm 0mm 10mm, clip]{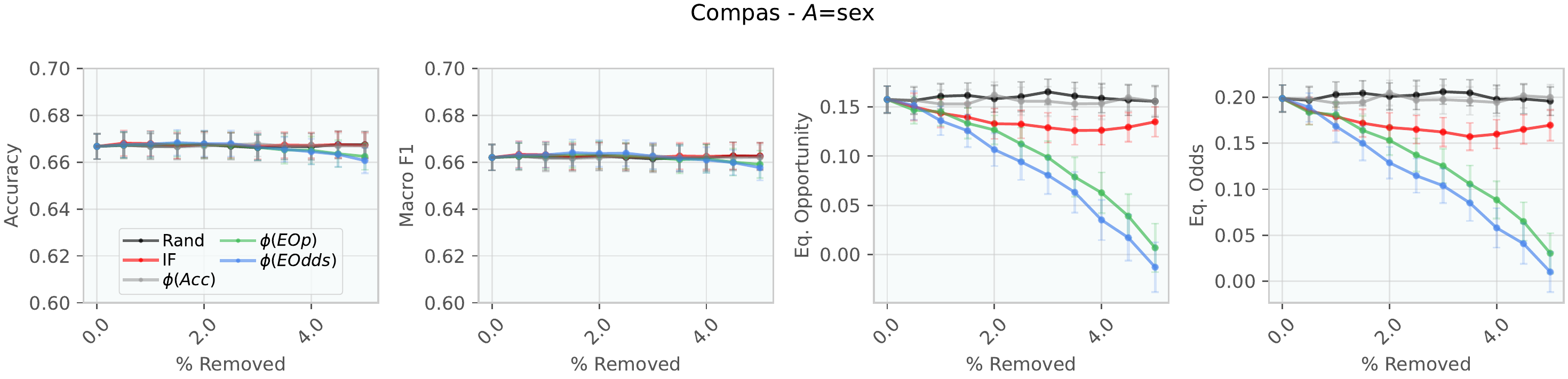}
}

\subfloat[]{
    \includegraphics[width=.82\textwidth,trim=0mm 0mm 0mm 10mm, clip]{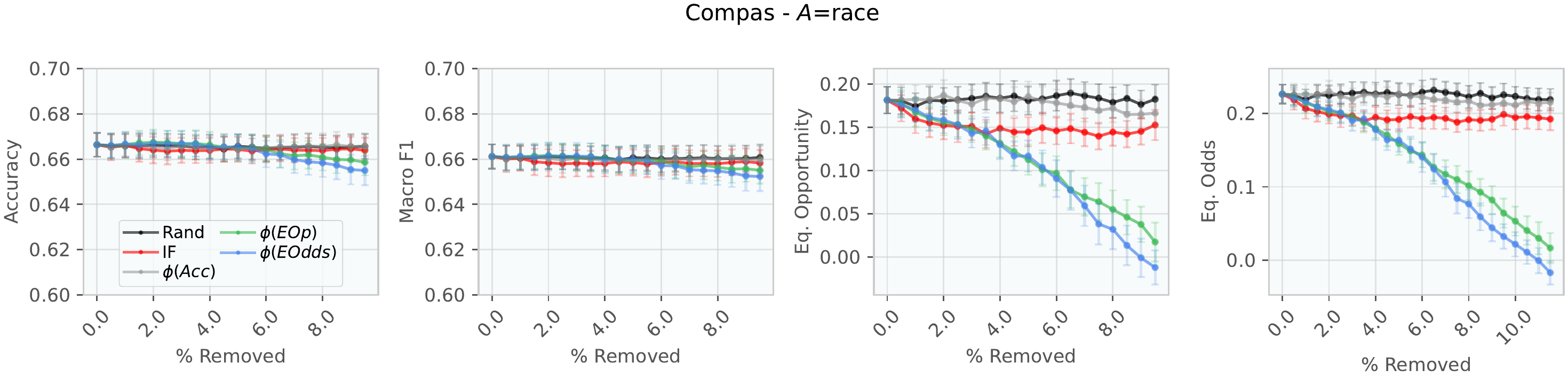}
}

\subfloat[]{
    \includegraphics[width=.82\textwidth,trim=0mm 0mm 0mm 10mm, clip]{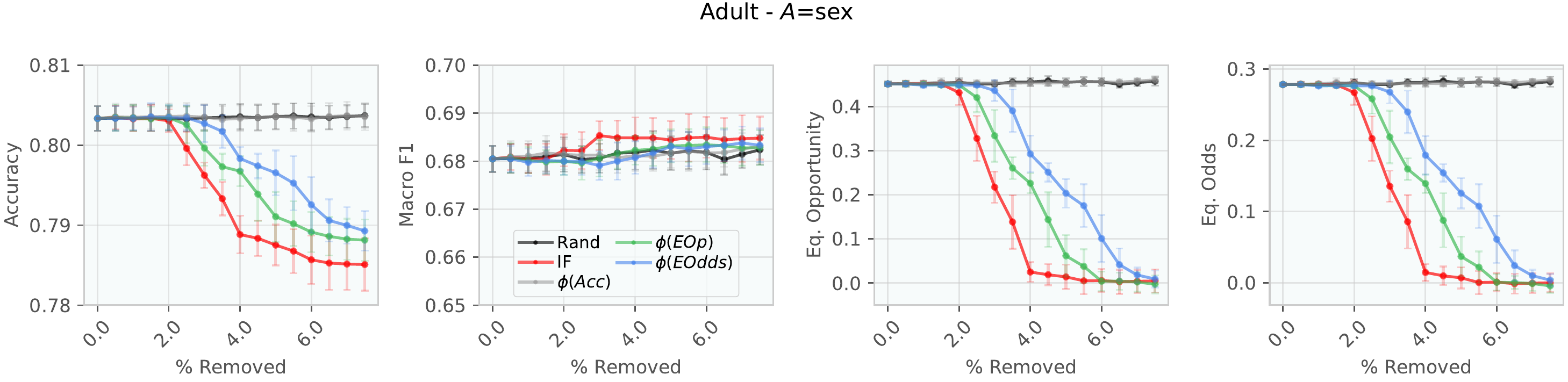}
}

\subfloat[]{
    \includegraphics[width=.82\textwidth,trim=0mm 0mm 0mm 10mm, clip]{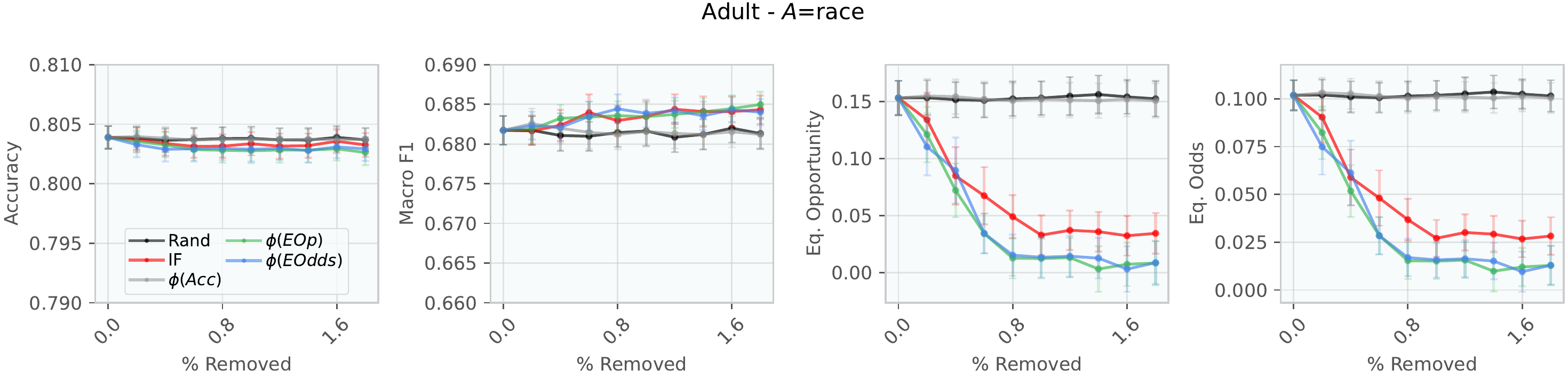}
}

\subfloat[]{
    \includegraphics[width=.82\textwidth,trim=0mm 0mm 0mm 10mm, clip]{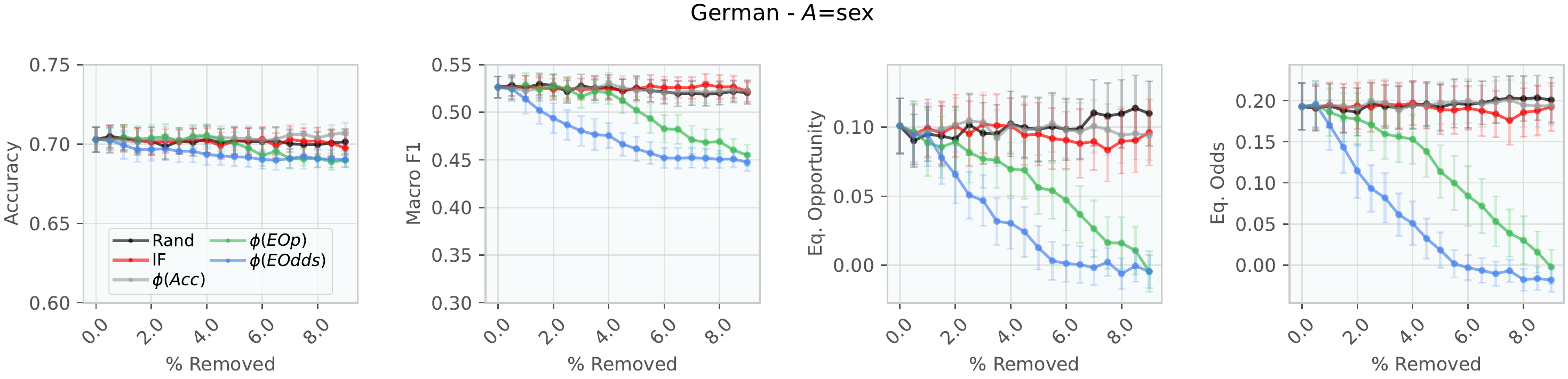}
}

\subfloat[]{
    \includegraphics[width=.82\textwidth,trim=0mm 0mm 0mm 10mm, clip]{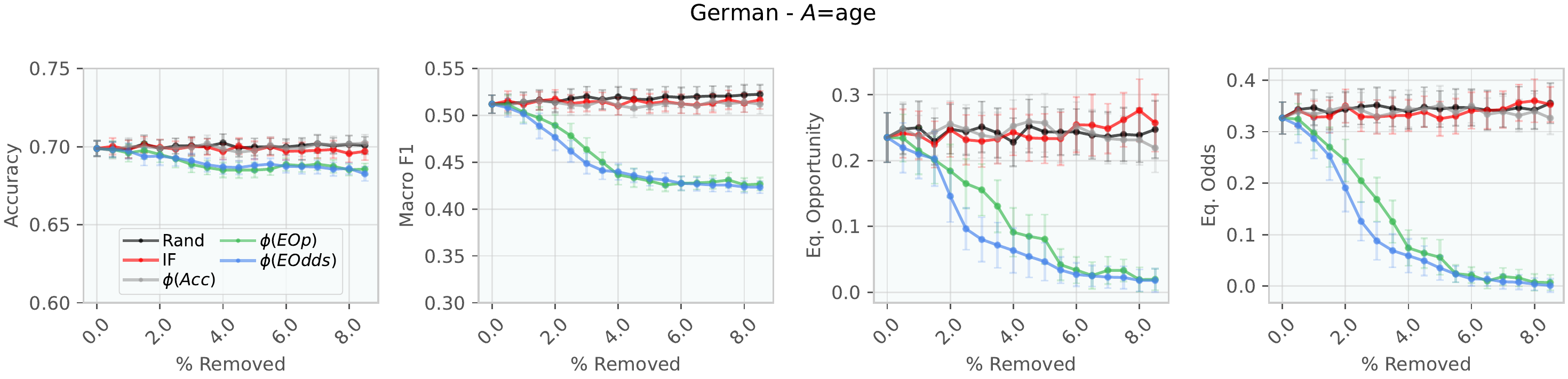}
}
\caption{Data pruning experiments. Subfigures depicts results for (a) Compas - Sex, (b) Compas - Race, (c) Adult - Sex, (d) Adult - Race, (e) German - Sex and (f) German - Age.}
\label{app:fig:pruning}
\end{figure}

The experimental results included in the main paper describe a subset of all the considered tasks.
\cref{app:fig:pruning} provides the results of the complete set of datasets and metrics.

\subsection{Impact of the reference dataset's size}
\label{app:sec:validationsize}

\begin{figure}[ht]
\centering
\subfloat[]{
\includegraphics[width=.28\linewidth]{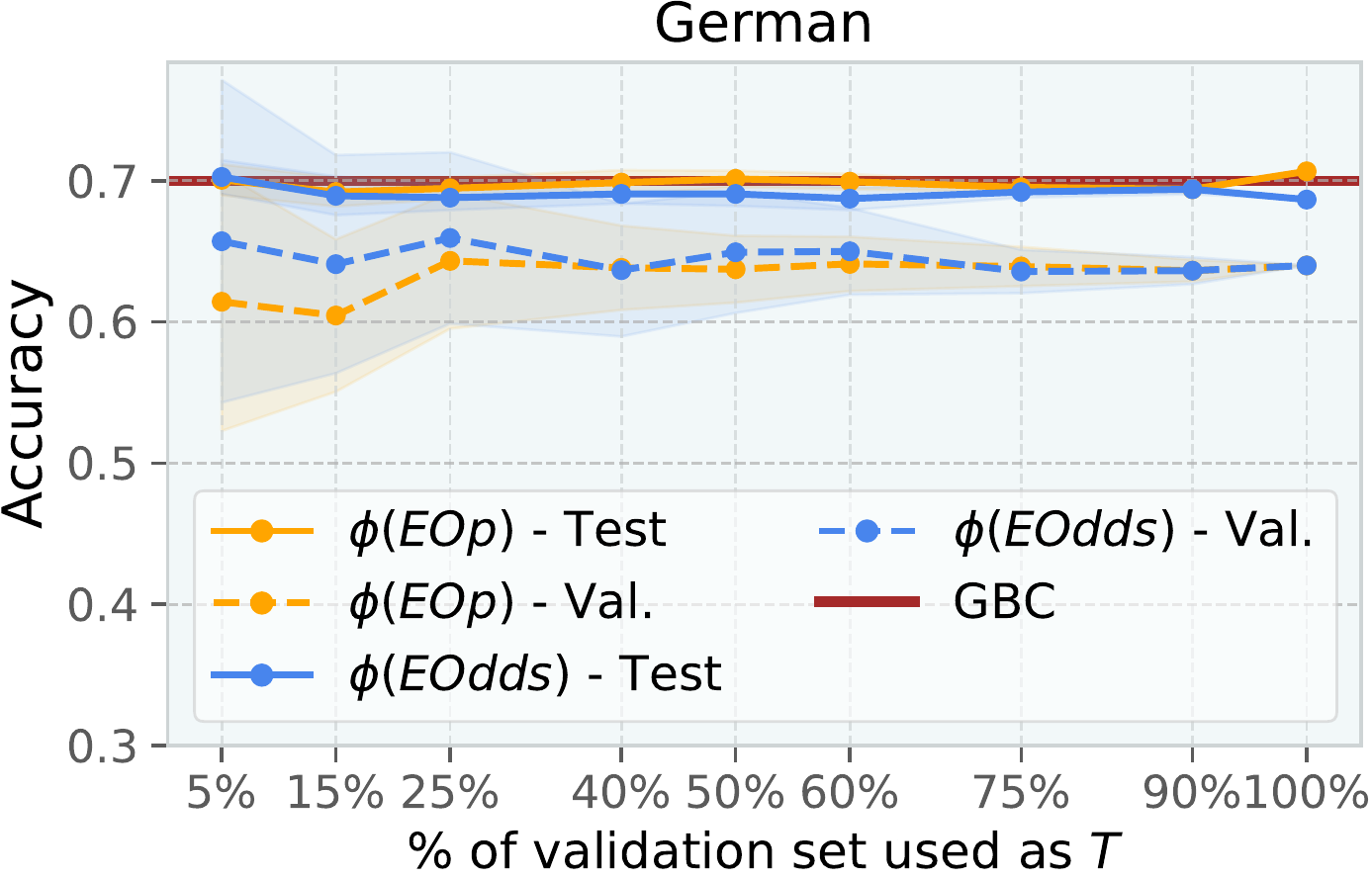}
}
\hfill
\subfloat[]{
\includegraphics[width=.28\linewidth]{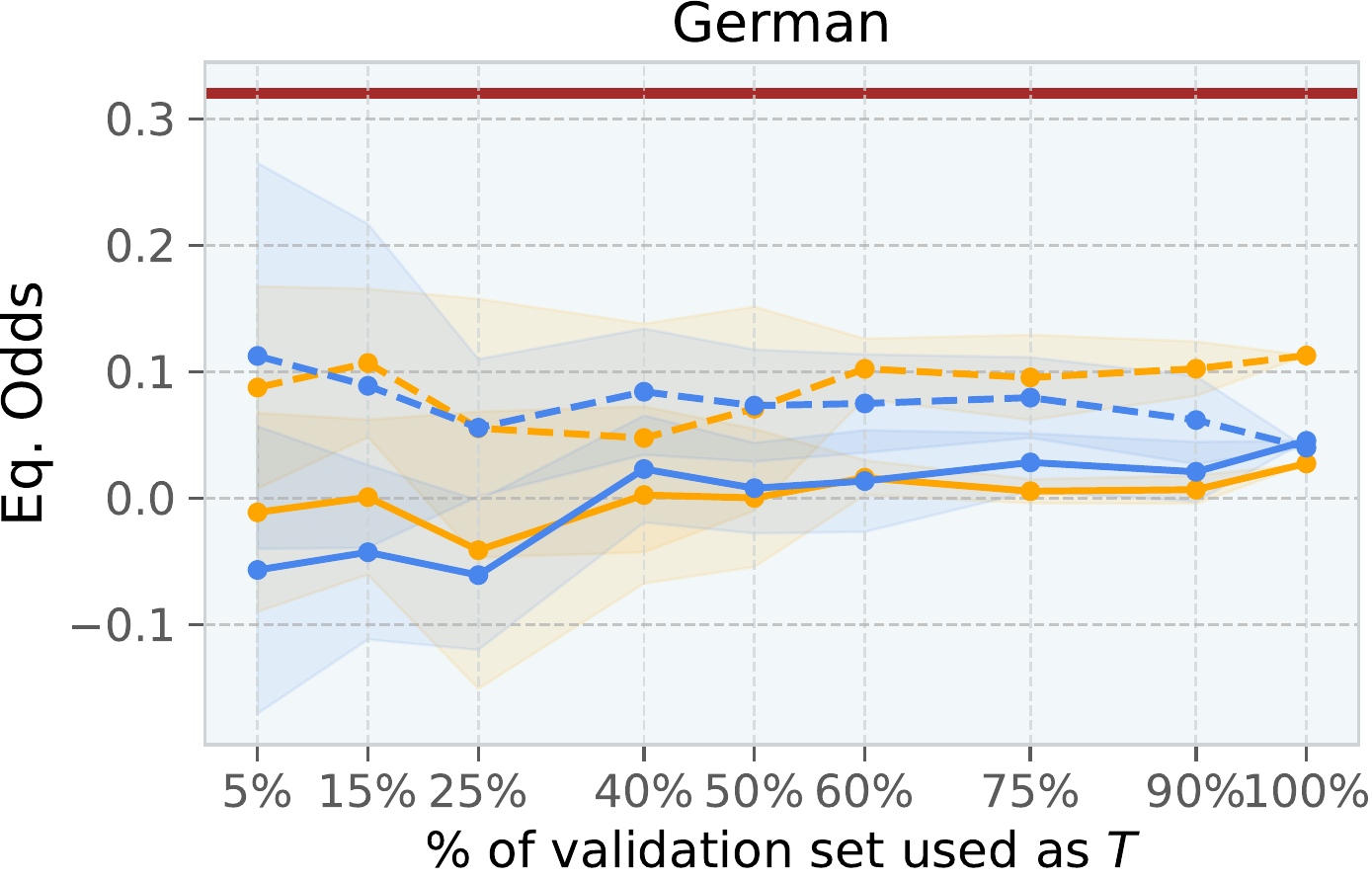}
}
\hfill
\subfloat[]{
\includegraphics[width=.28\linewidth]{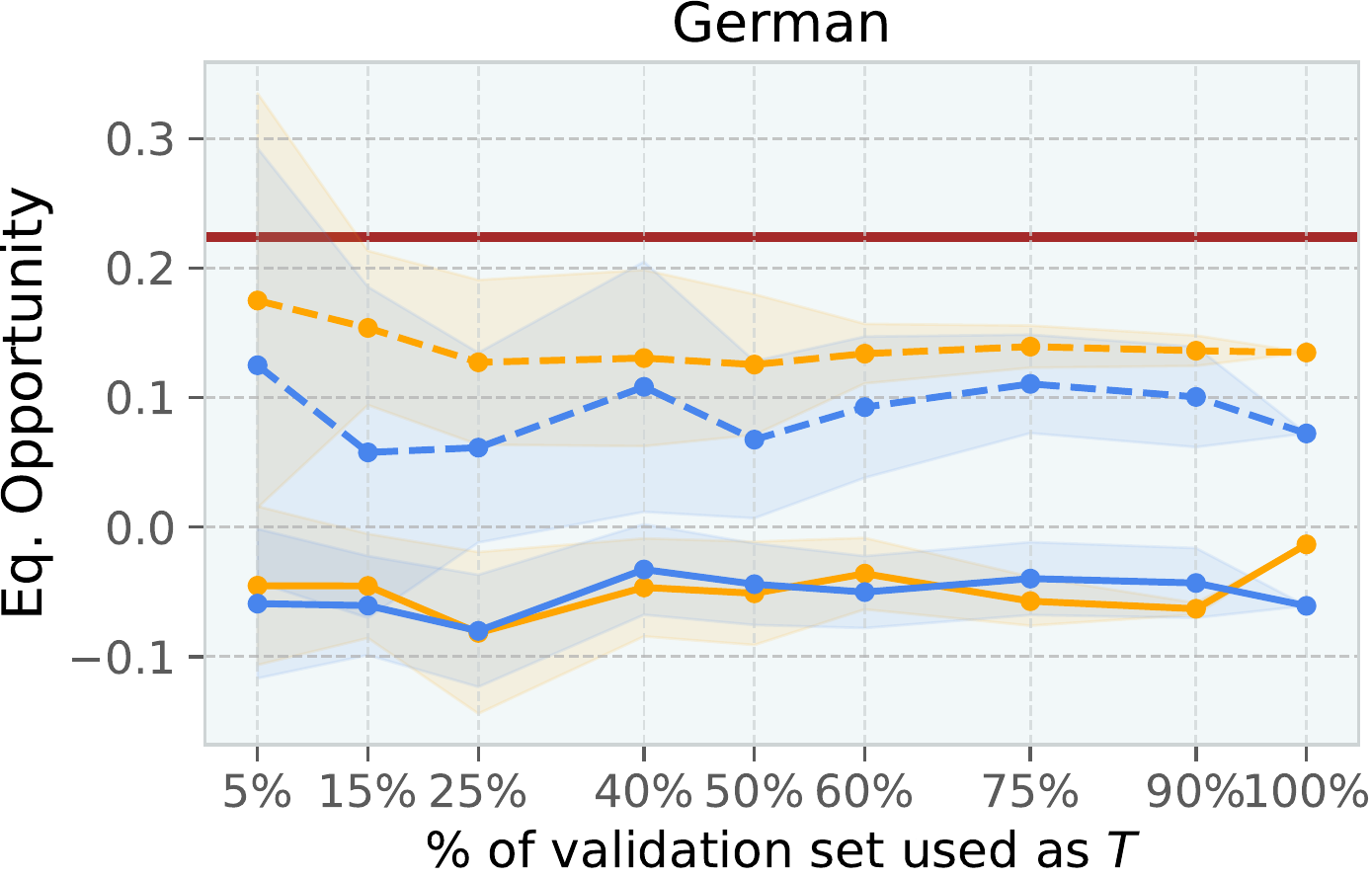}
}

\subfloat[]{
\includegraphics[width=.28\linewidth]{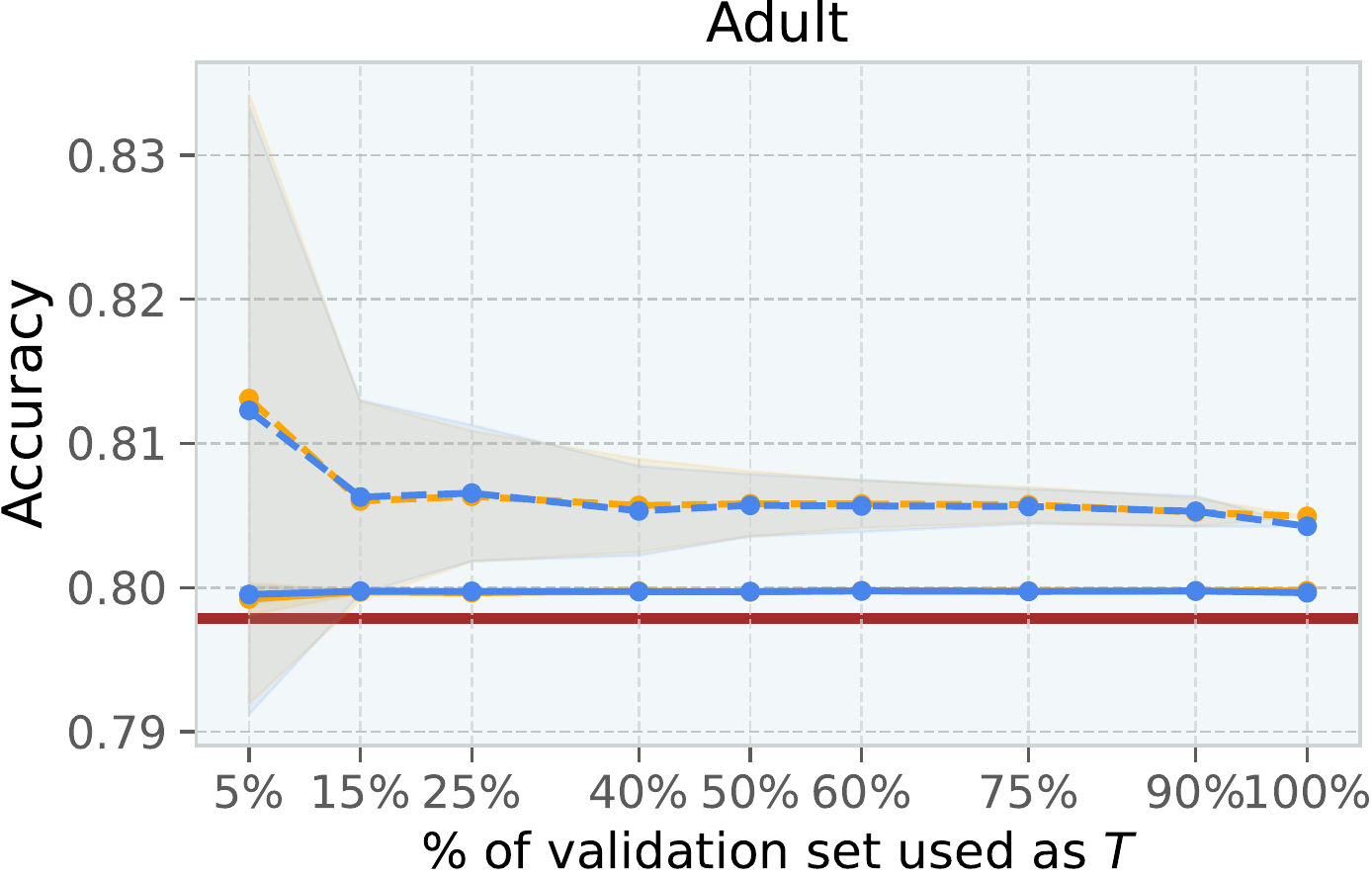}
}
\hfill
\subfloat[]{
\includegraphics[width=.28\linewidth]{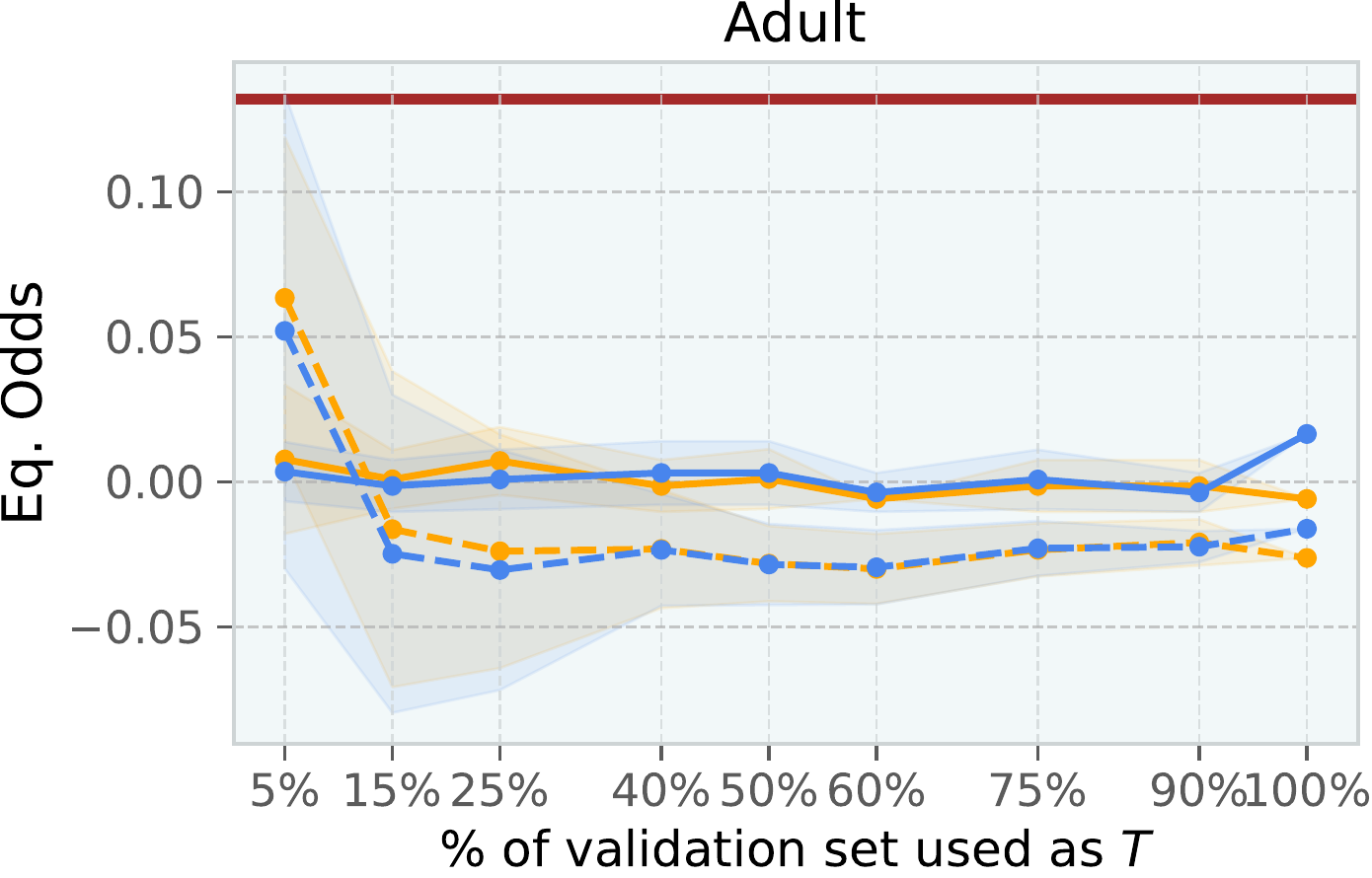}
}
\hfill
\subfloat[]{
\includegraphics[width=.28\linewidth]{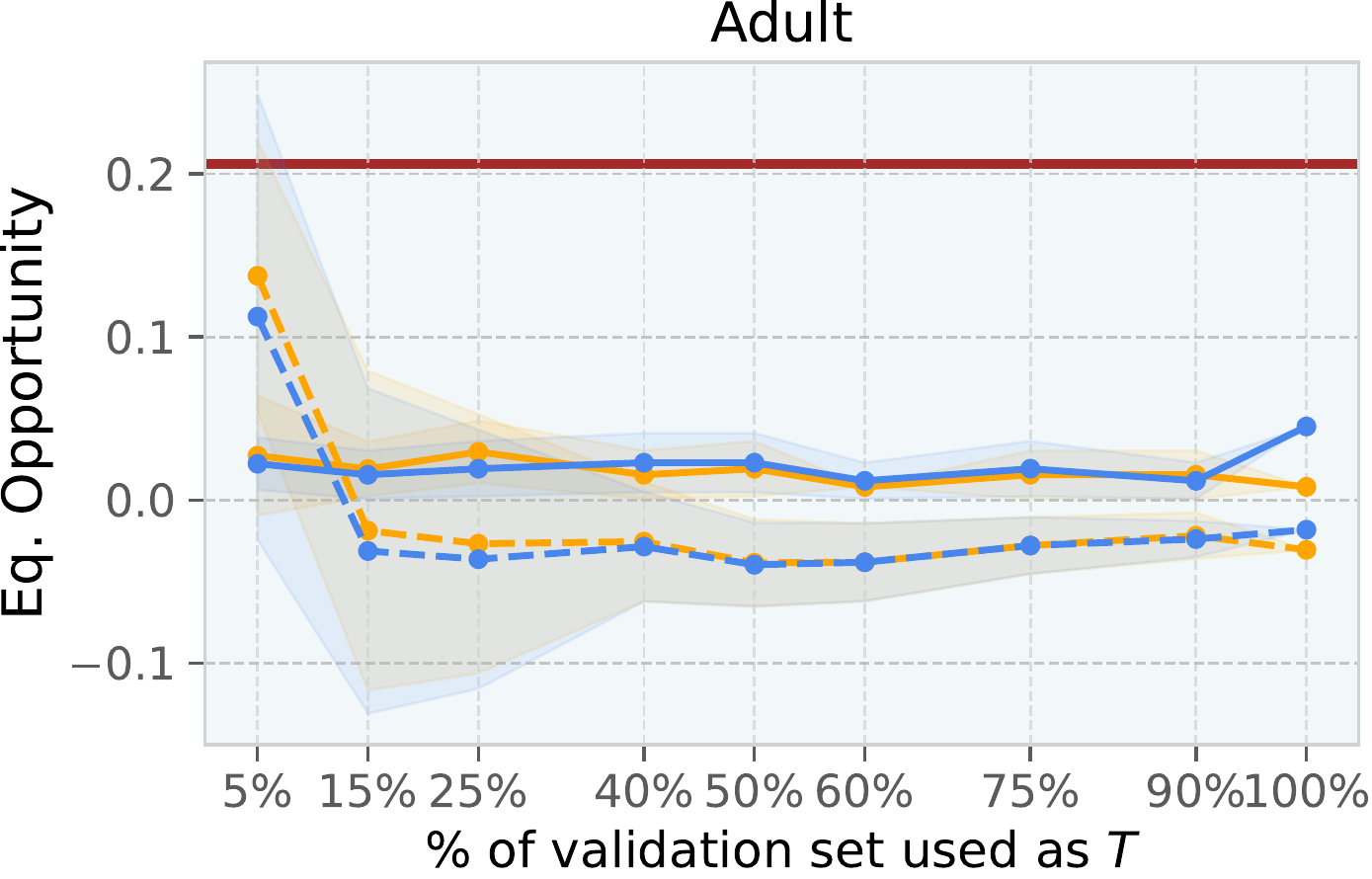}
}

\subfloat[]{
\includegraphics[width=.28\linewidth]{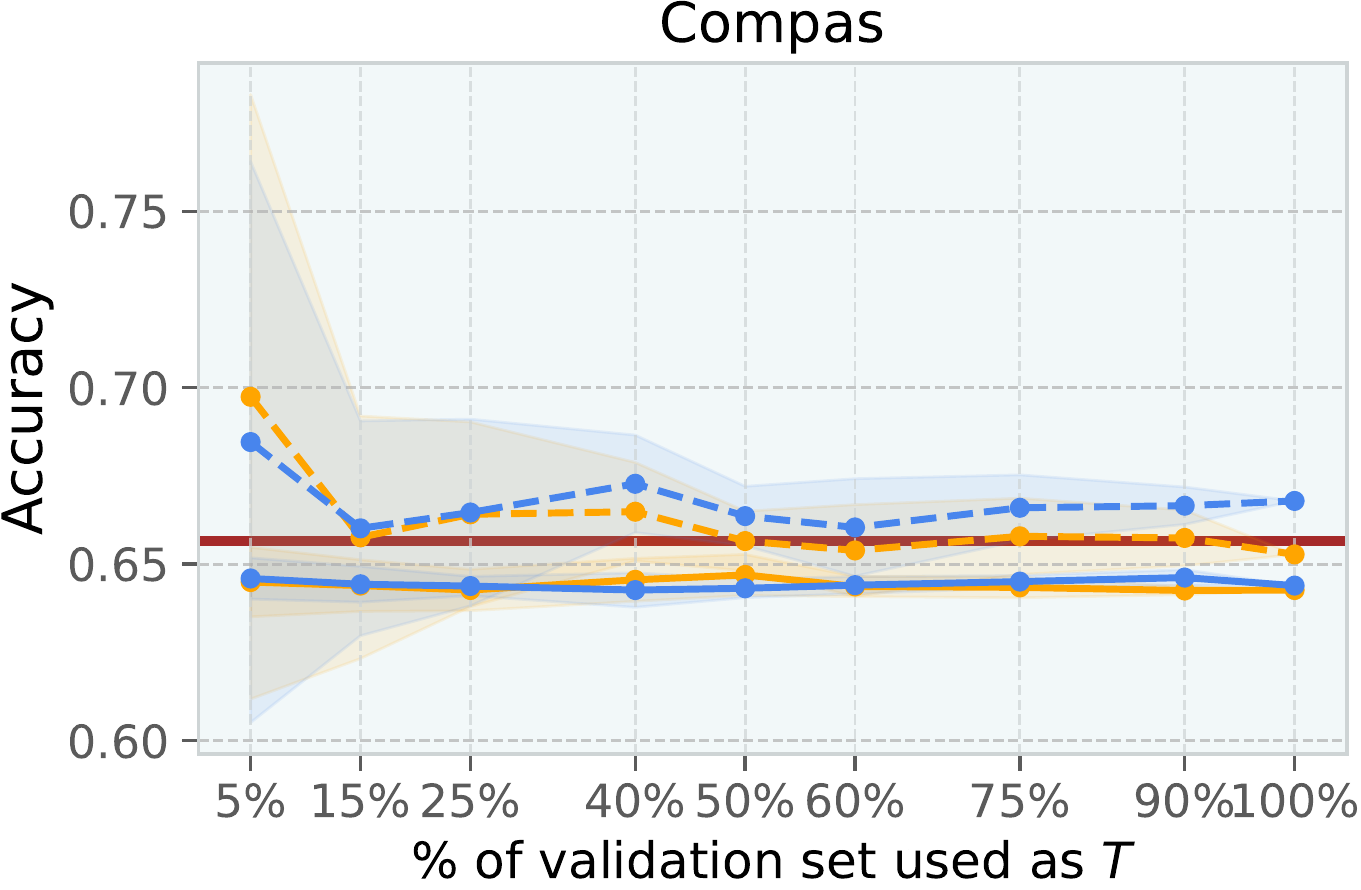}
}
\hfill
\subfloat[]{
\includegraphics[width=.28\linewidth]{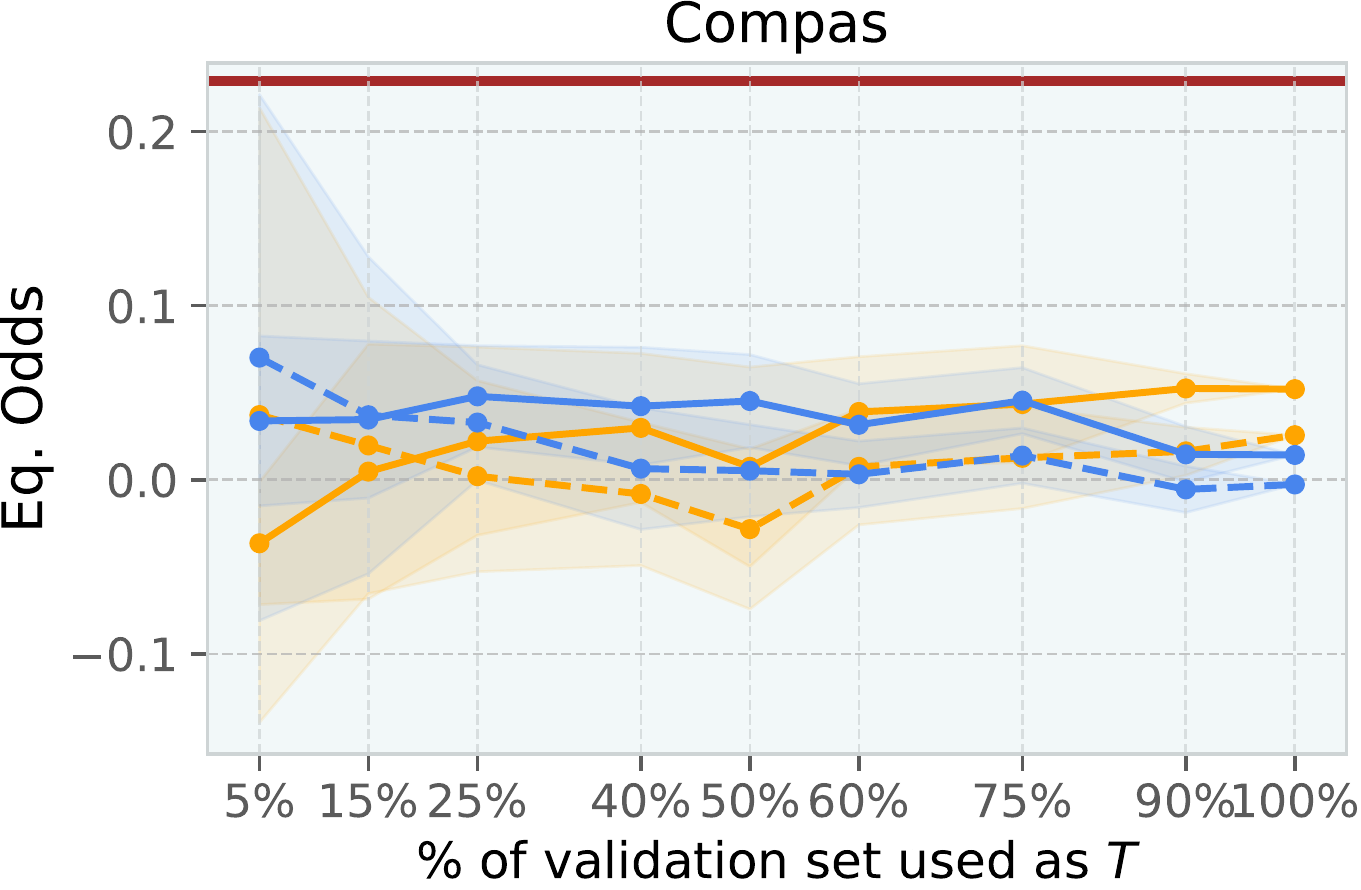}
}
\hfill
\subfloat[]{
\includegraphics[width=.28\linewidth]{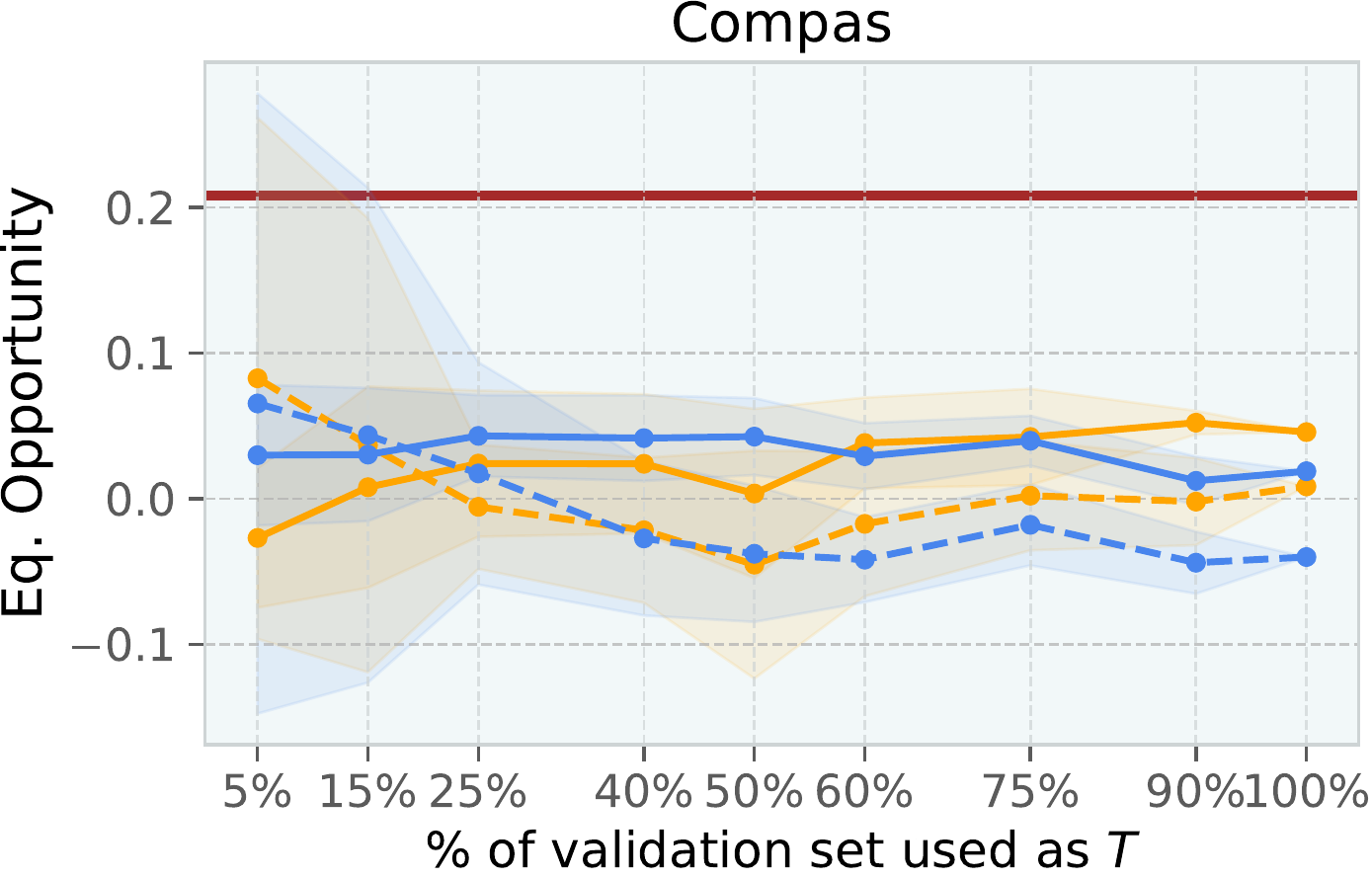}
}
\caption{Accuracy and fairness metrics when applying data re-weighting via \texttt{FairShap} (\textcolor{myorange}{$\phi(\text{EOp})$} and \textcolor{myblue}{$\phi(\text{EOdds})$}) as the size of the validation sets $\T$ increases, evaluated on both validation (- -) and test sets (---). The performance of the baseline GBC without re-weighting is shown as a red line. From top to bottom, the rows correspond to the German, Adult and COMPAS datasets, respectively. From left to right, the columns depict the Accuracy, Equalized Odds and Equal Opportunity, respectively.} \label{fig:ablation}
\end{figure}

In this section, we examine the influence of the size of the reference dataset, $\T$, and the impact of the alignment between $\T$ and the test set on the effectiveness of \texttt{FairShap}'s re-weighting. 
To do so, we perform an ablation study. 
We partition the three benchmark datasets (German, Adult and COMPAS) into training (60\%, $\D$), validation (20\%), and testing (20\%). We select subsets from the validation dataset --ranging from 5\% to 100\% of its size-- and use them as $\T$. For each subset, we compute \texttt{FairShap}'s weights on $\D$ with respect to $\T$, train a Gradient Boosting Classifier (GBC) model and evaluate its performance on the test set.  
This process is repeated 10 times with reported results comprising both mean values and standard deviations shown in \cref{fig:ablation}. 

As shown in the Figures, the size of the validation dataset has a discernible impact on the variance of the evaluation metrics, both in terms of accuracy and fairness. Increasing the size of the reference dataset $T$ leads to a notable reduction in the variability of the outcomes. However, averages for all the metrics remain stable across different sizes. 

\subsection{Accuracy metrics}
\label{app:utilitymetrics}

The reported experiments with the tabular datasets (i.e. German, Adult and COMPAS) include different accuracy metrics to evaluate the performance of the models.  

In imbalanced datasets, where one class significantly outweighs the other in terms of the number of examples, accuracy does not provide a good assessment of a model's performance, given that high accuracies might be obtained by a simple model that predicts the majority class. In these cases, the F1 metric is more appropriate, defined as~$2\times\frac{\text{precision}\times\text{recall}}{\text{precision}+\text{recall}}$, where precision is given by $\frac{\text{TP}}{\text{TP}+\text{FP}}$ and recall by $\text{TPR}=\frac{\text{TP}}{\text{TP}+\text{FN}}$. 

However, the F1 metric is only meaningful when the positive class is the minority class. Otherwise, i.e. if the positive class is the majority class, a constant classifier that always predicts the positive class can achieve a high F1 values. For example, in a scenario where the positive class has 100 examples and the negative class only 20, a simple model that always predicts the positive class will get an accuracy of 0.83 and a F1 score of 0.91. However, the F1 score for the negative class would be 0 in this case. 

The Macro-F1 metric arises as a solution to this scenario. Unlike the standard F1 score, the Macro-F1 computes the average of the F1 scores for each class. Thus, the Macro-F1 score can provide insight into the model's performance on every class for imbalanced datasets. 

Thus, in our experiments with tabular data, we report the Macro-F1 scores. 

\subsection{Dataset statistics} \label{app:datasetstats}

\textbf{Image Datasets} Total number of images and male/female distribution from the CelebA, LFWA and FairFace datasets are shown on \cref{tab:imagestatistics}.

\begin{table}[ht]
\caption{Face Datasets Statistics. Rows stand for \texttt{\#male$|$\#female}.}
\label{tab:imagestatistics}
\begin{center}
\begin{small}
\begin{tabular}{lccc}
\toprule
Dataset     & Train         & Validation    & Test      \\
\midrule
CelebA      & 94,509$|$68,261 & 11,409$|$8,458  & 12,247$|$7,715  \\
LFWA        & 7,439$|$2,086   & 2,832$|$876    & --  \\
FairFace   & 45,986$|$40,758 & 9,197$|$8,152   & 5,792$|$5,162  \\
\bottomrule
\end{tabular}
\end{small}
\end{center}
\end{table}

\noindent \textbf{Fairness Benchmark Datasets} The tables below summarize the statistics of the German, Adult and COMPAS datasets in terms of the distribution of labels and protected groups. Note that all the nomenclature regarding the protected attribute names and values is borrowed from the official documentation of the datasets.

\cref{tab:germanstats} shows the distributions of sex and label for the German dataset \citep{kamiran2009german}. It contains 1,000 examples with target binary variable the individual's \emph{credit risk} and protected groups \emph{age} and \emph{sex}. We use `Good Credit' as the favorable label (1) and `Bad Credit` as the unfavorable one (0). Regarding \emph{age} as a protected attribute, `Age>25' and 'Age<25' are considered the favorable and unfavorable groups, respectively. When using \emph{sex} as protected attribute, \textit{male} and \textit{female} are considered the privileged and unprivileged groups, respectively. Features used are the one-hot encoded credit history (delay, paid, other), one-hot encoded savings (>500, <500, unknown) and one-hot encoded years of employment (1-4y, >4y, unemployed).

\begin{table}[ht]
\caption{Tabular datasets statistics. (a) German Credit, (b) Adult Income and (c) COMPAS.} \label{tab:datasetstats}
\centering
\subfloat[]{
        \begin{small}
        \begin{tabular}{l|cc|c}
            \toprule
            A\textbackslash Y & Bad & Good & Total \\
            \midrule
            Male    & 191   & 499  & 690 (69\%)\\
            Female  & 109   & 201  & 310 (31\%)\\
            \midrule
            Age$>$25 & 220 & 590 & 810 (81\%)\\
            Age$<$25 & 80 & 110 & 190 (19\%)\\
            \midrule
            Total   &  300 (30\%) & 700 (70\%)  & 1,000\\
            \bottomrule
        \end{tabular}
         \label{tab:germanstats}
        \end{small}
}
\subfloat[]{
        \begin{small}
        \begin{tabular}{l|cc|c}
            \toprule
            A\textbackslash Y & $<$50k & $>$50k  & Total \\
            \midrule
            White           & 31,155 & 10,607 & 41,762 (86\%)\\
            non-White       & 6,000  & 1,080  & 7,080 (14\%)\\
            \midrule
            Male            & 22,732 & 9,918  & 32,650 (67\%)\\
            Female          & 14,423 & 1,769  & 16,192 (33\%)\\
            \midrule
            Total           & 37,155 (76\%)&   11,687 (24\%)   & 48,842 \\
            \bottomrule
        \end{tabular}
        \label{tab:adultstats}
        \end{small}
}
    
\subfloat[]{
        \begin{small}
        \begin{tabular}{l|cc|c}
            \toprule
            A\textbackslash Y & Recid & No Recid & Total \\
            \midrule
            Male        & 2,110      & 2,137      &   4,247 (80\%)\\
            Female      & 373       & 658       &   1,031 (20\%)\\
            \midrule
            Caucasian       & 822       & 1,281      & 2,103 (40\%)\\
            non-Cauc.  & 1,661      & 1,514      & 3,175 (60\%)\\
            \midrule
            Total       & 2,483 (47\%) & 2,795 (53\%)  & 5,278\\
            \bottomrule
        \end{tabular}
         \label{tab:compasstats}
        \end{small}
}
\end{table}

\cref{tab:adultstats} depicts the data statistics for the Adult Income dataset \citep{kohavi1996adult}. This dataset contains 48,842 examples where the task is to predict if the \emph{income} of a person is more than 50k per year, being >50k considered as the favorable label (1) and <50k as the unfavorable label (0). The protected attributes are \emph{race} and \emph{sex}. When \emph{race} is the protected attribute, \textit{white} refers to the privileged group and \textit{non-white} to the unprivileged group. With \emph{sex} as protected attribute, \textit{male} is considered the privileged group and \textit{female} the disadvantaged group.  The features are the one-hot encoded age decade (10, 20, 30, 40, 50, 60, >70) and education years (<6, 6, 7, 8, 9, 10, 11, 12, >12).

\cref{tab:compasstats} contains the statistics about the COMPAS \citep{angwin2016machine} dataset. This dataset has 5,278 examples with target binary variable \emph{recidivism}. We use \textit{Did recid} as the unfavorable label (0) and \textit{No recid} as the favorable label (1). When \emph{sex} is the protected attribute, \textit{male} is the disadvantaged group and \textit{female} as the privileged one. When using \emph{race} as protected attribute, \textit{caucasian} is the privileged group and \textit{non-caucasian} the disadvantaged one. Regarding the features, we use one-hot encoded age (<25, 25-45, >45), one-hot prior criminal records of defendants (0, 1-3, >3) and one-hot encoded charge degree of defendants (Felony or Misdemeanor).

All datasets are pre-processed using AIF360 by \citep{aif360}, which use the same pre-processing as in Calmon et al.~\citep{calmon2017optimized}.

\end{document}